%% file: main.tex
\let\clineorig\cline
\documentclass[sn-mathphys-num]{sn-jnl}
\let\cline\clineorig
\usepackage{graphicx}%
\usepackage{amsmath,amssymb,amsfonts}%
\usepackage{amsthm}%
\usepackage{mathrsfs}%
\usepackage[title]{appendix}%
\usepackage[table]{xcolor}
\usepackage{textcomp}%
\usepackage{manyfoot}%
\usepackage{algorithm}%
\usepackage{algorithmicx}%
\usepackage{algpseudocode}%
\usepackage{listings}%
\usepackage{array}
\usepackage{microtype}
\usepackage{tabularx}
\usepackage{bm}
\usepackage{booktabs}
\usepackage{tikz}        
\usepackage{comment}
\usepackage{hyperref}
\usepackage{amssymb}
\usepackage{mathtools}
\usepackage{amsthm}
\usepackage{multirow}
\usepackage{float}
\usepackage{graphicx}
\usepackage{makecell}
\usepackage{longtable}
\usepackage{caption}
\usepackage{subcaption}
\usepackage{booktabs}   
\usepackage{graphicx}   
\usepackage{xurl}
\makeatletter
\providecommand{\headerps@out}[1]{}
\makeatother

\usepackage[capitalize,noabbrev]{cleveref}
                               
\theoremstyle{plain}

\theoremstyle{definition}

\theoremstyle{remark}

\begin{document}

\title[]{Benchmarking Generalist and Specialist Time-Series Foundation Models for PPG-Based Health Assessment}

\author[1]{\fnm{Saurabh} \sur{Kataria*}} \email{saurabh.kataria@emory.edu}
\equalcont{Corresponding Authors}
\author[1,2]{\fnm{Yi} \sur{Wu*}}\email{yi.wu-1@ou.edu}
\equalcont{Corresponding Authors}
\author[1]{\fnm{Zhaoliang} \sur{Chen*}}\email{david.chen.czl@outlook.com}
\author[1]{\fnm{Hyunjung Gloria} \sur{Kwak}} 
\author[4]{\fnm{Yuhao} \sur{Xu}}    
\author[4]{\fnm{Lovely Yeswanth} \sur{Panchumarthi}} 
\author[1]{\fnm{Ran} \sur{Xiao}}
\author[1]{\fnm{Jiaying} \sur{Lu}} 
\author[1]{\fnm{Ayca} \sur{Ermis}}  
\author[1]{\fnm{Anni} \sur{Zhao}}
\author[1]{\fnm{Runze} \sur{Yan}}
\author[1]{\fnm{Minxiao} \sur{Wang}}
\author[1]{\fnm{Alex} \sur{Fedorov}}
\author[4]{\fnm{Zewen} \sur{Liu}}
\author[2]{\fnm{Xu} \sur{Wu}}   
\author[2]{\fnm{Hongyu} \sur{Wu}}

\author[4]{\fnm{Wei} \sur{Jin}}
\author[4]{\fnm{Carl} \sur{Yang}}
\author[3]{\fnm{Jocelyn} \sur{Grunwell}}
\author[3]{\fnm{Stephanie R.} \sur{Brown}}
\author[5]{\fnm{Amit} \sur{Shah}}
\author[6]{\fnm{Craig} \sur{Jabaley}}
\author[7]{\fnm{Tim} \sur{Buchman}}
\author[8]{\fnm{Sivasubramanium V} \sur{Bhavani}}
\author[9]{\fnm{Randall J.} \sur{Lee}}
\author[1]{\fnm{Xiao} \sur{Hu*}}\email{xiao.hu@emory.edu}
\equalcont{Corresponding Authors}


\affil[1]{\orgdiv{Nell Hodgson Woodruff School of Nursing}, \orgname{Emory University, Atlanta, USA}}

\affil[2]{\orgdiv{School of Computer Science}, \orgname{University of Oklahoma, USA}}

\affil[3]{\orgdiv{Department of Pediatrics}, \orgname{Emory University School of Medicine, Atlanta, USA}}

\affil[4]{\orgdiv{Department of Computer Science}, \orgname{Emory University, Atlanta, USA}}

\affil[5]{\orgdiv{Department of Epidemiology}, \orgname{School of Public Health, Emory University, Atlanta, USA}}

\affil[6]{\orgdiv{Department of Anesthesiology}, \orgname{Emory University School of Medicine, Atlanta, USA}}

\affil[7]{\orgdiv{Department of Surgery}, \orgname{Emory University School of Medicine, Atlanta, USA}}

\affil[8]{\orgdiv{Department of Medicine}, \orgname{Emory University School of Medicine, Atlanta, USA}}

\affil[9]{\orgdiv{School of Medicine}, \orgname{University of California, San Francisco, USA}}

\abstract{
Time-series Foundation Models (FMs) are pretrained on either heterogeneous corpora or specific signal domains.
Using Photoplethysmography (PPG) as a representative modality, we compare generalist time-series FMs, PPG-specialist FMs, and traditional machine learning (ML) baselines for PPG-based health assessment.
We conduct a two-stage benchmark.
First, we evaluate multiple FM families across 23 downstream evaluations (e.g., atrial fibrillation detection, vital-sign estimation).
Second, we conduct a focused four-FM comparison (MOMENT, Chronos-2, GPT-PPG, MOMENT-PPG) at multiple model sizes on an expanded 51-task suite, under both frozen-backbone and full-model tuning, to study tuning strategy, model scaling, and cross-dataset transfer.
The multi-model benchmark shows that PPG-specialist FMs achieve the strongest aggregate performance, but their advantage is not uniform: generalist FMs remain competitive on several tasks, and traditional ML baselines outperform FMs on selected clinical regression tasks.
The focused analysis shows that the advantage of PPG-specific pretraining appears mainly with frozen backbones and on targets that the waveform directly reflects, such as atrial fibrillation and vital signs. Full-model tuning narrows this advantage, model rankings change with the tuning strategy, and the largest model variant is the best choice in only about one-third of cases.
Additionally, controlled experiments with two generalist models show that initializing PPG pretraining from generalist checkpoints provides an early advantage over training from scratch, even when the pretrained features offer little downstream benefit. 
Together, these findings suggest that PPG-specific pretraining is useful but task-dependent, and that FMs should be evaluated against simpler statistical, traditional ML, and clinical history-based baselines.
}


\maketitle

\input{1_intro}

\input{2_related_work}

\input{3_background}

\input{4_evaluation_methodology}

\input{5_results_multiple_model}

\input{6_results_classification}

\input{7_results_regression}

\input{8_results_others}


\begin{appendices}

\input{appendix_1}

\input{appendix_2}

\end{appendices}

\end{document}

%% file: 1_intro.tex
\section{Introduction}

Photoplethysmography (PPG) is widely used in wearable and clinical sensing.
Beyond basic pulse monitoring, PPG supports a broad range of downstream tasks, including cardiac state assessment, activity recognition, vital-sign estimation, and blood-oxygen monitoring~\cite{pillai2024papagei}.
This diversity motivates the development of Foundation Models (FMs) for PPG, since a reusable representation learner could reduce the need to train a separate task-specific model for every dataset, label type, and clinical application.
FMs learn reusable representations through large-scale pretraining and adapt them to downstream tasks using prediction heads or fine-tuning.
Models such as BERT~\cite{devlin2018bert}, GPT-4~\cite{achiam2023gpt}, Stable Diffusion~\cite{rombach2022high}, and Whisper~\cite{radford2023robust} illustrate their success in language, vision, and speech.
This approach is gaining attention in time-series analysis and physiological sensing, where labels are costly to obtain and data vary across subjects, devices, and clinical settings~\cite{goswami2024moment}.
Recent FMs for PPG~\cite{chen2025gpt}, Electroencephalography (EEG)~\cite{thapa2024sleepfm}, and Electrocardiography (ECG)~\cite{li2024electrocardiogram} illustrate its application to physiological signals.
Time-series language models such as OpenTSLM~\cite{langer2025opentslm} also support reasoning over signals and text; our study focuses on representations for downstream classification and regression.

Physiological FMs are often specialists, pretrained on the same or closely related signal domains as their downstream tasks.
For instance, ECG-BERT~\cite{choi2023ecgbert} and ECG-FM~\cite{mckeen2024ecg} for ECG, and GPT-PPG~\cite{chen2025gpt} and PaPaGei~\cite{pillai2024papagei} for PPG.
In parallel, recent generalist time-series FMs have been pretrained on heterogeneous temporal corpora spanning weather, traffic, electricity, finance, and other domains like MOMENT~\cite{goswami2024moment}, Chronos~\cite{ansari2024chronos}, TimesFM~\cite{das2024timesfm}, and Moirai~\cite{liu2025moirai}.
These models aim to learn broadly transferable temporal representations rather than specializing in a single time-series modality.
This creates a practical question for PPG-based health assessment: when do generalist time-series FMs transfer effectively to PPG, when does PPG-specific pretraining help, and how do both compare with traditional ML baselines trained directly on downstream tasks?
We therefore compare FMs with simpler statistical and clinical history baselines, and assess whether scaling achieves a universally better model.

\begin{figure*}[t]
    \centering
\includegraphics[width=0.99\linewidth]{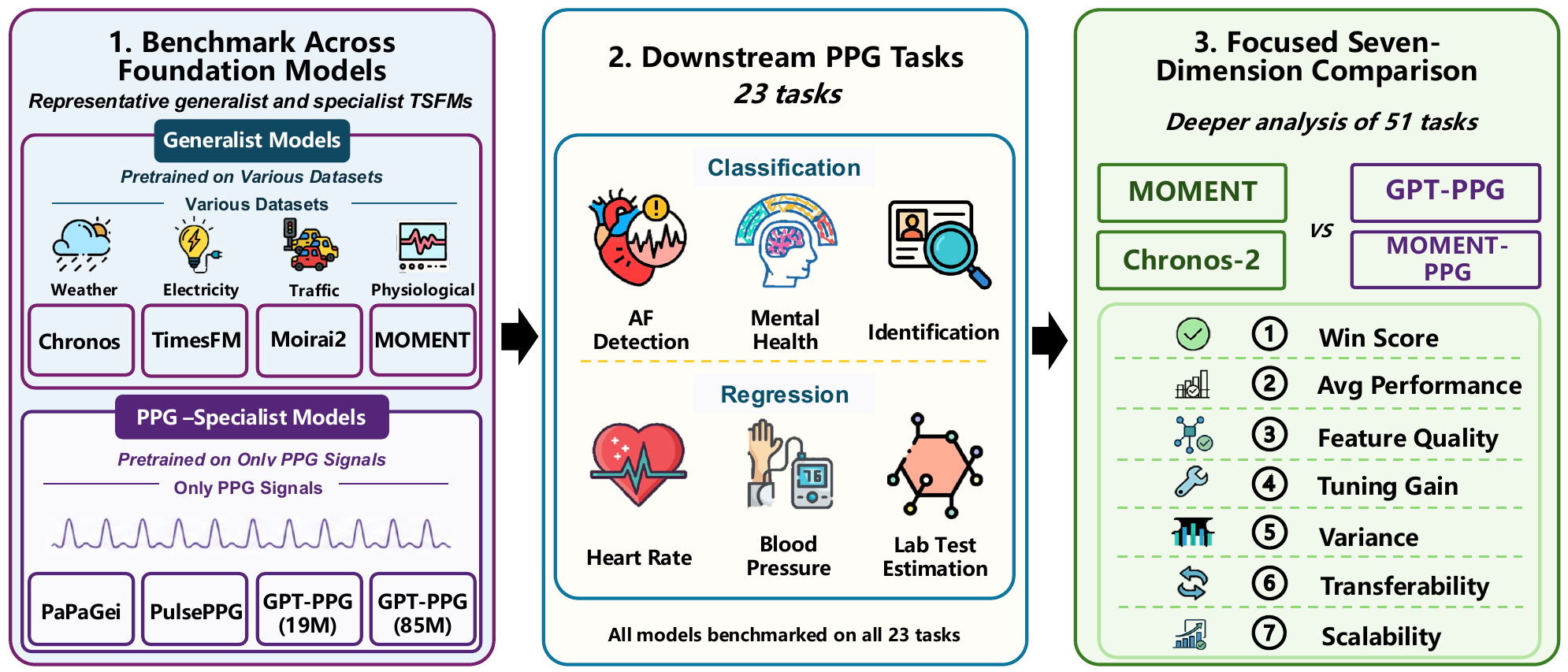}
    \caption{Overview of the two-stage benchmark.
    The first stage compares generalist time-series FMs, PPG-specialist FMs, and traditional ML baselines under a common PPG segment-level prediction protocol.
    The second stage uses MOMENT, Chronos-2, MOMENT-PPG, and GPT-PPG as representative scalable FMs for more detailed analysis.}
    \label{fig:overview}
\end{figure*}

We address these questions through a two-stage PPG-based health-assessment benchmark, followed by detailed analyses (Fig.~\ref{fig:overview}).
In the first stage, we compare multiple generalist time-series FMs, multiple PPG-specialist FMs, and traditional ML baselines across 23 task-level evaluations.
This multi-model comparison is designed to separate model-family trends from the behavior of any single model pair.
We find specialist performance to be the best, followed by generalist and traditional ML models.
As FM performance depends on scale, pretraining data, adaptation data, and the training objective, we also conduct focused benchmarking and analysis.
In the second stage, we perform a focused analysis of four FM families at different model scales across an expanded 51-task suite, adding tasks and datasets for deeper comparison.
We select MOMENT and Chronos-2 as generalists and GPT-PPG and MOMENT-PPG as PPG specialists because these models are available at multiple scales and differ significantly in design, enabling deeper analysis.
Using a holistic evaluation framework, we compare the families along seven dimensions: win score, average performance, feature quality, tuning gain, variance, transferability, and scalability. 
We also contextualize our findings with simpler statistical and clinical history-based baselines and discuss when they surpass FMs.
The performance gap between generalists and specialists narrows under full-model tuning, but both can fall short of simpler baselines in certain cases.

We also conduct controlled experiments on architecture and pretraining.
We compare self-attention (the dominant choice in PPG FMs) with non-attention token mixers under reconstruction and contrastive objectives, and compare PPG pretraining from random weights with continued pretraining from MOMENT and Chronos-2 checkpoints.
Finally, we conduct a controlled pretraining study of two generalists, MOMENT and Chronos-2, that fixes the architecture and varies only where PPG pretraining starts: from random weights or from the generalist checkpoint.
It yields three interesting findings.
First, starting from the generalist checkpoint gives PPG pretraining a head start in downstream performance.
Second, this head start does not require useful features: MOMENT's frozen features are statistically indistinguishable from random weights on PPG tasks, yet it gives a head start of the same order as Chronos-2, so the checkpoint transfers to PPG as a parameter initialization rather than as a representation.
Third, the checkpoint's off-the-shelf performance predicts which tasks gain the most in the early training stages, so we can anticipate the benefit of continued pretraining under a limited data or compute budget without any pretraining.
These analyses provide deeper insights into the benefits and limitations of computationally heavy pretraining.

While we do not introduce another PPG FM, we aim to provide a systematic benchmark and analysis framework for understanding when different classes of time-series FMs succeed or fail on PPG signals.
This paper makes the following contributions:
\begin{itemize}
    \item \textbf{A multi-family benchmark for PPG health assessment.}
    We compare multiple generalist and PPG-specialist FMs with traditional ML baselines across 23 task-level evaluations using common task protocols, validation-based baseline and checkpoint selection, and task-level uncertainty estimates.
    Additional statistical and clinical-history baseline comparisons establish whether FMs improve on simpler alternatives in the evaluated settings.

    \item \textbf{Task-dependent model selection and adaptation.}
    We evaluate MOMENT, Chronos-2, GPT-PPG, and MOMENT-PPG across multiple model sizes on an expanded 51-task suite under head-only and full-model tuning.
    A seven-dimensional analysis characterizes their strengths and limitations, showing that model selection should depend on the target, adaptation setting, and cross-dataset transfer rather than on pretraining domain or parameter count alone.

    \item \textbf{Controlled analysis of architecture--objective interactions.}
    We replace self-attention with convolutional, recurrent, and state-space alternatives in a common Transformer backbone, with matched model sizes and training settings.
    All seven alternatives learn better frozen representations overall under masked reconstruction, and three tested alternatives also outperform self-attention under next-patch prediction.
    Their advantages narrow under contrastive learning, and low reconstruction loss does not guarantee useful downstream features.
    
    \item \textbf{Budget-dependent analysis of generalist initialization.}
    We compare continued PPG pretraining from MOMENT and Chronos-2 checkpoints with PPG pretraining from random initialization.
    Both checkpoints improve downstream performance early in pretraining, but this advantage diminishes as training continues.
    The checkpoint's off-the-shelf probe performance indicates which tasks benefit most under a limited pretraining budget.
\end{itemize}

%% file: 2_related_work.tex
\section{Related Work}
\label{sec:related_work}
FMs have been developed for language and vision, including Llama~\cite{touvron2023llama}, CLIP~\cite{radford2021learning}, and SAM 2~\cite{ravi2024sam}, and for biomedical text and health records, including BioBERT~\cite{lee2020biobert}, PubMedBERT~\cite{gu2021domain}, and MED-BERT~\cite{rasmy2021med}.
These models use large-scale pretraining and subsequent downstream adaptation, motivating similar approaches for physiological signals.
Physiological FMs include LIFT-PD, which uses masked prediction for accelerometry~\cite{soumma2024wearable}, ECG-BERT~\cite{choi2023ecgbert} and ECG-FM~\cite{mckeen2024ecg}, which use masked modeling and contrastive learning for ECG representations, and FM fusion for acute coronary syndrome detection~\cite{meng2025fusion}.
EEG approaches include autoregressive Neuro-GPT~\cite{cui2024neuro} and EEG-GPT, which prompts language models with EEG statistics for abnormality assessment~\cite{kim2024eeg}; HeartBEIT applies a vision transformer to two-dimensional ECG inputs~\cite{vaid2023foundational}.

General time-series representation methods include TS2Vec~\cite{yue2022ts2vec}, PatchTST~\cite{nie2022time}, and multivariate representation learning~\cite{franceschi2019unsupervised}.
MOMENT~\cite{goswami2024moment} extends this direction through masked pretraining on a heterogeneous time series dataset.
For PPG, GPT-PPG uses autoregressive next-patch prediction~\cite{chen2025gpt} and has been evaluated on clinical tasks~\cite{panchumarthi2025fairtune,kataria2025continuous,kataria2025wav2arrest}.
PaPaGei uses morphology-aware contrastive learning on public PPG datasets~\cite{pillai2024papagei}, while Pulse-PPG studies pretraining on field recordings~\cite{saha2025pulse}.
Distillation has been explored for PaPaGei and GPT-PPG deployment~\cite{ni2025ppg}, while AnyPPG~\cite{nie2025anyppg} and Cardiac Sensing Foundation Model (CSFM)~\cite{gu2026cardiac} explore multimodal pretraining and ECG--PPG alignment.
Additionally, related work includes wearable biosignal pretraining with NormWear~\cite{luo2024toward} and Apple PPG/ECG-FM~\cite{abbaspourazad2023large}, ECG benchmarks covering interpretation labels, reduced LVEF, and abnormal troponin~\cite{wan2025openecg}, and vision-language models for echocardiography~\cite{christensen2024vision}.

Our benchmark expands the task coverage of the PPG FM evaluation.
In addition to commonly studied PPG tasks such as atrial fibrillation detection, heart-rate estimation, blood-pressure estimation, and SpO$_2$ estimation, we evaluate less-explored tasks, including biometric identification, non-invasive cholesterol estimation~\cite{arguello2025non}, blood electrolyte and biomarker estimation, ECG-metric estimation, and Echocardiogram (ECHO)-metric estimation.
Some related targets, such as elevated troponin and potassium levels, have been studied using ECG FMs~\cite{mckeen2024ecg}, and multimodal models can estimate auxiliary physiological variables, such as skin temperature.
However, a unified PPG benchmark covering generalist FMs, PPG-specialist FMs, and traditional ML baselines across this task range has not been established.
In prior work, we demonstrate continuous laboratory-test estimation from PPG within the UNIPHY+ framework~\cite{wang2025unified,wang2025estimating}.
Here, we focus instead on short-window PPG prediction, using 30-second or task-specific short segments to evaluate task-level transfer, model-family differences, and the limits of FMs relative to traditional ML, statistical, and clinical history baselines.
Additional related benchmark and deployment considerations are discussed in Section~\ref{sec:discussion}.

%% file: 3_background.tex
\section{FMs and baselines}
\label{sec:models}

\textbf{FMs.}
We benchmark three groups of models: generalist time-series FMs, PPG-specialist FMs, and traditional ML baselines.
For each downstream task, all evaluated FMs use the same task-specific prediction-head architecture.
In the second stage, the focused four-family analysis of MOMENT, Chronos-2, GPT-PPG, and MOMENT-PPG further examines both head-only tuning and full-model tuning settings across each family's available model sizes.
Table~\ref{tab:models} summarizes the eight FMs, including their pretraining corpora, architectures, and self-supervised objectives.
Appendix~\ref{sec:foundation_model_intro} provides full descriptions of each model, including architecture and pretraining details.
 
\begin{table*}[t]
\small
\centering
\caption{FMs evaluated in this benchmark. Generalist FMs are pretrained on heterogeneous time-series corpora that are not specific to PPG; specialist FMs are pretrained on PPG data. Full details are given in the Appendix.}
\label{tab:models}
\resizebox{\textwidth}{!}{%
\begin{tabular}{|l|l|l|l|}
\hline
\textbf{Model} & \textbf{Pretraining corpus} & \textbf{Architecture} & \textbf{Objective} \\
\hline
\multicolumn{4}{|l|}{\textit{Generalist time-series FMs}} \\
\hline
Chronos-2  & Public + synthetic time series     & T5-family transformer    & Multi-step quantile forecasting \\
\hline
TimesFM  & $\sim$100B web/traffic time points & Decoder-only transformer & Next-patch prediction \\
\hline
Moirai-2 & Tens of millions of series         & Decoder-only transformer & Quantile multi-token prediction \\
\hline
MOMENT   & Time Series Pile, multi-domain     & Transformer encoder      & Masked time-series modeling \\
\hline
\multicolumn{4}{|l|}{\textit{PPG-specialist FMs}} \\
\hline
PaPaGei   & $\sim$57k h public PPG             & 1D convolutional encoder & Morphology-aware contrastive learning \\
\hline
PulsePPG & Field PPG, 100-day study           & 1D ResNet encoder      & Contrastive self-supervision \\
\hline
GPT-PPG-19M & $>$100M hospital PPG segments     & GPT-style transformer    & Autoregressive next-patch prediction \\
\hline
GPT-PPG-85M & $>$100M hospital PPG segments     & GPT-style transformer    & Autoregressive next-patch prediction \\
\hline
\end{tabular}
}
\end{table*}

\noindent
\textbf{Traditional ML baselines.}
We include a set of traditional ML baselines trained from scratch on each downstream task: logistic regression~\cite{cox1958regression}, ridge regression~\cite{hoerl1970ridge}, random forests~\cite{breiman2001random}, gradient-boosted trees including XGBoost~\cite{chen2016xgboost} and LightGBM~\cite{ke2017lightgbm}, TabPFN~\cite{hollmann2025tabpfn}, a one-dimensional convolutional network (CNN1D)~\cite{kiranyaz2021one}, and a long short-term memory network (LSTM)~\cite{hochreiter1997long}.
The linear, tree-based, and TabPFN baselines are trained on handcrafted statistical and morphological features extracted from each PPG window, whereas CNN1D and LSTM are trained directly on the raw normalized waveform.
For each task, we report the strongest traditional ML baseline from this pool.
These handcrafted features comprise 37 descriptors per PPG window:
13 time-domain statistics (e.g., mean and standard deviation),
11 pulse morphology features (e.g., mean inter-beat interval and mean pulse width),
8 spectral features (e.g., dominant frequency and spectral entropy),
and 5 autocorrelation and complexity features (e.g., lag-1 autocorrelation and Hjorth mobility).
The complete feature list and extraction details are provided in Appendix~\ref{app:features}.
Within each baseline, early stopping and checkpoint selection use the validation
split. For each task, we then report the strongest of the eight baselines, also
selected on validation and evaluated on test, matching the validation-based
checkpoint selection used for the foundation models.

%% file: 4_evaluation_methodology.tex
\section{Evaluation Methodology}

\subsection{Datasets \& Data Preprocessing}

We evaluate the models on a collection of PPG-based health assessment tasks spanning classification and regression.
The classification tasks include atrial fibrillation detection, mental workload assessment, activity recognition, and biometric identification.
The regression tasks include vital-sign estimation (HR, BP, RR, Skin temperature, Blood Oxygen), blood-biomarker estimation, ECG-metric estimation, ECHO-metric estimation, and cholesterol estimation.
Appendix Table~\ref {tab:data_overview} summarizes the datasets and downstream tasks used in the full benchmark, with detailed dataset descriptions in Appendix~\ref{sec:dataset_details}.

We use two evaluation scopes.
The multi-family benchmark compares generalist FMs, PPG-specialist FMs, and traditional ML baselines across 23 task-level evaluations derived from 20 task--dataset settings.
The focused analysis of MOMENT, Chronos-2, GPT-PPG, and MOMENT-PPG uses a broader 51-task suite, consisting of 6 classification task-level evaluations and 45 regression task-level evaluations, to study model scale, tuning strategy, feature quality, variance, transferability, and scalability.
This separation reflects model availability and computational scope: the multi-family benchmark prioritizes a common task subset that can be run across all evaluated FMs and traditional ML baselines, whereas the focused analysis expands task coverage for four model families with multiple available size variants evaluated under common head-only and full-model adaptation protocols.
At the dataset level, the full benchmark covers 19 unique task types and 21 datasets.

Each input sample is formulated as a segment-level prediction problem.
The input is a short PPG window, and the target is either a categorical label or a scalar physiological or clinical value.
For vital-sign labels, the target is temporally aligned with the PPG window.
For laboratory-test labels, we select PPG windows before the laboratory measurement time and use the corresponding laboratory value as the scalar target.
We apply min--max normalization to each segment, scaling the data to the range $[0,1]$.
Given an original PPG segment $X$, the normalized segment $X_n$ is
$X_n = \frac{X - \min(X)}{\max(X)-\min(X)}$.
Details of the train, validation, and test splits are provided in Appendix~\ref{subsec:train_test_split}.

\subsection{Fine-tuning Strategies}

We use a common downstream adaptation setup for all evaluated FMs. For a given task, the same classification or regression head architecture is used across models, and every FM is evaluated under two fine-tuning strategies. In head-only tuning, the FM backbone is frozen and only the prediction head is trained. This setting evaluates the transferability of pretrained representations under limited adaptation. In full-model tuning, both the FM backbone and the prediction head are updated. This setting evaluates whether task-specific adaptation improves performance beyond frozen-feature transfer. Results are compared across model families separately within each tuning setting. The focused four-family analysis additionally examines these strategies across each family's available model sizes.


\subsection{Evaluation Metrics}
\label{sec:metrics}

For classification tasks, we report the F1 score or accuracy, depending on the task.
For regression tasks, we report mean absolute error (MAE).
For the main multi-family benchmark, we report the mean performance along with uncertainty estimates and compute a \emph{win score}, defined as the number of task-level evaluations in which a model achieves the best mean result.
Unless otherwise specified, uncertainty is reported as the task-level standard deviation from the repeated-run, fold, or split procedure used for that dataset; for the frozen-encoder nearest-neighbor baseline, we additionally report bootstrap-based standard errors and confidence intervals.
We also aggregate win scores within model families to compare generalist FMs, PPG-specialist FMs, and traditional ML baselines.

For the focused comparison of MOMENT, Chronos-2, GPT-PPG, and MOMENT-PPG, we additionally use a seven-dimensional evaluation framework:
(1)~\emph{win score};
(2)~\emph{average performance};
(3)~\emph{feature quality}, measured by performance under head-only tuning with a frozen backbone;
(4)~\emph{tuning gain}, measured as the relative improvement of full-model tuning over head-only tuning;
(5)~\emph{variance}, measured as dispersion across tasks, model sizes, and tuning strategies;
(6)~\emph{transferability}, measured by performance on unseen datasets; and
(7)~\emph{scalability}, measured as the slope of performance against log model size.
To keep the win-score axis consistent between classification and regression, the focused seven-dimensional tables compute win score from full-model tuning for both task types; head-only tuning is used for the feature-quality and tuning-gain axes and is reported in the detailed appendix tables.
The focused seven-dimensional summaries are aggregate point estimates
derived from the task-level tables, and we do not estimate separate
confidence intervals for these aggregate axes.
For regression, cross-task summaries use task-normalized log errors rather than directly averaging MAEs in different units.
In the regression radar plot, lower-is-better metrics are sign-reversed, and each axis is min--max normalized across the four model families so that larger plotted values indicate better scores.
Formal definitions are provided in Appendix~\ref{sec:metrics_details}.

%% file: 5_results_multiple_model.tex
\section{Performance of Multiple Model Families on Downstream PPG Tasks}
\label{sec:multi_family_performance}

We compare multiple model families on downstream PPG tasks,
including generalist time-series FMs, PPG-specialist FMs, and
traditional ML baselines.
This multi-family benchmark contains 23 task-level evaluations
derived from 20 task--dataset settings.
All FMs are evaluated under full-model tuning with the same
task-specific prediction-head architecture.
For MOMENT and Chronos-2, we use the medium-size variants
(MOMENT-125M and Chronos-2-120M).
Table~\ref{tab:main_representative_tasks} reports representative
task-level results across classification, mental workload,
activity recognition, vital-sign estimation, blood-pressure
estimation, and laboratory-test prediction.
The full task-level comparison, including the selected traditional ML
baseline for each task, is provided in
Appendix~\ref{app:detailed_multi_family_results}.

Overall, PPG-specialist FMs achieve the highest aggregate win count,
with PulsePPG obtaining the largest number of
task-level wins and GPT-PPG-19M remaining competitive on AF
detection and selected clinical tasks.
Among the generalist FMs, Chronos-2 obtains the highest overall
task-level win count, while MOMENT remains strong on selected
classification and clinical tasks.
However, the advantage of FMs is not uniform, and the
margin between specialist and generalist model families is modest.
Traditional ML baselines remain competitive and win selected clinical
or physiological regression tasks, including respiration-rate
estimation, systolic blood-pressure estimation on PPG-BP, and A1c
estimation.
This pattern suggests that PPG-specific pretraining is useful for
many PPG-centered tasks, but traditional ML remains a strong baseline
on selected clinical and physiological tasks.
The representative tasks illustrate this heterogeneity.
For AF detection, PulsePPG achieves the highest mean F1 on Stanford,
whereas Chronos-2 achieves the highest mean F1 on Simband.
For workload and activity tasks, PulsePPG achieves the best MAUS
mental-workload F1 score, whereas MOMENT has the best Pulse Transit
activity-recognition accuracy.
Regression tasks show an even more mixed pattern:
PulsePPG is strongest on DALIA heart-rate estimation and troponin
estimation, Chronos-2 is strongest on WESAD heart-rate estimation,
and traditional ML baselines win representative respiration rate,
systolic blood pressure, and A1c tasks.
These examples show why the benchmark is best interpreted as a
multi-family comparison rather than as a single ranking of all models.

\begin{table}[!t]
\small
\centering
\resizebox{\textwidth}{!}{%
\begin{tabular}{|l|l|c|l|l|l|}
\hline
\textbf{Task} & \textbf{Dataset} & \textbf{Metric} & \textbf{Best generalist FM} & \textbf{Best PPG-specialist FM} & \textbf{Best traditional ML} \\
\hline
AF Detection & Stanford & F1 (\%) $\uparrow$ & Chronos-2: 87.77 $\pm$ 0.85 & \textbf{PulsePPG: 93.93 $\pm$ 0.43} & CNN1D: 86.49 $\pm$ 0.90 \\
\hline
AF Detection & Simband & F1 (\%) $\uparrow$ & \textbf{Chronos-2: 69.66 $\pm$ 3.78} & GPT-PPG-19M: 67.91 $\pm$ 3.84 & LogReg: 68.17 $\pm$ 3.56 \\
\hline
Mental Workload & MAUS & F1 (\%) $\uparrow$ & Chronos-2: 68.54 $\pm$ 2.00 & \textbf{PulsePPG: 72.75 $\pm$ 2.13} & LSTM: 67.20 $\pm$ 2.26 \\
\hline
Activity Recognition & Pulse Transit & Acc. (\%) $\uparrow$ & \textbf{MOMENT: 87.96 $\pm$ 1.06} & PulsePPG: 87.46 $\pm$ 0.94 & CNN1D: 84.16 $\pm$ 0.99 \\
\hline
Heart Rate Estimation & WESAD & MAE $\downarrow$ & \textbf{Chronos-2: 9.60 $\pm$ 0.07} & PulsePPG: 9.93 $\pm$ 0.08 & CNN1D: 9.94 $\pm$ 0.07 \\
\hline
Heart Rate Estimation & DALIA & MAE $\downarrow$ & Chronos-2: 9.03 $\pm$ 0.10 & \textbf{PulsePPG: 5.91 $\pm$ 0.09} & XGBoost: 24.25 $\pm$ 0.13 \\
\hline
Respiration Rate Estimation & MIMIC-III & MAE $\downarrow$ & Chronos-2: 4.17 $\pm$ 0.04 & PulsePPG: 4.24 $\pm$ 0.04 & \textbf{CNN1D: 4.15 $\pm$ 0.04} \\
\hline
Systolic BP Estimation & PPG-BP & MAE $\downarrow$ & Chronos-2: 15.36 $\pm$ 1.16 & GPT-PPG-19M: 15.31 $\pm$ 1.28 & \textbf{RidgeReg: 15.27 $\pm$ 1.21} \\
\hline
A1c Estimation & Inst. A & MAE $\downarrow$ & TimesFM / MOMENT: 1.35 $\pm$ 0.03 & GPT-PPG-85M: 1.33 $\pm$ 0.03 & \textbf{LSTM: 1.31 $\pm$ 0.03} \\
\hline
Troponin Estimation & MIMIC-III & MAE $\downarrow$ & TimesFM: 1.16 $\pm$ 0.02 & \textbf{PulsePPG: 1.11 $\pm$ 0.02} & LSTM: 1.15 $\pm$ 0.02 \\
\hline
\end{tabular}%
}
\caption{Representative task-level results across model families.
For F1 score and accuracy, higher is better; for MAE, lower is better.
Bold indicates the best result among the three displayed model-family
representatives. All FM results use full-model tuning.
The complete task-level results are provided in
Appendix~\ref{app:detailed_multi_family_results}.}
\label{tab:main_representative_tasks}
\end{table}

Table~\ref{tab:main_win_summary} summarizes the task-level win
counts across individual models and the traditional ML baseline pool.
The benchmark comprises 20 task--dataset settings; because each of
the three blood-pressure datasets contributes a systolic and a
diastolic task, this yields 23 task-level evaluations in total.
Each task-level evaluation is credited to the model with the best
mean score.
The traditional ML baseline pool receives 5.5 of the 23 task-level
win credits, demonstrating its competitiveness on selected tasks.
PPG-specialist FMs obtain the highest aggregate win score (10),
compared with 7.5 for generalist FMs, with PulsePPG achieving the
highest individual win score (8).
Sections~\ref{sec:classification}--\ref{sec:further_analysis}
examine these patterns through the focused four-family comparison,
baseline and transfer analyses, and controlled pretraining experiments.

\begin{table}[!t]
\small
\centering
\resizebox{\textwidth}{!}{%
\begin{tabular}{|l|l|c|l|}
\hline
\textbf{Model family} & \textbf{Model / baseline} & \textbf{Win score} & \textbf{Main pattern} \\
\hline
Generalist FM & Chronos-2 & 3.5/23 & Highest generalist win count; competitive on several regression tasks \\
\hline
Generalist FM & TimesFM & 0/23 & Competitive on selected tasks but no task-level win \\
\hline
Generalist FM & Moirai-2 & 1/23 & One task-level win; otherwise mixed performance \\
\hline
Generalist FM & MOMENT & 3/23 & Strong on selected classification and clinical tasks \\
\hline
PPG-specialist FM & PaPaGei & 0.5/23 & Shares the best SpO$_2$ result on MIMIC-III; otherwise no task-level win \\
\hline
PPG-specialist FM & PulsePPG & 8/23 & Highest overall win count, especially on PPG-centered tasks \\
\hline
PPG-specialist FM & GPT-PPG-19M & 1/23 & Competitive on AF detection; wins temperature estimation on MIMIC-III \\
\hline
PPG-specialist FM & GPT-PPG-85M & 0.5/23 & Larger scale does not consistently improve task-level performance \\
\hline
Traditional ML baselines & Best traditional ML baseline & 5.5/23 & Wins selected regression tasks \\
\hline
\end{tabular}%
}
\caption{Task-level win summary across FMs and
traditional ML baselines. Each of the 23 task-level evaluations is
credited to the model with the best mean score.
Each task-level evaluation contributes one win credit, split equally
among models with the best reported mean score.
Complete task-level results, including the selected traditional ML
baselines, are provided in
Appendix~\ref{app:detailed_multi_family_results}.}
\label{tab:main_win_summary}
\end{table}

%% file: 6_results_classification.tex
\section{Focused FM comparison}
To test how performance varies with model scale, we evaluate
each model family across its available size variants.
MOMENT, GPT-PPG, and MOMENT-PPG each have three variants,
whereas Chronos-2 has two.
Scaling trends are assessed within each family over its
available parameter range.
Due to the scarce availability of open PPG FMs, we train MOMENT-PPG from scratch on multiple scales (40M, 125M, 385M) by using the MOMENT pretraining recipe and UCSF/Inst.A dataset that was used to pretrain GPT-PPG.

\subsection{Performance on Classification Tasks}
\label{sec:classification}

\begin{figure}[htbp]
  \centering
  \captionsetup{position=bottom,skip=6pt}

  \begin{minipage}[t]{0.51\textwidth}
    \vspace{0pt}
    \begin{minipage}[c][8cm][c]{\linewidth}
      \centering
      \includegraphics[width=\linewidth]{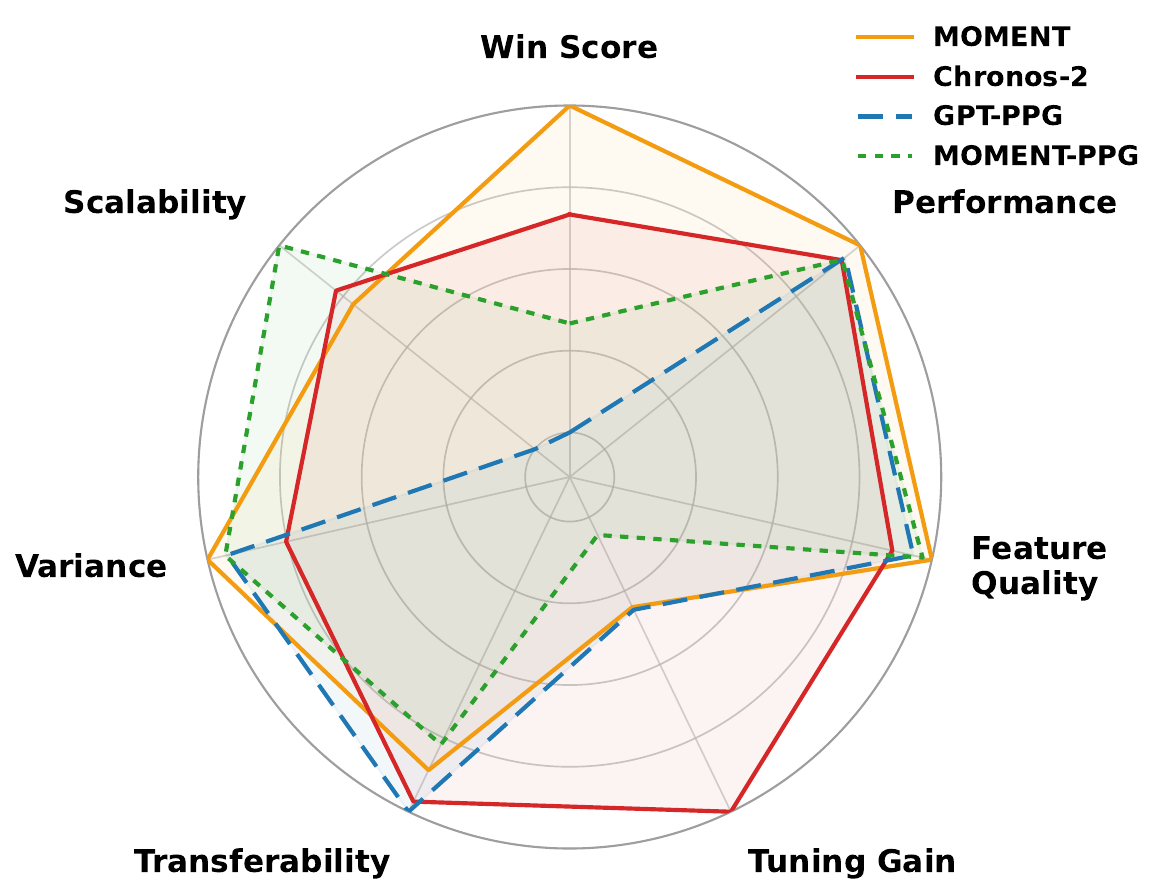}
    \end{minipage}
    \caption{Seven-dimensional comparison on classification tasks.
    Metrics are normalized so that larger radar area indicates
    better performance across axes.}
    \label{fig:cls_7dim}
  \end{minipage}%
  \hfill
  \begin{minipage}[t]{0.45\textwidth}
    \vspace{0pt}
    \begin{minipage}[c][8cm][c]{\linewidth}
      \centering
      \resizebox{\linewidth}{!}{%
        \begin{tabular}{lcc}
          \toprule
          \textbf{Metric}
            & \textbf{MOMENT}
            & \textbf{Chronos-2} \\
          \midrule
          Win Score ($\uparrow$)
            & \textbf{3} & 2 \\
          Performance ($\uparrow$)
            & \textbf{74.03\%} & 68.67\% \\
          Feature Quality ($\uparrow$)
            & \textbf{70.37\%} & 61.63\% \\
          Tuning Gain ($\uparrow$)
            & 11.17 & \textbf{36.57} \\
          Variance ($\downarrow$)
            & \textbf{15.25} & 20.24 \\
          Transferability ($\uparrow$)
            & 57.46\% & 64.60\% \\
          Scalability ($\uparrow$)
            & 0.47 & 0.83 \\
          \midrule
          \textbf{Metric}
            & \textbf{GPT-PPG}
            & \textbf{MOMENT-PPG} \\
          \midrule
          Win Score ($\uparrow$)
            & 0 & 1 \\
          Performance ($\uparrow$)
            & 69.68\% & 68.83\% \\
          Feature Quality ($\uparrow$)
            & 66.27\% & 68.31\% \\
          Tuning Gain ($\uparrow$)
            & 11.50 & 2.21 \\
          Variance ($\downarrow$)
            & 16.25 & 16.15 \\
          Transferability ($\uparrow$)
            & \textbf{66.90\%} & 51.65\% \\
          Scalability ($\uparrow$)
            & -3.34 & \textbf{2.02} \\
          \bottomrule
        \end{tabular}%
      }
    \end{minipage}
    \captionof{table}{Seven-dimensional metric summary for
    classification tasks.}
    \label{tab:classification_comparison}
  \end{minipage}
\end{figure}

We next compare two generalist model families, MOMENT and
Chronos-2, with two PPG-specialist families, GPT-PPG and
MOMENT-PPG, on classification tasks.
This analysis covers atrial fibrillation detection, mental
workload and stress assessment, activity recognition, and
human identification across six task--dataset evaluations.
We evaluate both head-only tuning and full-model tuning
across all 11 available model configurations.
Detailed per-task results are provided in
Appendix~\ref{app:classification_details}; here, we summarize
the comparison through the seven-dimensional analysis in
Figure~\ref{fig:cls_7dim} and
Table~\ref{tab:classification_comparison}.
As defined in Section~\ref{sec:metrics}, the win-score axis
uses full-model tuning.
Wins are determined jointly across all 11 configurations,
with tied wins split equally, and summed across sizes
within each model family.
These tasks probe different kinds of information in the
PPG signal: rhythm irregularity, physiological changes
associated with workload and stress, motion and activity
patterns, and identity-specific physiology.
The classification setting therefore tests both label
separability and the reuse of learned representations
across diverse PPG classification tasks.

As shown in Fig.~\ref{fig:cls_7dim} and
Table~\ref{tab:classification_comparison}, MOMENT leads in
full-model classification win score (3/6), aggregate
classification performance (74.03\%), and frozen-feature quality
(70.37\%), while also achieving the lowest variance score.
The other model families exhibit complementary strengths:
Chronos-2 achieves the largest tuning gain (36.57),
GPT-PPG has the highest transferability score (66.90\%),
and MOMENT-PPG achieves the highest scalability score (2.02).
The detailed results in Appendix~\ref{app:classification_details}
show that task-level winners differ across model families.
Based on the reported means under full-model tuning, MOMENT
wins Mental Stress PPG, Pulse Transit, and Real-world PPG.
Chronos-2 wins Simband AF detection and MAUS, with the 120M
variant achieving F1 scores of 69.66\% and 68.54\%, respectively.
MOMENT-PPG wins Stanford AF detection, reaching 91.02\% F1
with its 385M variant.
GPT-PPG does not win any of the six evaluations under
full-model tuning.

The head-only results show a different distribution of wins.
GPT-PPG wins both AF detection datasets, achieving 85.83\% F1
on Stanford and 64.89\% F1 on Simband with its 85M variant,
while MOMENT wins the remaining four evaluations.
Chronos-2 benefits particularly strongly from full-model
adaptation: its 120M variant improves from 30.61\% to 87.98\%
accuracy on human identification.
These results show that frozen-feature quality and performance
after full-model adaptation capture different strengths.
Overall, classification advantages depend on the task,
tuning protocol, and evaluation metric.
This four-family analysis complements the broader benchmark
in Section~\ref{sec:multi_family_performance}, which includes
additional FM families and traditional ML baselines.
The main findings are:

\begin{itemize}
    \item The effect of pretraining domain is clearest when the backbone is frozen. With a frozen backbone, both PPG-specialist families rank above both generalist families on the two AF datasets, while MOMENT performs best on all four non-AF tasks. After full-model tuning, the gap between the best and worst family narrows on five of the six tasks. The best specialist and the best generalist also fall within bootstrap uncertainty on both AF datasets. Evaluating frozen features alone therefore overstates how much pretraining domain matters for classification.

    \item Rankings under frozen and fully tuned backbones disagree. Chronos-2 has the weakest frozen features but the largest tuning gain, and it wins two tasks after full tuning. MOMENT-PPG has the second-strongest frozen features but ranks last on activity recognition and human identification after full tuning. A model chosen by frozen-feature quality may not be the best choice when full tuning is planned, so both settings should be reported.

    \item In-domain accuracy does not predict cross-dataset transfer. MOMENT has the highest in-domain classification performance but ranks third in transfer. GPT-PPG and Chronos-2 transfer best despite lower in-domain performance.
    
    \item Larger models are rarely the best choice. For the three families with three sizes, the largest variant gives the best full-tuning result in only 5 of 18 family-task pairs. That is about what choosing a size at random would give. The best size also changes from task to task within the same family, so model size should be selected per task rather than maximized.
\end{itemize}

%% file: 7_results_regression.tex
\subsection{Performance on Regression Tasks}
\label{sec:regression}

We next examine the four model families on regression tasks.
The analysis covers 14 task types and 45 task-level evaluations,
counting systolic/diastolic blood pressure and HDL/LDL cholesterol
separately.
Detailed results, scaling trends, and data-regime analyses are
provided in Appendix~\ref{app:regression_details};
Fig.~\ref{fig:regression_radar} and
Table~\ref{tab:regression_comparison} summarize the
seven-dimensional profiles.
These tasks span vital signs, blood pressure, laboratory
biomarkers, ECG-derived and echocardiographic metrics, and
cholesterol-related measures.
The targets differ in how directly they are reflected in the
PPG waveform, allowing us to examine whether the advantages
of PPG-specialist pretraining extend across physiological
targets and persist under adaptation and cross-dataset transfer. As shown in Fig.~\ref{fig:regression_radar} and
Table~\ref{tab:regression_comparison}, Chronos-2 achieves the
lowest task-normalized regression composite, while MOMENT-PPG
leads in full-model win score and frozen-feature quality.
MOMENT-PPG also achieves the lowest normalized cross-dataset
transfer error.
The coexistence of a strong generalist and a strong specialist
suggests that pretraining domain alone does not explain the
regression ranking.
In particular, MOMENT-PPG's transfer performance shows that
GPT-PPG's weaker transfer should not be interpreted as an
inherent limitation of PPG-specialist representations.

\begin{figure}[htbp]
  \centering
  \captionsetup{position=bottom,skip=6pt}
  \begin{minipage}[t]{0.48\textwidth}
    \vspace{0pt}
    \begin{minipage}[c][8cm][c]{\linewidth}
      \centering
      \includegraphics[width=\linewidth]{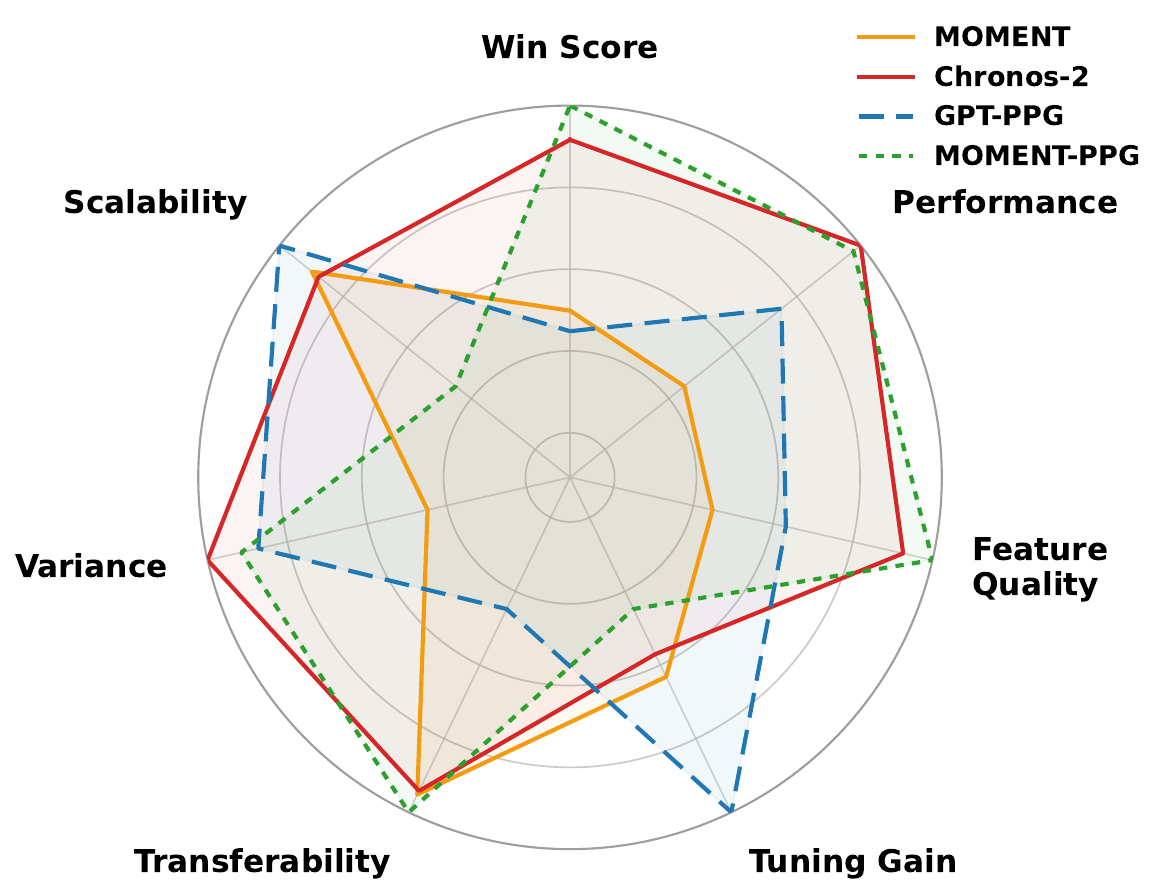}
    \end{minipage}
    \caption{Seven-dimensional comparison on regression tasks.
    MAE-based metrics are inverted and normalized in the radar plot
    so that larger area indicates better performance across axes.}
    \label{fig:regression_radar}
  \end{minipage}%
  \hfill
  \begin{minipage}[t]{0.48\textwidth}
    \vspace{0pt}
    \begin{minipage}[c][8cm][c]{\linewidth}
      \centering
      \resizebox{\linewidth}{!}{%
\begin{tabular}{lcc}
  \toprule
  \textbf{Metric}
    & \textbf{MOMENT}
    & \textbf{Chronos-2} \\
  \midrule
  Win Score ($\uparrow$)
    & 8.00 & 14.25 \\
  Performance ($\downarrow$)
    & 0.29 & \textbf{-0.20} \\
  Feature Quality ($\downarrow$)
    & 0.53 & -0.12 \\
  Tuning Gain ($\uparrow$)
    & 0.05 & 0.04 \\
  Variance ($\downarrow$)
    & 1.06 & \textbf{0.93} \\
  Transferability ($\downarrow$)
    & -0.36 & -0.33 \\
  Scalability ($\uparrow$)
    & 0.04 & 0.03 \\
  \midrule
  \textbf{Metric}
    & \textbf{GPT-PPG}
    & \textbf{MOMENT-PPG} \\
  \midrule
  Win Score ($\uparrow$)
    & 7.25 & \textbf{15.50} \\
  Performance ($\downarrow$)
    & 0.02 & -0.18 \\
  Feature Quality ($\downarrow$)
    & 0.28 & \textbf{-0.22} \\
  Tuning Gain ($\uparrow$)
    & \textbf{0.11} & 0.02 \\
  Variance ($\downarrow$)
    & 0.96 & 0.95 \\
  Transferability ($\downarrow$)
    & 1.20 & \textbf{-0.51} \\
  Scalability ($\uparrow$)
    & \textbf{0.09} & -0.18 \\
  \bottomrule
\end{tabular}%
      }
    \end{minipage}
    \captionof{table}{Seven-dimensional metric summary for regression
tasks. Performance, frozen-backbone Feature Quality, and
cross-dataset Transferability report task-normalized log errors.}
\label{tab:regression_comparison}
  \end{minipage}
\end{figure}

The detailed results in Appendix~\ref{app:regression_details}
make the target dependence of specialization more concrete.
In the matched 40M MOMENT--MOMENT-PPG comparison under full-model
tuning, MOMENT-PPG reduces MAE on PR interval, QRS duration,
QT interval, and LVEF, but increases MAE on A1c and both
cholesterol targets.
Thus, matching the PPG input domain does not ensure that a
representation becomes more useful for every physiological
target.
The benefits of specialization must be assessed at the target
level, including for clinical quantities whose relationship
to the waveform is indirect.
Adaptation and scaling reveal another distinction.
GPT-PPG achieves the largest tuning gain and strongest
scalability score, yet still trails Chronos-2 and MOMENT-PPG
in the task-normalized regression composite and normalized
cross-dataset transfer score.
A larger improvement from tuning or additional parameters
therefore does not necessarily produce a stronger final predictor.
Conversely, MOMENT-PPG combines strong frozen-feature quality
with a small tuning gain and a negative scalability score.
Its regression strengths consequently do not translate into
a general recommendation to unfreeze the backbone or select
the largest variant.
These comparisons should also be interpreted alongside
Section~\ref{sec:further_analysis}.
Lower MAE than another FM establishes a relative advantage,
but does not by itself demonstrate practical clinical utility,
particularly where traditional ML or history-based baselines
remain competitive.
The main findings are:

\begin{itemize}
    \item PPG-specialist pretraining helps most when the waveform directly reflects the target. With a frozen backbone, a specialist gives the best result on 8 of 11 vital-sign targets (SpO2, heart rate, respiration rate). On blood-chemistry targets, a specialist is best only about a quarter of the time; the best result is spread across all four families, and margins are small. The gap between the two target groups persists after full tuning, though it is smaller. This is consistent with Section 7.3, which finds that blood-chemistry values are not linearly recoverable from PPG. As with AF detection in classification, specialist pretraining helps where the PPG waveform carries the target signal.

    \item Full tuning narrows the specialist advantage, as in classification. With a frozen backbone, the specialist families win more regression evaluations than the generalists, led by MOMENT-PPG. After full tuning, the two classes are roughly even. Chronos-2 gains the most wins, and its 120M variant becomes the best single configuration, while MOMENT-PPG loses the most. This repeats the classification pattern, where Chronos-2 also benefited most from full tuning and MOMENT-PPG least, so regression rankings likewise depend on the tuning strategy.

    \item Transfer rankings depend on the model and task type.
In regression, MOMENT-PPG achieves the best normalized
cross-dataset transfer score, followed by MOMENT and Chronos-2,
whereas GPT-PPG transfers worst.
This reverses the classification ranking of the two specialist
families, where GPT-PPG transferred best and MOMENT-PPG worst.
Pretraining domain alone therefore does not determine
cross-dataset transfer performance.

    \item Larger models are again rarely the best choice. As in classification, the largest variant gives the best full-tuning result in only about one in three family-task pairs for the three-size families. That is no better than choosing a size at random. In some cases, larger models are clearly worse: MOMENT-PPG’s smallest variant gives its best result on most Inst. A clinical target. 
    
\end{itemize}

\subsection{Architectural Ablation Experiments}
\label{sec:arch_ablation}

Six of the eight FMs in Table~\ref{tab:models}, including MOMENT and GPT-PPG, are Transformers. While Transformers is ubiquitous and shown to be powerful for language modeling, there is no obvious reason why this particular architecture is also suitable for PPG signals, if at all. We therefore run an architectural ablation in this section. We focus on the component that defines the Transformer, the self-attention layer, replace it with other layers while keeping everything else fixed, and measure how this changes the usefulness of the pre-trained representations. We first describe the setup, then show that self-attention is a weak choice when the model is pre-trained to reconstruct the signal, as both MOMENT and GPT-PPG are, and finally show that this weakness arises during pre-training and is not caused by a worse fit to the training objective.

\subsubsection{Setup}

\paragraph{Token mixers.}
Each block of a Transformer has two layers: a token mixer, which lets each patch of the signal exchange information with the other patches, and a feed-forward layer, which processes each patch on its own. In the Transformer, the token mixer is self-attention. We compared it with seven other token mixers, covering the three main families of alternatives to attention: convolutions (a local depthwise convolution, a global depthwise convolution, and Hyena), state-space models (a selective state-space mixer) and recurrent networks (a linear recurrent unit, LRU; a GRU; and an LSTM). We also included a second attention variant, which replaces the Transformer's rotary position embeddings with ALiBi position biases~\cite{press2021train}. Every variant starts from the GPT-PPG 85M Transformer architecture and changes only its token mixer, in every block. All variants have the same number of parameters (within about 2\%) and are trained in the same way; only the token mixer differs.

\paragraph{Pre-training objectives.}
We used three pre-training objectives. Masked reconstruction, MOMENT's objective, hides parts of the signal and trains the model to reconstruct them. Next-patch prediction, GPT-PPG's objective, trains the model to predict each patch of the signal from the patches before it. We refer to both as signal-reconstruction objectives. The third objective is contrastive: it does not reconstruct the signal, but trains the model to give similar representations to two windows taken from the same 30-s segment. We use it as a control.

\paragraph{Two pre-training experiments.}
We first ran a large-scale experiment on the UCSF corpus (104.8M 30-s PPG segments), in which we pre-trained all nine variants, the Transformer, ALiBi attention and the seven alternatives, with masked reconstruction, using three seeds each. We then analyzed the result in more depth. We selected the three alternatives that performed best in the first experiment (local convolution, the selective state-space mixer and the LRU) and pre-trained them, together with the Transformer, on a second PPG corpus (9.6M segments from PulseDB, which combines MIMIC-III and VitalDB, and from MC-MED), this time under each of the three objectives and again with three seeds each. This second experiment tests whether the result holds on another corpus and under other objectives. Ideally, all nine variants would have been trained under every objective on the second corpus, and for as long as in the first experiment, but this would have required computing resources beyond our reach; the second experiment therefore uses three alternatives and one fifth of the training steps. Training details are in Appendix~\ref{app:arch_ablation}.

\paragraph{Evaluation.}
We evaluated every pre-trained model in the same way: we froze it and fitted a linear probe on its features for each task of the 51-task suite. We report the 16 tasks on which the probes learn something, selected by a fixed rule on the validation data: all six classification tasks, and the ten regression tasks on which the best model reaches a validation $R^2$ above 0.2. This rule measures if \textbf{any} of the model variant achieves a sensible performance; the tasks that fails this bar are deemed unsolvable by any of the variants tried and therefore left out of the analysis. On each task, we compare a variant with the Transformer by its relative improvement in F1 (classification) or MAE (regression), and we count the tasks on which the variant is better. To rank the variants overall, we also combine the 16 tasks into one composite score. Because the tasks use different metrics and units, we take each task's error on a log scale ($\log(1-\mathrm{F1})$ or $\log\mathrm{MAE}$), so that the same relative improvement counts the same on every task; we standardize it across models, so that every task carries the same weight; and we average over the 16 tasks. A lower composite is better.

\begin{table}[t]
\centering
\caption{The nine token-mixer variants pre-trained with masked reconstruction on the UCSF corpus, sorted by composite score. Tasks: the number of the 16 tasks on which the variant is better than the Transformer. Median gain: the median relative improvement over the Transformer across the 16 tasks (positive is better). Composite: the standardized log error averaged over the 16 tasks (lower is better). Tasks and median gain are computed from means over three seeds; $\pm$ gives the standard deviation over seeds (median gain: bootstrap over each variant's three seeds; composite: over the three runs).}
\label{tab:arch_ucsf}
\small
\begin{tabular}{llccc}
\toprule
\textbf{Token mixer} & \textbf{Type} & \textbf{Tasks} & \textbf{Median gain} & \textbf{Composite} \\
\midrule
Local convolution & Convolution & 14/16 & $+6.3 \pm 1.2\%$ & $-0.43 \pm 0.14$ \\
LRU & Recurrent & 13/16 & $+4.2 \pm 0.7\%$ & $-0.42 \pm 0.11$ \\
Selective state-space & State-space & 13/16 & $+5.7 \pm 1.4\%$ & $-0.41 \pm 0.10$ \\
Hyena & Convolution & 12/16 & $+5.5 \pm 0.6\%$ & $-0.37 \pm 0.03$ \\
Global convolution & Convolution & 14/16 & $+9.0 \pm 1.2\%$ & $-0.33 \pm 0.15$ \\
GRU & Recurrent & 11/16 & $+3.5 \pm 0.9\%$ & $+0.04 \pm 0.24$ \\
LSTM & Recurrent & 10/16 & $+0.8 \pm 0.7\%$ & $+0.32 \pm 0.12$ \\
\midrule
ALiBi attention & Attention & 9/16 & $+0.6 \pm 0.4\%$ & $+0.68 \pm 0.04$ \\
Transformer & Attention & -- & -- & $+0.91 \pm 0.10$ \\
\bottomrule 
\end{tabular}
\end{table}

\subsubsection{Under signal-reconstruction objectives, attention is a weak token mixer}

\paragraph{Masked reconstruction.}
We start with masked reconstruction on the UCSF corpus (Table~\ref{tab:arch_ucsf}). Every one of the seven alternatives learns better representations than the Transformer: each is better on 10 to 14 of the 16 tasks, with median gains between 0.8\% and 9.0\%. The result therefore does not depend on choosing the best alternative; any of the seven beats self-attention. Changing how attention encodes position does not help: ALiBi attention is better than the Transformer on only 9 of the 16 tasks. The two attention variants also have the two worst composite scores of the nine, and they do so separately in each of the three seeds. If the choice of token mixer did not matter, each of the 36 pairs of variants would be equally likely to have the two worst scores in a given seed, so the chance that the attention pair does so in all three seeds would be $(1/36)^3 \approx 2 \times 10^{-5}$. Accordingly, each of the 21 runs with a non-attention mixer has a better composite score than each of the six attention runs.

\begin{figure}[t]
\centering
\includegraphics[width=0.68\linewidth]{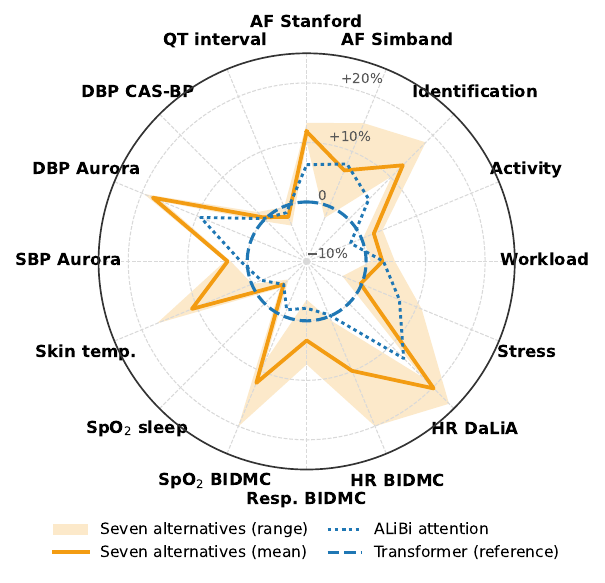}
\caption{Per-task improvement over the Transformer of the token mixers that replace self-attention (UCSF corpus, masked reconstruction; relative improvement in F1 or MAE, means over three seeds; positive is better). The Transformer is the dashed zero circle. The seven non-attention mixers are shown as their mean (solid line) and their range (shaded band); the dotted line is ALiBi attention.}
\label{fig:arch_radar}
\end{figure}

Figure~\ref{fig:arch_radar} shows where the gains come from. They fall on the same tasks for all seven alternatives. Averaged over the seven, the gains are largest for heart rate (DaLiA, $+20\%$), oscillometric diastolic blood pressure ($+18\%$), human identification ($+13\%$), SpO$_2$ (BIDMC, $+12\%$) and AF detection (Stanford, $+12\%$), and all seven are slightly worse than the Transformer on the same two tasks, SpO$_2$ in the sleep-disorder cohort and the QT interval.

The result holds on the second corpus (Table~\ref{tab:arch_second}). Under masked reconstruction, local convolution, LRU, and the selective state-space mixer outperform the Transformer on 11, 13, and 14 of the 16 tasks, and each run achieves a better composite score than the corresponding Transformer run. They also improve on the same tasks as on the UCSF corpus: across the 16 tasks, their gains on the two corpora have Spearman correlations of 0.75, 0.61 and 0.47.

\paragraph{Next-patch prediction.}
Like masked reconstruction, next-patch prediction also attempts to recover the signal from limited context, and this objective yields a similar picture. In next-patch prediction, a patch must not see the patches that it is trained to predict, so every mixer has to be causal: the Transformer uses causal attention, as in GPT-PPG, and we used causal versions of the three alternatives (Appendix~\ref{app:arch_ablation}). The causal alternatives outperform the Transformer on 13, 14, and 14 of the 16 tasks, and each run has a better composite score than the corresponding Transformer run.

\paragraph{The contrastive control.}
Under the contrastive objective, which does not reconstruct the signal, the gap narrows. The local convolution and the LRU are no longer better than the Transformer: they are better on only 6 and 8 of the 16 tasks, with median gains of $-0.5\%$ and $+0.3\%$, and their runs overlap with the Transformer's. The selective state-space mixer is still better on 13 of the 16 tasks, but by a median of only 0.6\%, against 5.8\% under masked reconstruction. Self-attention is therefore not uniformly bad for PPG: its deficit is large when the model is trained to reconstruct the signal and small otherwise, although it is still a suboptimal choice under the contrastive objective. Changing the objective has the matching effect on the Transformer itself: trained contrastively instead of with masked reconstruction, the Transformer improves on 10 of the 16 tasks, whereas the three alternatives improve on only 7 to 9 of the 16 tasks.

\begin{table}[t]
\centering
\caption{The Transformer and the three best alternatives of Table~\ref{tab:arch_ucsf}, pre-trained on the second corpus (PulseDB and MC-MED) under three objectives. Each cell gives the number of the 16 tasks on which the mixer is better than the Transformer trained with the same objective, and in brackets the median relative improvement (positive is better). Computed from means over three seeds; $\pm$ gives the standard deviation of the median gain from a bootstrap over each mixer's three seeds. Next-patch prediction uses causal versions of all mixers. The contrastive objective, which does not reconstruct the signal, is a control.}
\label{tab:arch_second}
\small
\setlength{\tabcolsep}{4pt}
\begin{tabular}{lccc}
\toprule
\textbf{Token mixer} & \multicolumn{3}{c}{\textbf{Pre-training objective}} \\
\cmidrule(lr){2-4}
 & Masked & Next-patch & Contrastive (control) \\
\midrule
Local convolution & 11/16 ($+3.2 \pm 0.8\%$) & 13/16 ($+1.6 \pm 0.4\%$) & 6/16 ($-0.5 \pm 0.3\%$) \\
LRU & 13/16 ($+2.9 \pm 0.7\%$) & 14/16 ($+2.5 \pm 0.5\%$) & 8/16 ($+0.3 \pm 0.5\%$) \\
Selective state-space & 14/16 ($+5.8 \pm 0.7\%$) & 14/16 ($+3.8 \pm 0.6\%$) & 13/16 ($+0.6 \pm 0.3\%$) \\
\bottomrule
\end{tabular}
\end{table}

\subsubsection{The deficit is learned, and it is not a failure to fit}

\paragraph{The deficit arises during pre-training.}
To test whether attention is weak before any pre-training, we built a synthetic PPG task in which beat timing and pulse morphology vary independently, and measured how well a linear probe on each model's frozen features recovers these factors (Appendix~\ref{app:arch_ablation}). With random weights, the two attention variants are not separated from the alternatives: they fall within the range of the seven. After pre-training on the UCSF corpus, both are worse than all seven alternatives. The weakness is therefore a result of what attention learns during pre-training, not of the architecture as such.

\paragraph{Attention fits the reconstruction objective equally well yet fails downstream.}
Attention also does not learn less useful features because it fits the objective worse. On the second corpus, the Transformer's held-out reconstruction error (0.0091--0.0093 over three seeds) lies within the range of the three alternatives (0.0087--0.0088 for the selective state-space mixer, 0.0093 for the local convolution and 0.0095--0.0097 for the LRU), yet each alternative's features are better than the Transformer's on 11 to 14 of the 16 tasks. The LRU, in particular, reconstructs the signal less accurately than the Transformer and still learns better features. Under next-patch prediction, likewise, the alternatives' held-out loss is only slightly lower than the Transformer's ($-1.697$ to $-1.737$ against $-1.688$, means over three seeds). Attention reaches an equally good reconstruction through a less informative solution, so a low reconstruction error does not guarantee a useful representation.

\paragraph{Interpretation.}
A plausible explanation lies in the quasi-periodic structure of PPG. A masked or upcoming patch closely resembles the same phase of a neighboring beat, so self-attention, which can look up and copy similar content anywhere in the input, can solve reconstruction largely by retrieval. Mixers whose weights depend on the relative position of patches rather than on their content, as in convolutions, or that pass information along the signal step by step, as recurrent and state-space models do, cannot copy in this way as easily; they have to model how the waveform evolves, which may yield more informative features. A contrastive objective offers less of this shortcut, consistent with the smaller deficit under it. For the comparison in this paper, the ablation means that both MOMENT and GPT-PPG pair a signal-reconstruction objective with the token mixer that suits it least. More generally, it suggests that the token mixer and the pre-training objective of a PPG FM should be chosen together, and that neither should be judged by how low the reconstruction loss becomes.

%% file: 8_results_others.tex
\section{Further analysis}
\label{sec:further_analysis}

\subsection{Frozen-encoder nearest-neighbor baseline}
\label{subsec:moment_frozen_knn}

We evaluated FMs as a frozen encoder followed by cosine-distance $k$-nearest-neighbor prediction ($k=5$).
No model parameters were updated, so this setting tests whether the pretrained representation already contains task-relevant structure.
Because the predictor still uses labeled reference examples, however, it should be interpreted as zero-finetuning transfer rather than strict label-free zero-shot task prediction.
The frozen embeddings transfer non-trivially to several classification tasks, including AF detection, activity recognition, and human identification, but are weaker on smaller mental-state tasks and difficult clinical regression targets.
These results suggest that generalist pretraining provides useful structure, but that many PPG health tasks still require task-specific adaptation or labeled reference data.
Full results with bootstrap uncertainty estimates are reported in Appendix~\ref{app:moment_frozen_knn} (Table~\ref{tab:moment_frozen_knn_cls}).

\subsection{Baseline, robustness, and transfer analyses}

We also tested whether FMs add value beyond simpler statistical, clinical-history, and traditional ML baselines.
For BP and HR, demographic or morphology-based baselines are competitive in selected settings, showing that waveform FMs are not automatically preferable to task-specific feature models.
For laboratory-test tasks, Last Observation Carried Forward (LOCF) is a strong clinical-history baseline whenever the same variable has been measured previously, and it outperforms FMs on several stable-patient tasks.
It is not a PPG waveform-forecasting baseline and is not applied to tasks without longitudinal target measurements.
This result is important because it identifies a practical boundary for PPG FMs: when the target is temporally persistent and recent clinical history is available, a simple history-based rule can be difficult to beat.
FMs can sometimes improve over LOCF during transition periods, but this result should be viewed as hypothesis-generating rather than as evidence of clinical readiness; prospective validation would be required before using such discrepancies to guide retesting or intervention.

\begin{table}[t]
\centering
\resizebox{\textwidth}{!}{%
\begin{tabular}{l|c|c|c|c|c}
\hline
{Task $/$ baseline type} & MOMENT & Chronos-2 & GPT-PPG & MOMENT-PPG & Signal/morph. baseline \\ 
\hline
\multicolumn{6}{l}{\textbf{Heart Rate Estimation} (MAE) $\downarrow$ $/$ Inter-Beat Interval (IBI)} \\
\hline
WESAD & 10.87 ± 0.09 & 9.78 ± 0.07 & 10.84 ± 0.08 & \textbf{9.54 ± 0.07} & 19.52$\pm$0.15 \\
\hline
DALIA & 12.98 ± 0.11 & 9.30 ± 0.11 & 9.39 ± 0.09 & \textbf{7.01 ± 0.08} & 16.80$\pm$0.14 \\
\hline
\hline
\multicolumn{6}{l}{\textbf{Respiration Rate Estimation} (MAE) $\downarrow$ $/$ Baseline Wander (BW)} \\
\hline
BIDMC & 2.09 ± 0.06 & 1.57 ± 0.05 & \textbf{1.30 ± 0.05} & 1.40 ± 0.05 & 5.85$\pm$0.56 \\
\hline
MIMIC-III & 4.64 ± 0.04 & 4.45 ± 0.04 & 4.50 ± 0.04 & \textbf{4.38 ± 0.04} & 8.04$\pm$0.19 \\
\hline
\end{tabular}
}
\caption{Comparison of representative FM configurations with task-specific signal-processing and morphology-based baselines. A fixed small-model configuration is used rather than selecting the best model size or tuning strategy separately for each task. These baselines are distinct from the supervised traditional-ML baseline pool used in the multi-family benchmark.
}
\label{tab:trad_bline}
\end{table}

Table~\ref{tab:trad_bline} shows representative comparisons with traditional signal-processing or morphology-based baselines, using the fully tuned small MOMENT, Chronos-2, GPT-PPG, and MOMENT-PPG for the FM columns.
Using classical baselines, we estimated heart rate (HR) and respiratory rate (RR) from short PPG windows.
Inter-Beat Interval (IBI) (peak-to-peak)~\cite{allen2007photoplethysmography} achieved 19.52~bpm MAE on WESAD (more than 2x FM (MOMENT-PPG) MAE), where beats were located via a smoothed derivative-squared energy envelope, and HR was computed as 60 / median(IBI).
Baseline-Wander (BW)~\cite{charlton2017breathing} on MIMIC-III yielded 8.04 brpm (breaths/min) MAE (approx. 2x FM (MOMENT-PPG) MAE) by isolating the slow trend and taking the dominant frequency in 0.1–0.5~Hz via Welch Power Spectral Density (PSD).
Additional statistical, LOCF, and transition-state baseline results are provided in Appendix~\ref{app:baseline_details}.
We highlight cases when these simpler baselines can outperform FMs especially in patient state transitions.

Additional analyses reveal further insights into the FMs.
In cross-domain fine-tuning, generalists and specialists tie with GPT-PPG performing better by discarding the generative loss during fine-tuning (Table~\ref{tab:robustness}).
In non-PPG biosignal testing, Chronos-2 wins all tasks except EEG classification for which GPT-PPG performs best (Table~\ref{tab:nonppg}).
The gender bias analysis shows that balanced training does not reliably reduce subgroup disparities; domain differences remain the dominant factor (Table~\ref{tab:type5_hr_gender_fairness}).
Full discussion is provided in Appendices~\ref{app:robustness_transfer_details} and \ref{sec:bias_study}.

\subsection{Controlled study of pre-training schemes}
\label{sec:mechanistic}

The comparisons above differ in pre-training data, objectives, and architectures, making it difficult to isolate the effect of generalist pre-training on PPG. To study this, we kept the architecture fixed and compared PPG pre-training from scratch with continued pre-training from a generalist checkpoint. We conducted this analysis using two models with different pre-training objectives: MOMENT (masked reconstruction) and Chronos-2 (forecasting).
Three findings emerge. First, starting PPG pre-training from the generalist checkpoint gives it a head start in downstream performance. Second, the head start does not depend on the quality of the checkpoint's representations: MOMENT's frozen features are statistically indistinguishable from random weights on PPG tasks, yet it gives a head start of the same order as Chronos-2, whose features are useful, so the checkpoint transfers to PPG as a parameter initialization rather than as a representation. Third, the off-the-shelf model's own downstream performance predicts which tasks gain the most early on, so the benefit of continued pre-training under a limited data or compute budget can be anticipated with a single linear probe, before any pre-training.

\subsubsection{Setup}

We compared three versions of each generalist: PT, the architecture pre-trained on PPG from random weights; CPT, the same pre-training started from the generalist checkpoint; and the off-the-shelf checkpoint without PPG pre-training. PT and CPT were trained on 9.6M 30-s segments from PulseDB and MC-MED with the generalist's own objective and recipe: MOMENT-Small (40M parameters) for 26{,}000 steps of 2{,}048 segments, as three independently seeded CPT/PT pairs, and Chronos-2 (120M parameters) for 112{,}896 steps of 512 segments, as one pair, with checkpoints saved during training. As a reference, we also built each architecture with random weights. Every model was frozen and evaluated with a linear probe on the 51-task suite. The probe's regularization was chosen by cross-validation on the training split, because a fixed regularization misranks pre-trained against random features when training data are few (Appendix~\ref{app:pretraining_scheme_details}). We report the 15 regression tasks on which any of these models extracts signal and the 6 classification tasks; the other 30 regression tasks carry no signal for any of them. Results over MOMENT's seed pairs are given as mean $\pm$ standard deviation.

\begin{figure}[t]
\centering
\includegraphics[width=0.8\linewidth]{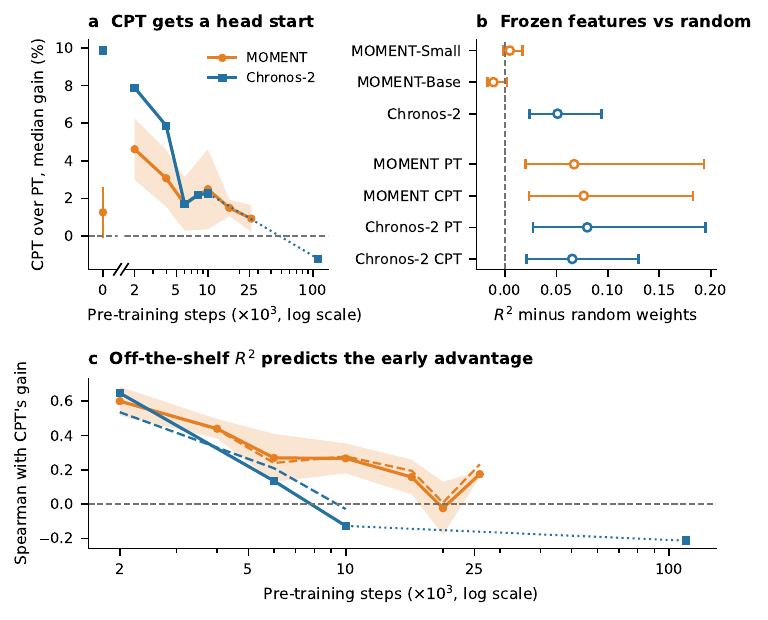}
\caption{Continued PPG pre-training from a generalist checkpoint (CPT) compared with pre-training from scratch (PT). \textbf{(a)}~Median relative gain of CPT over PT on the 15 informative tasks. \textbf{(b)}~Median difference in $R^2$ between frozen features and random weights of the same architecture over the 15 regression tasks, with bootstrap 95\% intervals. \textbf{(c)}~Spearman correlation across the 45 regression tasks between CPT's gain over PT and the $R^2$ of the off-the-shelf checkpoint (solid) or of random weights (dashed). MOMENT: mean $\pm$ s.d. over three seed pairs; Chronos-2: one pair.}
\label{fig:pretrain_schemes}
\end{figure}

\subsubsection{A generalist checkpoint gives a head start}

For both generalists, CPT is ahead of PT early in pre-training (Fig.~\ref{fig:pretrain_schemes}a; Appendix Table~\ref{tab:pretrain_headstart}). After 2{,}000 steps, MOMENT's CPT is better on $12.3 \pm 0.6$ of the 15 tasks (median relative gain $+4.6 \pm 1.6\%$), and Chronos-2's on 12 of 15 ($+7.9\%$). This demonstrates that the generalist model serves as a robust parameter initialization to train a specialist model. Such phenomenon is also observed in the context of language modeling, where some literatures name it "pre-pre-training" \cite{hu2025circuitschomskyprepretrainingformal, zhang2025intelligenceedgechaos}.     

\subsubsection{The head start does not require useful features}

MOMENT gives a clear head start even though its pre-trained checkpoint is no better than a randomly initialized network on PPG tasks. Its frozen features match those of the same network with random weights: the median difference in $R^2$ is $+0.005$ (95\% interval $-0.001$ to $+0.017$) over the 15 regression tasks, and $0.000$ in F1 over the 6 classification tasks (Fig.~\ref{fig:pretrain_schemes}b). This holds at every layer of the encoder, for other pooling readouts and input rates, at every training-set size, and for the larger MOMENT-Base (125M parameters) (Appendix~\ref{app:pretraining_scheme_details}, Table~\ref{tab:pretrain_random}). Accordingly, MOMENT's CPT starts level with PT, which starts from random weights (median gain $+1.3 \pm 1.3\%$ at step 0), and takes the lead only once PPG pre-training begins (Fig.~\ref{fig:pretrain_schemes}a). Chronos-2, in contrast, does carry useful features: its checkpoint beats random weights ($+0.051$, $+0.024$ to $+0.094$; better on 12 of the 15 regression tasks and on all 6 classification tasks), so its CPT starts ahead of PT (on 14 of 15 tasks, $+9.9\%$). Yet the head starts that the two checkpoints give PPG pre-training are of the same order. Representation quality decides where CPT starts, not whether it gains a head start: the generalist's weights lead PPG pre-training to good solutions faster without encoding those solutions themselves. Whether a checkpoint's frozen features carry anything for a new physiological signal therefore cannot be read off its pre-training; it has to be tested against the checkpoint's own random initialization with a tuned probe.

\subsubsection{Off-the-shelf performance predicts the early advantage}

Which tasks gain the most is predictable before any PPG pre-training, from the checkpoint's off-the-shelf performance alone and without knowing where the checkpoint comes from. Across the 45 regression tasks, the off-the-shelf model's $R^2$ predicts CPT's relative gain over PT after 2{,}000 steps (Spearman $+0.60 \pm 0.08$ for MOMENT and $+0.65$ for Chronos-2; Fig.~\ref{fig:pretrain_schemes}c). The two generalists confirm that what matters is the measured performance. MOMENT's off-the-shelf performance equals that of random weights (their $R^2$ correlate at $0.97$ across tasks), and accordingly the $R^2$ of random weights predicts MOMENT's early advantage equally well ($+0.60 \pm 0.07$; partial Spearman of the off-the-shelf $R^2$ given the random-weight $R^2$, $+0.10 \pm 0.08$). Chronos-2's off-the-shelf performance exceeds that of random weights, and its own $R^2$ predicts the early advantage beyond what random weights predict (partial Spearman $+0.45$). The early advantage is largest on the tasks where the checkpoint already extracts signal.

Like the head start itself, this prediction belongs to a limited pre-training budget. With enough PPG data, pre-training from scratch can learn whatever a checkpoint provides, so a head start must eventually vanish, as that of ImageNet pre-training does for long training schedules in vision~\cite{he2019rethinking}. Here it fades within about 3M segments for Chronos-2 and 12M for MOMENT, roughly a twentieth and a quarter of their pre-training budgets. By the end, Chronos-2's CPT and PT are level (median difference in $R^2$ $-0.015$, 95\% interval $-0.031$ to $+0.010$) and MOMENT's CPT keeps only a residual edge ($+0.007 \pm 0.007$), small next to what PPG pre-training adds over random weights ($+0.067$ for MOMENT's PT and $+0.080$ for Chronos-2's; $+0.06$ to $+0.11$ for every PPG-pre-trained encoder of Section~\ref{sec:arch_ablation}). The predictive power of the off-the-shelf $R^2$ fades accordingly (MOMENT $+0.27 \pm 0.14$ at 6{,}000 steps and $+0.17 \pm 0.02$ at the end; Chronos-2 $+0.13$ and $-0.21$; Fig.~\ref{fig:pretrain_schemes}c).

A limited budget is, however, exactly when continued pre-training is worth considering. When data or compute for PPG pre-training are scarce, the head start is the benefit a generalist checkpoint offers, and a single linear probe of the off-the-shelf checkpoint, fitted without any pre-training, indicates on which tasks that benefit will be largest.

\section{Discussion}
\label{sec:discussion}

This study examines the generalist \textit{vs} specialist question through a two-stage benchmark spanning multiple model families, model sizes, and adaptation settings.
Across the multi-family benchmark, PPG-specialist FMs achieve the highest aggregate win score, but their advantage is modest and task-dependent.
Chronos-2 leads the generalists by task-level win count; PulsePPG~\cite{saha2025pulse} has the highest individual win count overall; and traditional ML baselines~\cite{chen2016xgboost,ke2017lightgbm,breiman2001random} win on selected physiological and clinical regression tasks.
The focused comparison of MOMENT, Chronos-2, GPT-PPG, and MOMENT-PPG further shows that model rankings depend on the target, tuning strategy, and cross-dataset transfer setting.
Strong frozen features, large tuning gains, and larger model sizes do not consistently identify the best downstream model.

Our findings complement recent work on wearable and physiological FMs, including NormWear~\cite{luo2024toward}, Large Sensor Model (LSM)~\cite{narayanswamy2024scaling}, SleepFM~\cite{thapa2024sleepfm}, WBM~\cite{beyond-sensor}, PaPaGei~\cite{pillai2024papagei}, PulsePPG~\cite{saha2025pulse}, Apple PPG/ECG-FM~\cite{abbaspourazad2023large}, ECGFounder~\cite{li2024electrocardiogram}, and DeepECG-SSL~\cite{nolin2025foundation}.
Our study provides a common PPG evaluation framework for generalist FMs, PPG-specialist FMs, and traditional ML baselines, alongside experiments examining architecture and pre-training.
The architectural ablations show that token-mixer performance depends on the pre-training objective, and that comparable reconstruction losses can yield different representation quality.
The MOMENT and Chronos-2 experiments show an early downstream advantage from continued PPG pre-training over training from scratch, with this advantage diminishing as training continues.
These findings support choosing models and pre-training strategies according to the target task, adaptation setting, and available training budget.

\subsection{Data, architecture, and scaling}
 
The focused comparison of MOMENT, Chronos-2, GPT-PPG, and MOMENT-PPG illustrates how several design choices interact.
MOMENT~\cite{goswami2024moment} is more stable in several classification settings, while GPT-PPG~\cite{chen2025gpt,chen2024adapting} obtains stronger adaptation gains on several regression tasks after full tuning.
This pattern is consistent with differences in pre-training corpora, architectures, and objectives~\cite{he2022masked,touvron2023llama}.
The pre-training study sharpens this point. Under the evaluated training settings, continuing PPG pre-training from a generalist checkpoint gives a head start over pre-training from scratch for both MOMENT and Chronos-2. This suggests that generalist pre-training can provide a useful initialization without a corresponding advantage in frozen-feature performance. Because the checkpoint's off-the-shelf performance is associated with the size of the early advantage, a single linear probe may help assess the benefit of continued pre-training under a limited data or compute budget. Downstream performance, therefore, depends jointly on the alignment among pre-training objective, signal morphology~\cite{allen2007photoplethysmography,park2022photoplethysmogram}, target physiology, and adaptation strategy, rather than on pre-training domain alone~\cite{kornblith2019similarity}.

Model scaling also shows no simple monotonic pattern.
Larger models improve some settings but degrade or destabilize others~\cite{zhai2023stabilizing}, especially when the backbone is frozen, or the downstream dataset is small.
This suggests that brute-force scaling is not sufficient for physiological FMs~\cite{narayanswamy2024scaling}; scaling gains depend on how well the representation, objective, and adaptation protocol match the task.
The practical implication is that future PPG FM development should treat model size as one design axis among many rather than as a guaranteed route to better performance~\cite{ni2025ppg,ekambaram2024tiny}.
 
\subsection{Clinical benchmarking}
 
The clinical baseline results are central to interpreting the benchmark~\cite{bedi2025medhelm,karargyris2023federated}.
For stable patients and temporally persistent laboratory values, LOCF and other simple baselines can outperform complex FMs~\cite{im2025labtop,von2024electrolyte}.
PPG FMs should therefore be viewed as complementary tools, not replacements for clinical history or task-specific models~\cite{rasmy2021med}.
Their potential value is more plausible when prior measurements are unavailable, delayed, or unreliable during physiological transitions~\cite{wang2025estimating,kataria2025continuous}, but this use case remains to be validated prospectively.
This framing is also important for future benchmarks: reporting FM performance without clinical history and traditional ML baselines can overstate practical utility, particularly for low-dimensional physiological or laboratory targets~\cite{im2025labtop,von2024electrolyte}.
 
\subsection{Limitations and Future Directions}

Several limitations remain that can be addressed in future work.
First, our benchmarking operates on retrospective segment-level PPG evaluation.
Models are trained and evaluated on task-specific labels and short waveform windows.
This evaluates predictive capability rather than clinical actionability or improvements in patient outcomes.
For tasks with limited physiological predictability from PPG or weak labels, the benchmarking protocol is further limited.
Future work should therefore align with clinical decision-making points, use a longer history of signals (such as UNIPHY+~\cite{wang2025estimating}), and conduct prospective validation~\cite{liu2020reporting,collins2024tripod+}.

Second, our large-scale benchmark selects a broad range of heterogeneous datasets.
Reporting on various sources of variation, such as device types, population, and target variables, is challenging, especially when metadata is limited due to privacy concerns~\cite{im2024exploring}.
We partially mitigate this through task-type reporting, dataset sizes, gender, uncertainty estimates, and cross-dataset analyses.
Future work should incorporate richer metadata, including motion variations~\cite{ismail2021heart}, comorbidities, medication use, clinical history, skin factors~\cite{desquins2022survey}, and sensor type.

Third, zero-shot evaluation also remains unresolved.
Our frozen-encoder nearest-neighbor analysis evaluates zero-finetuning transfer, since the FM backbone is fixed and no task-specific model parameters are learned~\cite{alain2016understanding}, yet it still uses labeled reference examples and is distinct from strict direct zero-shot task prediction.
Because segment-level classification and scalar physiological or clinical regression do not yet have standardized prompting protocols, future benchmarks should distinguish direct zero-shot inference, frozen-representation retrieval, few-shot adaptation, and full fine-tuning~\cite{radford2021learning,ekambaram2024tiny}.

Fourth, the benchmarking setup needs improvement in task selection to ensure balance.
For instance, we find generalists to be vastly superior to specialists on classification tasks, but such tasks are much fewer (only six compared to 45 regression tasks) to make a general conclusion.
Also, by including transient baselines like clinical history-based LOCF, we assume availability of prior measurement.
Future physiological FM benchmarks should evaluate performance across stable and changing clinical states, and account for measurement recency and informative clinical testing patterns~\cite{sun2024incorporating}.

Given these limitations, our study is timely because it shifts from model-centered performance comparisons of PPG foundation models to task-regime characterization, clarifying when generalist pretraining transfers, when PPG-specific pretraining helps, when traditional or history-based baselines suffice, and when the target may not be physiologically observable from short PPG segments.
Furthermore, future benchmarks may move towards continuously updated, community-driven evaluation platforms inspired by the pairwise evaluation paradigm of Chatbot Arena and its recent extensions to medical LLMs.
The emergence of signal–language models further motivates extending these benchmarks beyond numerical prediction to include physiological question answering and signal-grounded explanations.
The evaluations should therefore assess numerical accuracy, calibration, and clinical usefulness of generated explanations~\cite{chiang2024chatbot,arai2025evaluating}.

\section{Conclusion}
The growing range of physiological FMs makes systematic evaluation essential for model selection and development.
Our two-stage PPG benchmark finds the clearest specialist advantages for targets directly reflected in the waveform, such as AF and vital signs, particularly with frozen backbones.
Generalists and simpler baselines remain competitive on many tasks.
Full-model tuning improves performance overall, changes model rankings, and narrows the specialist advantage.
Rankings also differ between in-domain and cross-dataset evaluation.
Larger models are not reliably better: the largest variant performs best in only about one third of family--task comparisons among the three-size families under full-model tuning. Controlled experiments suggest that architecture and pretraining objective should be chosen jointly: under signal-reconstruction objectives, non-attention token mixers yield stronger frozen representations than self-attention, the gap shrinks under contrastive learning, and low reconstruction loss does not guarantee useful features.
Generalist checkpoints give continued PPG pretraining an early advantage even when their frozen features are no better than random weights, but this advantage largely fades within about 3 million (Chronos-2) and 12 million (MOMENT) segments seen; an off-the-shelf linear probe indicates in advance which tasks benefit most.

These findings support reporting both frozen-backbone and full-model tuning, comparing frozen representations against random weights of the same architecture using tuned probes, and including strong traditional ML and relevant clinical-history baselines.
They also support task-specific validation of model size and adaptation strategy.
By linking model comparisons with controlled design experiments, this study provides a basis for more rigorous physiological FM benchmarking and practical guidance on model development and the allocation of training resources for PPG-based health assessment.

\section*{Acknowledgments}
We want to acknowledge the efforts of the Center for Data Science (School of Nursing, Emory University) cluster support team, including Viren Patel.
We also appreciate the Emory HPC support team, including Prasad Reddy and Circe Tsui.

\section*{Data Availability}
The datasets used in this study comprise private institutional collections, datasets requiring access agreements (DUA), and public datasets.

\textbf{Private datasets.} Simband, UCSF (also referred to as Inst.~A), and the multicentre CBFV-ICP collection are not currently publicly available. We plan to seek approval from the relevant institutional review boards to share de-identified data where permitted.

\textbf{Datasets requiring access agreements.} Access to Stanford, Microsoft Aurora (including its oscillometric and auscultatory subsets), CAS-BP, and Vital Videos is governed by the respective providers' data use agreements (DUAs); requests should follow the procedures described in the original dataset publications. The MIMIC-III Clinical Database (\url{https://physionet.org/content/mimiciii/1.4/}) and Multimodal Clinical Monitoring in the Emergency Department (MC-MED; \url{https://physionet.org/content/mc-med/}) require PhysioNet credentialing.

\textbf{Public datasets.} The following resources are available subject to their respective licenses: MIMIC-III Waveform Database Matched Subset v1.0 (\url{https://doi.org/10.13026/c2294b}); BIDMC (\url{https://doi.org/10.13026/C2208R}); BUT-PPG (\url{https://doi.org/10.13026/tn53-8153}); Mental Stress PPG (\url{https://github.com/talhaanwarch/PPG_mental_stress}); MAUS (\url{https://ieee-dataport.org/open-access/maus-dataset-mental-workload-assessment-n-back-task-using-wearable-sensor}); Pulse Transit Time PPG v1.1.0 (\url{https://doi.org/10.13026/jpan-6n92}); Real-World PPG (\url{https://doi.org/10.17632/yynb8t9x3d.1}); Sleep Disorder (\url{https://doi.org/10.6084/M9.FIGSHARE.1209662}); WESAD (\url{https://doi.org/10.24432/C57K5T}); PPG-DaLiA (\url{https://doi.org/10.24432/C53890}); Gyro-Acc-PPG (\url{https://github.com/hooseok/gyro_acc_ppg}); Welltory PPG (\url{https://github.com/Welltory/welltory-ppg-dataset}); PPG-BP (\url{https://figshare.com/articles/dataset/PPG-BP_Database_zip/5459299}); BCG (\url{https://doi.org/10.6084/m9.figshare.28416896.v1}); expanded Sleep-EDF (\url{https://doi.org/10.13026/C2X676}); PulseDB (\url{https://github.com/pulselabteam/PulseDB}); and VitalDB (\url{https://vitaldb.net/dataset/}).

We plan to release a GitHub repository consolidating dataset access information, links to the source publications, and details of the data used in this study.

\section*{Declarations}
\subsection*{Funding Statement}
This work was partially supported by NHLBI grant award R01HL166233.

\subsection*{Competing Statement}
No competing interest to report.

\subsection*{Author Contributions}
S.K., Y.W.,  Z.C., and X.H. conceptualized the project. S.K., Y.W., and Z.C. wrote major parts of the manuscript. H.G.K. and L.Y.P. conceptualized and conducted the fairness experiments. Y.X., Z.L., M.W., H.W. and X.W. helped in model training. R.X. and J.L. produced data processing code. A.Y. and A.Z. helped in acquiring and processing blood pressure and ICP datasets respectively. R.Y. helped in data analysis. A.F. helped in model pre-training. W.J., C.Y., J.G., S.R.B. helped in results interpretation. A.S. helped in further data acquisition. C.J., T.B., S.V.B, R.J.L., and X.H. did overall project direction and supervision of all above tasks mentioned. All authors reviewed the manuscript.

%% file: appendix_1.tex
\section{Foundation Models and Baselines}
\label{sec:foundation_model_intro}
We benchmark three groups of models: generalist time-series FMs pretrained on heterogeneous corpora that are not specific to PPG, PPG-specialist FMs pretrained on PPG signals, and traditional ML baselines evaluated on each downstream task. All FMs are adapted to downstream tasks through a lightweight prediction head, using either frozen-feature or full-model fine-tuning.
\subsection{Generalist time-series FMs}

\textbf{Chronos-2}~\cite{ansari2025chronos} is an encoder-only time-series foundation model that produces probabilistic forecasts using a quantile forecasting objective. Its group-attention mechanism supports information sharing across related time series. We evaluate the 28M and 120M variants in the focused four-family analysis.

\textbf{TimesFM}~\cite{das2024timesfm} is a decoder-only transformer of approximately 200M parameters for univariate forecasting. It divides a series into input patches and is trained to predict subsequent patches. TimesFM is pretrained on roughly 100 billion real-world time points, drawn mostly from web search-interest and online traffic data, supplemented with synthetic series.

\textbf{Moirai-2}~\cite{liu2025moirai} is a decoder-only time-series FM that combines quantile forecasting with multi-token prediction. It is pretrained on a curated corpus of tens of millions of series spanning many domains and frequencies. Moirai-2 is the successor to the original masked-encoder Moirai model.

\textbf{MOMENT}~\cite{goswami2024moment} is a transformer encoder pretrained with masked time-series modeling, in which randomly masked patches of the input are reconstructed from their surrounding context. It is pretrained on the Time Series Pile, a multi-domain collection of public time-series datasets, and is released in several sizes (approximately 40M, 125M, and 385M parameters).

None of the generalist FMs is exposed to PPG-specific morphology during pretraining; each reaches PPG only through the downstream adaptation step.

\subsection{PPG-specialist FMs}

\textbf{PaPaGei}~\cite{pillai2024papagei} is an open PPG FM pretrained on more than 57,000 hours of unlabeled PPG (about 20 million segments) drawn exclusively from publicly available datasets. It uses a morphology-aware contrastive objective that exploits differences in PPG pulse morphology across individuals, and can be used as a feature extractor or as an encoder for downstream models.

\textbf{PulsePPG}~\cite{saha2025pulse} is an open PPG FM trained with contrastive self-supervised learning on raw PPG collected during a 100-day field study of 120 participants. In contrast to specialist FMs pretrained on curated or clinical PPG, PulsePPG is pretrained on uncurated real-world wearable data, which the original work reports to improve generalization across both laboratory and field settings.

\textbf{GPT-PPG}~\cite{chen2024adapting,chen2025gpt} adapts the GPT architecture to PPG. The waveform is divided into one-second patches that serve as tokens, and the model is trained by autoregressive next-patch prediction; it uses rotary positional embeddings and RMSNorm. GPT-PPG is pretrained on more than 100 million 30-second PPG segments from over 25,000 hospitalized patients and is released at several scales. Note that we use only 40\% training data for GPT-PPG pre-training in this paper since we found its performance close to the original GPT-PPG model which used 100\% data (>200M segments).  We evaluate the 19M and 85M variants in the multi-family benchmark, and the 19M, 85M, and 345M variants in the focused four-family analysis.

\textbf{MOMENT-PPG} is identical to MOMENT except the pretraining data (UCSF/Inst. A data) follows GPT-PPG model as described above.
We train it on three model scales (40M, 125M, 385M) to enable fair comparison with model scales of other FMs.

These specialist FMs differ not only in objective and architecture but also in the source of their pretraining PPG: curated public datasets (PaPaGei), uncurated field recordings (PulsePPG), and hospital monitoring (GPT-PPG, MOMENT-PPG). We return to this distinction when interpreting the results.

\subsection{Traditional ML baselines: handcrafted features}
\label{app:features}

The linear, tree-based, and TabPFN baselines operate on a fixed 37-dimensional
feature vector extracted from each PPG window. Windows are min--max normalised
to $[0,1]$ before extraction, and features are computed at the dataset's
sampling rate (40\,Hz for the resampled signals used here). Table~\ref{tab:features}
lists the full set.

\begin{table}[htbp]
\small
\centering
\begin{tabular}{llp{0.56\textwidth}}
\toprule
\textbf{Group} & \textbf{\#} & \textbf{Features} \\
\midrule
Time-domain statistics & 13 & mean, standard deviation, minimum, maximum, median, interquartile range, skewness, kurtosis, root mean square, zero-crossing rate, mean absolute deviation, range, energy \\
\addlinespace
Pulse morphology & 11 & peak count, peak-to-peak amplitude, inter-beat interval (IBI) mean, IBI standard deviation, IBI RMSSD, pNN50, mean rise time, mean fall time, mean maximum upstroke slope, mean pulse area, mean pulse width \\
\addlinespace
Spectral & 8 & dominant frequency, dominant power ratio, spectral entropy, VLF power (0--0.5\,Hz), LF power (0.5--4\,Hz), HF power (4--10\,Hz), total power, signal-to-noise ratio \\
\addlinespace
Autocorrelation and complexity & 5 & autocorrelation at lag 1, autocorrelation at lag 10, Hjorth mobility, Hjorth complexity, sample-entropy proxy \\
\bottomrule
\end{tabular}
\caption{The 37 handcrafted features used by the linear, tree-based, and TabPFN
baselines. CNN1D and LSTM instead take the raw normalised waveform.}
\label{tab:features}
\end{table}

Peaks are detected with a prominence threshold of 0.05 and a minimum separation
of $0.3\,f_s$ samples; troughs are detected the same way on the inverted signal.
Pulse morphology features are averaged over the detected pulses, and a window in
which fewer than two peaks are found is assigned zero for all eleven morphology
features. Spectra are estimated by Welch's method with segment length
$\min(T, \max(64, 4f_s))$, and the signal-to-noise ratio is the summed power in
the LF and HF bands divided by the remaining power. The sample-entropy proxy is
the standard deviation of the absolute first difference, used as an inexpensive
stand-in for sample entropy rather than as its exact value. Any non-finite
feature value is replaced with zero.

We use the latest TabPFN models (v3). Because TabPFN performs in-context learning rather than gradient-based fitting, the training split is supplied as context and subsampled to at most 10,000 examples when necessary. PCA is applied when the number of features exceeds the model’s feature limit.

\begin{table*}[htbp]
\centering
\resizebox{1\textwidth}{!}{%
\begin{tabular}{|c|c|c|c|c|c|c|}
\hline
Task & Task Type ($N_{\text{class}}$) & Data Source & Channels & \#participants & Data size & Length \\
\hline

\multirow{2}{*}{AF Detection}
& \multirow{2}{*}{Classification (2)}
& Stanford~\cite{torres2020multi} & 1 & 163 & 285 h & 25 s \\
\cline{3-7}
& & Simband~\cite{shashikumar2017deep} & 1 & 98 & 7 h & 30 s \\
\hline

\multirow{2}{*}{\makecell{Mental Load\\Assessment}}
& \multirow{2}{*}{Classification (2)}
& Mental Stress PPG~\cite{anwar2022machine} & 1 & 27 & 4.77 h & 60 s \\
\cline{3-7}
& & MAUS~\cite{beh2021maus} & 1 & 22 & 4.375 h & 5 s \\
\hline

Activity Recognition & Classification (3)
& Pulse Transit PPG~\cite{mehrgardt2022pulse} & 2 & 22 & 9 h & 5 s \\
\hline

Human Identification & Classification (35)
& Real World PPG~\cite{realworldppg} & 1 & 35 & 3.5 h & 6 s \\
\hline

\multirow{3}{*}{SpO$_2$ Estimation}
& \multirow{3}{*}{Regression}
& Sleep Disorder~\cite{garde2014development} & 1 & 146 & 1,208.3 h & 10 s \\
\cline{3-7}
& & BIDMC~\cite{pimentel2017bidmc} & 1 & 53 & 7 h & 30 s \\
\cline{3-7}
& & MIMIC-III~\cite{moody2020mimiciiiwdb} & 1 & 4,382 & 6,500 h & 30 s \\
\hline

\multirow{2}{*}{\makecell{Temperature\\Estimation}}
& \multirow{2}{*}{Regression}
& Pulse Transit PPG~\cite{mehrgardt2022pulse} & 2 & 22 & 9 h & 5 s \\
\cline{3-7}
& & MIMIC-III~\cite{moody2020mimiciiiwdb} & 1 & 73 & 2,516 h & 30 s \\
\hline

\multirow{6}{*}{\makecell{Heart Rate\\Estimation}}
& \multirow{6}{*}{Regression}
& WESAD & 1 & 15 & 96.4 h & 8 s \\
\cline{3-7}
& & DALIA & 1 & 15 & 143.8 h & 8 s \\
\cline{3-7}
& & Gyro-Acc-PPG~\cite{8529266} & 3 & 24 & 1.2 h & 10 s \\
\cline{3-7}
& & BUT PPG~\cite{nemcova2021brno} & 1 & 50 & 10.8 h & 10 s \\
\cline{3-7}
& & Welltory PPG & 3 & 13 & 0.5 h & 5 s \\
\cline{3-7}
& & BIDMC & 1 & 53 & 7 h & 30 s \\
\hline

\multirow{2}{*}{\makecell{Potassium\\Estimation}}
& \multirow{2}{*}{Regression}
& MIMIC-III & 1 & 5,842 & 6,500 h & 30 s \\
\cline{3-7}
& & Inst. A~\cite{drew2014insights} & 1 & 21,000 & 143 h & 30 s \\
\hline

\multirow{2}{*}{\makecell{Sodium\\Estimation}}
& \multirow{2}{*}{Regression}
& MIMIC-III & 1 & 5,826 & 6,500 h & 30 s \\
\cline{3-7}
& & Inst. A & 1 & 21,446 & 150 h & 30 s \\
\hline

\multirow{2}{*}{\makecell{Glucose\\Estimation}}
& \multirow{2}{*}{Regression}
& MIMIC-III & 1 & 5,602 & 6,500 h & 30 s \\
\cline{3-7}
& & Inst. A & 1 & 6199 & 152 h & 30 s \\
\hline

A1c Estimation & Regression
& Inst. A & 1 & 1895 & 152 h & 30 s \\
\hline

\multirow{2}{*}{\makecell{Troponin\\Estimation}}
& \multirow{2}{*}{Regression}
& MIMIC-III & 1 & 1,548 & 6500 h & 30 s \\
\cline{3-7}
& & Inst. A & 1 & 4014 & 182 h & 30 s \\
\hline

\multirow{2}{*}{\makecell{Lactate\\Estimation}}
& \multirow{2}{*}{Regression}
& MIMIC-III & 1 & 4500 & 6500 h & 30 s \\
\cline{3-7}
& & Inst. A & 1 & 3746 & 300 h & 30 s \\
\hline

\multirow{2}{*}{\makecell{Respiration Rate\\Estimation}}
& \multirow{2}{*}{Regression}
& BIDMC & 1 & 53 & 7 h & 30 s \\
\cline{3-7}
& & MIMIC-III~\cite{moody2020mimiciiiwdb} & 1 & 4,397 & 6,500 h & 30 s \\
\hline

\multirow{7}{*}{\makecell{Blood Pressure\\Estimation}}
& \multirow{7}{*}{Regression}
& PPG-BP~\cite{liang2018new} & 1 & 219 & 0.4 h & 2.1 s \\
\cline{3-7}
& & Aurora-Oscillometric~\cite{mieloszyk2022comparison} & 1 & 1125 & 170.1 h & 30 s \\
\cline{3-7}
& & Aurora-Auscultatory~\cite{mieloszyk2022comparison} & 1 & 1125 & 21.0 h & 30 s \\
\cline{3-7}
& & CAS-BP~\cite{liu2023cuffless} & 4 & 1272 & 334 h & 30 s \\
\cline{3-7}
& & Vital Videos~\cite{toye2023vital} & 2 & 850 & 7.3 h & 30 s \\
\cline{3-7}
& & BCG & 1 & 40 & 4.6 h & 30 s \\
\cline{3-7}
& & BUT PPG~\cite{nemcova2021brno} & 1 & 50 & 10.8 h & 10 s \\
\hline

\multirow{3}{*}{\makecell{ECG metrics\\Estimation}}
& \multirow{3}{*}{Regression}
& Inst. A (PR Interval) & 1 & 7351 & 546 h & 30 s \\
\cline{3-7}
& & Inst. A (QRS Duration) & 1 & 7351 & 546 h & 30 s \\
\cline{3-7}
& & Inst. A (QT Interval) & 1 & 7351 & 546 h & 30 s \\
\hline

\multirow{2}{*}{\makecell{ECHO metrics\\Estimation}}
& \multirow{2}{*}{Regression}
& Inst. A (LVEF) & 1 & 3467 & 148 h & 30 s \\
\cline{3-7}
& & Inst. A (LVMass) & 1 & 2887 & 144 h & 30 s \\
\hline

\multirow{2}{*}{\makecell{Cholesterol\\Estimation}}
& \multirow{2}{*}{Regression}
& Inst. A (HDL) & 1 & 1645 & 172 h & 30 s \\
\cline{3-7}
& & Inst. A (LDL) & 1 & 1619 & 158 h & 30 s \\
\hline

\end{tabular}%
}
\caption{Overview of datasets for downstream tasks.
Temperature targets are dataset-specific: PPG sensor temperature
for Pulse Transit PPG and body temperature for MIMIC-III.}
\label{tab:data_overview}
\end{table*}

\section{Dataset details}
\label{sec:dataset_details}

This appendix provides detailed descriptions of the datasets used in the benchmark.

\subsubsection{Multi-purpose datasets}
Some datasets are endowed with multiple labels, which can be used to obtain subsets of data for different purposes and tasks.
Here, we list such datasets.

\noindent
\textbf{MIMIC-III~\cite{moody2020mimiciiiwdb}.}
Following standard practice, we use 30s, 125Hz, single-channel PPG signals (downsampled to 40Hz for experiments) from the MIMIC-III Waveform Database Matched Subset (MIMIC3WDB-matched)~\cite{moody2020mimiciiiwdb}, Version 1.0.
The MIMIC3WDB-matched dataset comprises various types of physiological signals, including PPG, ECG, and Arterial Blood Pressure (ABP), along with corresponding time series of vital signs collected from bedside monitors for 10,282 distinct adult and neonatal patients in the Intensive Care Unit (ICU).
We first derive three sub-datasets for vital signs estimation, including:
(1) SpO2 estimation,
(2) body temperature estimation, and
(3) respiration rate estimation, all sourced directly from MIMIC3WDB-matched.
Vital signs in the dataset are available at two frequencies: 1 Hz and 0.0167 Hz. For this study, we use the 1 Hz vital sign time series and link them to the corresponding high-frequency PPG signals. To ensure temporal alignment, we only retain samples where both the PPG signals and vital signs have continuous, overlapping 30s segments.
Additionally, we derive five laboratory measurement estimation sub-datasets by linking PPG signals from MIMIC3WDB-matched to corresponding laboratory test results from the MIMIC-III Clinical Database~\cite{johnson2016mimiciii}. The laboratory measurements include:
(1) Potassium,
(2) Sodium,
(3) Lactate,
(4) Glucose, and
(5) Troponin-T.
Laboratory tests occur at variable and irregular time intervals. For each lab result, we associate all PPG segments recorded within a one-hour window before the laboratory test time as corresponding PPG samples. Unlike prior studies~\cite{von2024electrolyte}, which may use a time window extending both before and after the lab test (e.g., -1 hour to +1 hour), our approach avoids potential data leakage by ensuring the PPG signals precede the lab result, thus preserving temporal causality for model development.

\noindent
\textbf{UCSF ICU dataset~\cite{drew2014insights,ding2024siamquality}.}
This multi-modal data was obtained in the intensive care units of UCSF Medical Center between March 2013 and December 2018.
It consists of multi-channel physiological waveforms, bedside monitor alarms, vital signs, and associated EHR from over 24,100 patients.
The waveforms consist of 7-lead ECGs (I, II, III, V, AVR, AVL, AVF), PPG, and respiration rate, which we all downsample to 40~Hz.
This collection was conducted with an approved waiver of written patient consent under the UCSF Institutional Review Board (IRB number: 14-13262).
For the pre-training of the GPT-PPG FM, we use 40\% of the total available data, i.e., 0.9 million hours, divided into non-overlapping 30-second segments.
For this dataset, we follow similar data and label extraction methodologies to those used in MIMIC-III.
Although the UCSF dataset is not publicly available, MIMIC-III provides a public comparison dataset for the overlapping vital-sign and laboratory-estimation tasks. The ECG, ECHO, and cholesterol tasks in this benchmark are evaluated using the UCSF data.
We can additionally utilize this dataset for conducting \emph{in-domain} testing.
As noted in Table~\ref{tab:data_overview}, we prepare 13 labeled subsets: six blood-chemistry targets (potassium, sodium, glucose, HbA1c, troponin, and lactate), three electrocardiographic (ECG) targets (PR interval, QRS duration, and QT interval), two echocardiographic (ECHO) targets (left ventricular ejection fraction and left ventricular mass), and two cholesterol targets (HDL and LDL cholesterol).
All these tasks belong to the regression task type, and the amount of data with labels present can vary from approximately 150 hours to over 500 hours, due to varying lab test frequencies.
For evaluating out-of-domain performance, we primarily leverage MIMIC-III data, which we noted is similar to UCSF.
In this paper, we also refer to the UCSF dataset as \emph{Institute A} or simply Inst. A for brevity.

\noindent
\textbf{BIDMC~\cite{pimentel2016toward}.}
The BIDMC PPG and Respiration Dataset was derived from the much larger MIMIC-II~\cite{saeed2011multiparameter} database, which was obtained from critically ill patients at the Beth Israel Deaconess Medical Centre (Boston, MA, USA).
Using the 53 recordings in the dataset, each lasting 8 minutes, we resample the PPG signals to 40~Hz and derive SpO$_2$, heart-rate, and respiration-rate labels.
To derive these labels per 30s segment, we choose the median value.

\noindent
\textbf{BUT PPG~\cite{nemcova2021brno}.}
We used the Brno University of Technology Smartphone PPG (BUT-PPG) database, consisting of 3,888 10s recordings of PPGs and corresponding ECG signals (for HR determination).
Simultaneous BP estimates were also provided by a device suitable for diabetes and hypertension.
This dataset is balanced across genders and can therefore be used to evaluate the bias of FMs (Table~\ref{tab:type5_hr_gender_fairness}).

\subsubsection{AF Detection}

\textbf{Stanford Dataset~\cite{torres2020multi}.}
The Stanford dataset comprises PPG recordings obtained from 163 individuals in ambulatory settings. 
We utilize the complete dataset with predefined train, validation, and test splits, ensuring no overlap of participants between sets to avoid data leakage and ensure fair evaluation.
Note that duplicate signals were removed from the original dataset, resulting in a total of 41,114 PPG segments (12,690 AF and 28,424 non-AF), each with a duration of 25 seconds, amounting to approximately 285 hours of recordings.

\noindent
\textbf{Simband Dataset~\cite{shashikumar2017deep}.}
The Simband dataset contains PPG recordings collected using the Samsung Simband smartwatch from 98 participants, including 45 with AF and 53 with other cardiac rhythms. The dataset is divided into 857 30-second segments, with 348 AF segments and 509 non-AF segments, totaling approximately 7 hours.
Simband was collected from patients at Emory University Hospital undergoing AF ablation, who consented to wear a study device for signal collection under IRB (00084629).


\subsubsection{Mental Load Assessment}

\noindent
\textbf{Mental Stress PPG~\cite{anwar2022machine}.}
This dataset includes PPG signals collected from 27 healthy university students (15 male, 12 female; mean age 21) using an ear-lobe sensor connected to an Arduino device. Recordings are taken during a baseline resting stage and while participants perform a Stroop test to induce stress, simulating real-life conditions outside a controlled lab.
The dataset comprises 153 stress segments and 133 non-stress segments, each lasting 60 seconds, totaling 4.77 hours.

\noindent
\textbf{MAUS~\cite{beh2021maus}.}
MAUS is a dataset for mental workload assessment using PPG signals collected from a wristband. It includes data from 22 participants who completed the PSQI questionnaire during resting sessions and performed N-back tasks during stress-inducing sessions. The dataset contains 1,575 stress segments and 1,575 non-stress segments, each lasting 5 seconds, totaling approximately 4.38 hours of recordings.

\subsubsection{Human Activity Recognition}

\noindent
\textbf{Pulse Transit Time PPG Dataset~\cite{mehrgardt2022pulse}.}
This dataset contains two-channel PPG recordings from 22 healthy subjects (16 male, six female) performing three physical activities: sitting, stationary walking, and running. Participants range in age from 20 to 53, with a mean age of 28.52 years. The dataset is evenly divided across the three activities, comprising 6,724 segments that last 5 seconds each, totaling approximately 9 hours.


\subsubsection{Human Identification}
\noindent
\textbf{Real World PPG~\cite{realworldppg}.}
The Real-World PPG dataset comprises 6-second wrist/finger PPG segments from 35 healthy participants (aged 10-75, with a mean age of 28.4 years), acquired using an IoT PPG sensor in everyday (uncontrolled) conditions.
Each segment has 300 samples at 50 Hz.
The authors provide a fixed split: 1,374 segments for training (66\%) and 700 for testing (34\%), which we adopt for 35-way subject identification.
We report top-1 accuracy on the official test set.
Several works utilize this corpus as a baseline for PPG biometrics and denoising under realistic noise conditions.

\subsubsection{SPO2 Estimation}

\noindent
\textbf{Sleep Disorder~\cite{garde2014development}.}
This pediatric cohort comprises 146 children referred for overnight polysomnography.
Additionally, a phone Oximeter recorded PPG at 62.5 Hz and SpO2 at 1 Hz on an adjacent finger, producing full-night traces per subject.
We segment the PPG into non-overlapping 10-s windows and pair each window with the contemporaneous SpO2 value (or short-term average) as the regression target, yielding 4.35 million seconds of data (1.21k hours) across all recordings.

\subsubsection{Heart Rate Estimation}

\noindent
\textbf{WESAD~\cite{schmidt2018introducing}.}
WESAD is a controlled lab corpus of 15 adults wearing a chest RespiBAN and wrist Empatic E4 under neutral, stress, and amusement conditions.
Numerous simultaneous signals are drawn, and we retain the PPG signals, resampling them to 40Hz.
Note that we extract signals for all mental and emotional states, resulting in diverse testing sets.

\noindent
\textbf{DALIA~\cite{reiss2019deep}.}
We used PPG-DaLiA, a daily living dataset of 15 adults wearing a wrist Empatica E4 and chest RespiBAN while performing eight real-world activities (e.g., sitting, walking, stair climbing, cycling, driving, lunch, table soccer, working) plus transitions.
Similar to WESAD, parallel signals are collected, like ECG and 3-axis acceleration, but we discard them.
A key difference between WESAD and DALIA is that the former is a controlled stress/affect lab corpus, and the latter is an ambulatory daily‑living PPG dataset targeting HR under motion.

\noindent
\textbf{Gyro-Acc-PPG~\cite{8529266}.}
This dataset includes 3-channel PPG recordings from 24 healthy subjects (10 male, 14 female; average age 26.9 ± 4.8 years) recruited at Wonkwang University. Participants perform walking and running on a treadmill while their heart rate is monitored. The dataset contains 432 PPG segments, each 10 seconds long, totaling approximately 1.2 hours.

\noindent
\textbf{Welltory PPG~\cite{neshitov2021wavelet}.}
This dataset contains 3-channel PPG recordings from 13 healthy participants aged 25–35. Each participant performs one or two measurements using a smartphone camera via the Welltory app, resulting in a total of 352 5-second PPG segments.

\subsubsection{Blood pressure estimation}\label{S418}

\noindent
\textbf{PPG-BP~\cite{liang2018new}.}
The PPG-BP dataset comprises 657 PPG segments, each 2.1 seconds in length, collected from 219 participants. Participants range in age from 20 to 89 years and include individuals with conditions such as hypertension and diabetes. The signals are sampled at a high rate of 1 kHz, though the total duration amounts to only 0.38 hours.

\noindent
\textbf{Microsoft Aurora~\cite{mieloszyk2022comparison}.}
This dataset, obtained with a Data Use Agreement (DUA), includes BP estimation results from 1,125 participants aged 21 to 85 (49.2\% female) across multiple hypertensive categories. BP is measured in the lab over 24 hours using both auscultatory and oscillometric methods, with a subset of participants also undergoing ambulatory monitoring. In our benchmark, the oscillometric and auscultatory subsets
contain 20,411 and 2,519 PPG segments, respectively. Each segment
lasts 30 seconds, corresponding to approximately 170.1 and
21.0 hours of PPG data, respectively.

\noindent
\textbf{CAS-BP~\cite{liu2023cuffless}.}
This dataset comprises 29,568 PPG recordings collected from 3,077 participants using a smartwatch. Participants range in age from 18 to 75 (65.16\% female), and 35.91\% of participants have a history of hypertension. Reference BP readings were obtained via dual-observer manual auscultation before and after each data recording session. Participants wore the smartwatch for 2 minutes to acquire signals. This procedure was repeated three times per day with a 5-minute interval. Recordings were collected on four different days within a month (D, D+7, D+14, and D+21), resulting in a maximum total of 12 recordings per participant.

\noindent
\textbf{Vital Videos~\cite{toye2023vital}.}
This corpus, obtained with a DUA, contains recordings from 850 unique participants.
For each participant, the study captured two 30-second uncompressed face videos synchronized with contact PPG and a single cuff BP measurement; demographics (age, sex, and skin tone) were also recorded.
We treat each 30s clip as one sample and 1) derive HR/PR labels from the simultaneously recorded contact PPG, and 2) pair both clips with the participant’s single BP label for BP estimation.

\noindent
\textbf{BCG~\cite{carlson2020bed}.}
This dataset includes time-aligned ECG, PPG, and Ballistocardiography (BCG) data, as well as continuous blood pressure data, collected from 40 participants aged 18 to 65 (57.5\% female). 10\% of participants have a history of cardiovascular-related conditions. Data collection was conducted in the lab using a custom bed system, resulting in over 4.5 hours of data being collected. All signals were sampled synchronously at a rate of 1 kHz.

\subsubsection{Cross-modality datasets}
\label{S419}

For cross-modality tasks, we investigate whether the PPG FM backbone can translate to brain-related targets as well as signals, which are derived from different sensors and anatomical mismatches.
First, we pursue Intra-Cranial Pressure (ICP)~\cite{KimHamilton2012,rajajee2024noninvasive} prediction using Cerebral Blood Flow Velocity (CBFV) recordings measured via Transcranial Doppler (TCD) technology.
In other words, we regress CBFV signals to invasive ICP values using a PPG FM model, 
This is important since the timely detection of intracranial hypertension can guide therapy.
Second, we pursue Electroencephalography (EEG) based sleep/wake binary classification~\cite{kemp2000analysis}.
In this task, we classify EEG signals into binary sleep/wake labels using the PPG FM backbone.
These two tasks are interrelated, and even their signal dynamics may share some commonalities.
Even though we pursue these tasks independently, effective adaptation of PPG FM can guide the development of non-invasive sensors at the bedside.

\noindent
\textbf{CBFV-ICP}
This dataset that we collected contains ICP and CBFV recordings from 156 patients across six institutions, including 75 from the University of California, Los Angeles (UCLA), 25 from the University of California, Davis (UCD), one from Emory University (IRB: STUDY00004039), 35 from Brain4Care (IRB: NCT03144219), one from NeuraSignal, Inc (IRB: 14430), and 19 from Wroclaw Medical University in Poland (IRB: KB-620/2020, KB-133/2023). The data from these institutions were obtained under Emory-approved DUAs. Each recording is 360 heartbeats with a sampling frequency of 400Hz, resulting in a total of 448 recordings. All the patients have various intracranial diseases, including traumatic brain injury (TBI), subarachnoid hemorrhage (SAH), and other diseases such as intraparenchymal hematoma, subdural hematoma, intraventricular hemorrhage (IVH), pontine hemorrhage, epidural hemorrhage, stroke, neoplasm, lumbar stenosis, and intracerebral hemorrhage (ICH). The participants range in age from 18 to 89, with a female population of 54.02\%.

\noindent
\textbf{Sleep-EDF~\cite{kemp2000analysis}}
For the EEG sleep classification task, we derived a binary sleep/wake dataset from the expanded Sleep-EDF Database.
Sleep-EDF offers comprehensive whole-night Polysomnography with Electroencephalography (EEG) (Fpz–Cz and Pz–Oz) and additional signals.
It also consists of expert hypnograms scored in 30-s epochs.
From each night, we extracted the Fpz–Cz EEG channel and down-sampled to 40 Hz.
We synchronized to the hypnogram and binarized labels, assigning wake (W) = 0 and sleep = 1 for all non-W stages (R, N1–N3), with movement time treated as sleep.
We then segmented the EEG into non-overlapping 30s windows (the standard scoring epoch) and assigned each window the majority label within the 30-s interval.

\subsection{Train \& Test Split}
\label{subsec:train_test_split}
\textbf{In-domain Evaluation.}
In-domain evaluation refers to the scenario where the training, validation, and testing sets are drawn from the same dataset. For all downstream tasks and datasets, except for user identification, we employ an inter-patient evaluation scheme in which the training and testing data are derived from different groups of participants to assess the model's generalizability. This approach is more practical and reflective of real-world scenarios. Specifically, for smaller datasets with fewer than 30 participants, we apply a leave-one-out methodology. In this setup, each participant's data serves as the testing set, while the remaining participants' data are split into training and validation sets in a 4:1 ratio. It is important to note that the training and validation sets share the same data distribution (i.e., data from the same participants can be included in both sets). This process is repeated until every participant has been used as the testing set, and the average performance across all iterations is reported as the final result for the given dataset. In each iteration, we select the model with the best performance on the validation set to assess its performance on the test set. 
For larger datasets without predefined splits, we randomly partition participant IDs into training, validation, and test sets using a 4:1:1 ratio. For Stanford AF detection, we retain the predefined training, validation, and test splits. In each case, the model achieving the best validation performance is evaluated on the corresponding test set.
Due to the inherent characteristics of user identification, the data from all participants must be included in the training set. Therefore, we simply randomly split the entire dataset into training, validation, and test sets. For user identification tasks, only in-domain evaluation is performed. The units for labels for heart rate are bpm (beats per minute), for potassium mEq/L, for sodium mEq/L, for glucose mg/dL, for troponin-T ng/ml, for SPO2 is percentage, for skin temperature Celsius, for blood pressure mmHg, for A1C percentage, for lactate mmol/L, for ECG metrics millisecond, for LVEF percentage, and for LVMass grams.

\textbf{Out-domain Evaluation.}
Additionally, we evaluate out-of-domain performance, where a model trained on one dataset is tested on a different dataset. All available data in the training set is used to fine-tune the FM, which is then evaluated on the entire testing set.

\subsection{Evaluation Metric Definations}
\label{sec:metrics_details}
To characterize the performance of MOMENT, Chronos-2, GPT-PPG, and MOMENT-PPG across datasets, model sizes, and tuning strategies, we evaluate them from seven complementary perspectives:

\noindent
\textbf{Win score.}
A straightforward descriptive metric to summarize benchmarking that counts the number of tasks for which the best performance is achieved by a model among competing models.
We also use \emph{win score combined}, which sums up the win scores of models belonging to a particular subset to facilitate group-level analysis.

\noindent
\textbf{Average Performance.}
For classification, we average F1 scores or accuracy across tasks, model sizes, and both tuning strategies.
For regression, we first normalize log MAE separately for each task:
\begin{equation}
z_{t,c,a}
=
\frac{\log \mathrm{MAE}_{t,c,a}-\mu_t}{\sigma_t}.
\end{equation}
Here, $t$ denotes the task, $c$ the model configuration, and $a$ the tuning strategy.
For each task, $\mu_t$ and $\sigma_t$ are the mean and population standard deviation of log MAE across all 11 model configurations and both tuning strategies.
Head-only and full-model results share the same normalization.
If $\sigma_t=0$, all normalized errors for that task are set to zero.
Each family's Performance score averages these normalized errors across tasks, its available model sizes, and both tuning strategies, giving each task equal weight.
Lower scores indicate better relative performance.

\noindent
\textbf{Scalability}. For each model family and size, we average task-level scores across tasks and both tuning strategies.
We use F1 or accuracy in percentage units for classification and negative task-normalized log error for regression.
We fit
\begin{equation}
S_{f,j}=a_f\log(n_{f,j})+b_f
\end{equation}
by ordinary least squares over each family's available sizes, where $n_{f,j}$ is the parameter count.
A positive slope indicates improvement with model size.

\noindent
\textbf{Feature Quality.}
Feature quality is reflected by the model’s average performance when only the task head is fine-tuned, with the FM kept frozen~\cite{chen2020simple}. This metric is crucial for evaluating the generality of the pretrained representations without relying on full model tuning. For classification, it averages F1 scores or accuracy.
For regression, it averages the head-only normalized log errors, using the same normalization shared with full-model tuning.

\noindent
\textbf{Tuning Gain.}
Tuning gain measures the relative improvement obtained by
full-model tuning over head-only tuning. For each matched task--model-size
comparison $i$,
\[
g_i=
\begin{cases}
\dfrac{S_i^{\mathrm{full}}-S_i^{\mathrm{head}}}{S_i^{\mathrm{head}}},
  & \text{classification},\\[10pt]
\dfrac{\mathrm{MAE}_i^{\mathrm{head}}-\mathrm{MAE}_i^{\mathrm{full}}}
       {\mathrm{MAE}_i^{\mathrm{head}}},
  & \text{regression},
\end{cases}
\qquad
\mathrm{Tuning\ Gain}=\frac{1}{N}\sum_{i=1}^{N}g_i ,
\]
where $S_i$ is the F1 score or accuracy and $N$ is the number of matched
task--model-size comparisons within a family. Head-only tuning updates only the
prediction head; full-model tuning updates the backbone as well. Positive values
indicate an average improvement from full-model tuning.

\noindent
\textbf{Variance.}
This axis describes dispersion across tasks, model sizes, and tuning strategies.
For classification, we retain the normalized standard deviation, calculated as the standard deviation divided by the mean and expressed as a percentage.
For regression, we report the population standard deviation of the task-normalized log errors.
Lower values indicate less dispersion.

\noindent
\textbf{Transferability.}
Transferability measures performance on datasets not used for downstream training.
Classification scores are averaged in their original F1 or accuracy scale.
For regression, log MAEs are standardized separately within each source--target task, using a common reference across all FM configurations included in that transfer comparison.
The resulting normalized errors are averaged across transfer tasks and the evaluated configurations within each family.
Lower regression scores indicate better relative transfer performance.

\section{Detailed multi-family results}
\label{app:detailed_multi_family_results}

This appendix reports the complete task-level results for all
eight FMs and the strongest traditional ML baseline,
extending the representative results in
Section~\ref{sec:multi_family_performance}. Each table lists every
task; for readability the models are split across two tables:
Table~\ref{tab:app_models_a} reports the four generalist foundation
models and PaPaGei, and Table~\ref{tab:app_models_b} reports the
remaining PPG-specialist FMs and the strongest traditional ML
baseline. Values are mean~$\pm$~standard deviation. Classification
tasks use F1 score (AF detection, mental tasks) or accuracy
(activity recognition, human identification), where higher is
better; regression tasks use MAE, where lower is better.
Blood-pressure rows report diastolic / systolic MAE.

\begin{table}[!h]
\scriptsize
\centering
\resizebox{\textwidth}{!}{%
\begin{tabular}{|l|c|c|c|c|c|}
\hline
\textbf{Task (Dataset)} & \textbf{Chronos-2 (120M)} & \textbf{TimesFM} & \textbf{Moirai-2} & \textbf{MOMENT (125M)} & \textbf{PaPaGei} \\
\hline
\multicolumn{6}{|l|}{\textit{Classification (F1 / accuracy \%, higher is better)}} \\
\hline
AF Detection (Stanford) & 87.77 $\pm$ 0.85 & 80.25 $\pm$ 1.14 & 83.51 $\pm$ 1.00 & 83.55 $\pm$ 1.09 & 81.43 $\pm$ 0.77 \\
\hline
AF Detection (Simband) & \textbf{69.66 $\pm$ 3.78} & 60.11 $\pm$ 4.19 & 54.76 $\pm$ 4.28 & 55.22 $\pm$ 3.94 & 61.75 $\pm$ 2.91 \\
\hline
Mental Workload (MAUS) & 68.54 $\pm$ 2.00 & 66.19 $\pm$ 2.18 & 62.45 $\pm$ 2.34 & 68.01 $\pm$ 2.37 & 58.42 $\pm$ 2.40 \\
\hline
Mental Stress (Mental Stress PPG) & 68.10 $\pm$ 5.71 & 70.11 $\pm$ 5.75 & 70.06 $\pm$ 5.52 & \textbf{72.73 $\pm$ 5.49} & 55.95 $\pm$ 7.33 \\
\hline
Activity Recognition (Pulse Transit) & 77.72 $\pm$ 1.14 & 85.85 $\pm$ 0.88 & 78.00 $\pm$ 1.13 & \textbf{87.96 $\pm$ 1.06} & 67.39 $\pm$ 1.23 \\
\hline
Human Identification (Real-world PPG) & 87.98 $\pm$ 1.26 & 96.30 $\pm$ 0.72 & 92.56 $\pm$ 1.00 & 92.43 $\pm$ 1.50 & 7.16 $\pm$ 1.01 \\
\hline
\multicolumn{6}{|l|}{\textit{Regression (MAE, lower is better; BP rows are Diastolic / Systolic)}} \\
\hline
SpO2 (MIMIC-III) & \textbf{2.21 $\pm$ 0.02} & 2.25 $\pm$ 0.02 & 2.25 $\pm$ 0.02 & 2.25 $\pm$ 0.02 & \textbf{2.21 $\pm$ 0.02} \\
\hline
Skin Temperature (MIMIC-III) & 1.97 $\pm$ 0.01 & 2.36 $\pm$ 0.01 & 2.25 $\pm$ 0.01 & 2.02 $\pm$ 0.01 & 1.60 $\pm$ 0.01 \\
\hline
Heart Rate (WESAD) & \textbf{9.60 $\pm$ 0.07} & 10.88 $\pm$ 0.08 & 10.87 $\pm$ 0.08 & 10.86 $\pm$ 0.08 & 10.36 $\pm$ 0.08 \\
\hline
Heart Rate (DALIA) & 9.03 $\pm$ 0.10 & 22.70 $\pm$ 0.15 & 22.35 $\pm$ 0.15 & 11.67 $\pm$ 0.11 & 7.94 $\pm$ 0.08 \\
\hline
Respiration Rate (MIMIC-III) & 4.17 $\pm$ 0.04 & 4.62 $\pm$ 0.04 & 4.76 $\pm$ 0.04 & 4.63 $\pm$ 0.04 & 4.31 $\pm$ 0.04 \\
\hline
BP (PPG-BP) & \textbf{8.50} / 15.36 & 9.02 / 17.03 & 38.22 / 92.36 & 9.14 / 17.10 & 8.69 / 17.11 \\
\hline
BP (Aurora-Oscillometric) & 7.02 / 19.13 & 25.14 / 25.08 & 20.87 / 25.08 & 22.19 / 25.12 & 23.83 / 23.12 \\
\hline
BP (Aurora-Auscultatory) & 10.07 / 13.25 & 9.77 / 13.81 & 9.77 / 18.63 & 9.70 / 13.77 & 9.76 / 13.86 \\
\hline
Potassium (MIMIC-III) & 0.47 $\pm$ 0.00 & 0.47 $\pm$ 0.00 & 0.50 $\pm$ 0.00 & 0.47 $\pm$ 0.00 & 0.47 $\pm$ 0.00 \\
\hline
Sodium (MIMIC-III) & 3.67 $\pm$ 0.03 & 3.69 $\pm$ 0.03 & 3.64 $\pm$ 0.03 & \textbf{3.63 $\pm$ 0.03} & 3.64 $\pm$ 0.03 \\
\hline
Glucose (MIMIC-III) & 35.64 $\pm$ 0.41 & 34.91 $\pm$ 0.45 & \textbf{34.87 $\pm$ 0.45} & 36.13 $\pm$ 0.39 & 35.45 $\pm$ 0.42 \\
\hline
A1c (Inst. A) & 1.77 $\pm$ 0.03 & 1.35 $\pm$ 0.03 & 1.51 $\pm$ 0.02 & 1.35 $\pm$ 0.03 & 1.42 $\pm$ 0.03 \\
\hline
Troponin (MIMIC-III) & 1.26 $\pm$ 0.02 & 1.16 $\pm$ 0.02 & 1.43 $\pm$ 0.02 & 1.26 $\pm$ 0.02 & 1.15 $\pm$ 0.02 \\
\hline
Lactate (MIMIC-III) & 1.25 $\pm$ 0.01 & 1.23 $\pm$ 0.02 & 1.18 $\pm$ 0.01 & 1.18 $\pm$ 0.01 & 1.19 $\pm$ 0.01 \\
\hline
\end{tabular}%
}
\caption{Detailed task-level results for the four generalist
FMs and PaPaGei. Classification tasks use F1
score or accuracy (higher is better); regression tasks use MAE
(lower is better). Blood-pressure rows report diastolic/systolic
MAE. Bold marks the best reported mean across both Tables~\ref{tab:app_models_a} and~\ref{tab:app_models_b}; ties at the displayed precision are bolded. Diastolic and systolic results are ranked separately. Bolding does not indicate statistical significance.}
\label{tab:app_models_a}
\end{table}

\begin{table}[!h]
\scriptsize
\centering
\resizebox{\textwidth}{!}{%
\begin{tabular}{|l|c|c|c|c|}
\hline
\textbf{Task (Dataset)} & \textbf{PulsePPG} & \textbf{GPT-PPG-19M} & \textbf{GPT-PPG-85M} & \textbf{Best traditional ML} \\
\hline
\multicolumn{5}{|l|}{\textit{Classification (F1 / accuracy \%, higher is better)}} \\
\hline
AF Detection (Stanford) & \textbf{93.93 $\pm$ 0.43} & 86.99 $\pm$ 0.91 & 87.63 $\pm$ 0.84 & CNN1D: 86.49 $\pm$ 0.90 \\
\hline
AF Detection (Simband) & 60.60 $\pm$ 3.12 & 67.91 $\pm$ 3.84 & 66.01 $\pm$ 3.82 & LogReg: 68.17 $\pm$ 3.56 \\
\hline
Mental Workload (MAUS) & \textbf{72.75 $\pm$ 2.13} & 61.57 $\pm$ 2.83 & 64.32 $\pm$ 2.58 & LSTM: 67.20 $\pm$ 2.26 \\
\hline
Mental Stress (Mental Stress PPG) & 65.59 $\pm$ 6.34 & 71.11 $\pm$ 5.55 & 64.86 $\pm$ 6.61 & CNN1D: 70.74 $\pm$ 5.43 \\
\hline
Activity Recognition (Pulse Transit) & 87.46 $\pm$ 0.94 & 78.81 $\pm$ 1.12 & 80.37 $\pm$ 1.07 & CNN1D: 84.16 $\pm$ 0.99 \\
\hline
Human Identification (Real-world PPG) & \textbf{97.82 $\pm$ 0.54} & 86.57 $\pm$ 1.31 & 88.14 $\pm$ 1.18 & RF: 77.12 $\pm$ 1.55 \\
\hline
\multicolumn{5}{|l|}{\textit{Regression (MAE, lower is better; BP rows are Diastolic / Systolic)}} \\
\hline
SpO2 (MIMIC-III) & 2.22 $\pm$ 0.02 & 2.46 $\pm$ 0.03 & 2.39 $\pm$ 0.03 & CNN1D: 2.23 $\pm$ 0.02 \\
\hline
Skin Temperature (MIMIC-III) & 2.06 $\pm$ 0.01 & \textbf{1.17 $\pm$ 0.01} & 1.18 $\pm$ 0.01 & LightGBM: 1.58 $\pm$ 0.01 \\
\hline
Heart Rate (WESAD) & 9.93 $\pm$ 0.08 & 10.84 $\pm$ 0.08 & 10.51 $\pm$ 0.08 & CNN1D: 9.94 $\pm$ 0.07 \\
\hline
Heart Rate (DALIA) & \textbf{5.91 $\pm$ 0.09} & 9.39 $\pm$ 0.09 & 8.03 $\pm$ 0.09 & XGBoost: 24.25 $\pm$ 0.13 \\
\hline
Respiration Rate (MIMIC-III) & 4.24 $\pm$ 0.04 & 4.50 $\pm$ 0.04 & 4.47 $\pm$ 0.04 & \textbf{CNN1D: 4.15 $\pm$ 0.04} \\
\hline
BP (PPG-BP) & 9.32 / 15.69 & 9.60 / 15.31 & 10.24 / 16.56 & LSTM 8.72 / RidgeReg \textbf{15.27} \\
\hline
BP (Aurora-Oscillometric) & \textbf{5.22} / \textbf{17.05} & 7.96 / 20.50 & 7.93 / 20.76 & CNN1D 8.95 / CNN1D 18.76 \\
\hline
BP (Aurora-Auscultatory) & 10.13 / 13.88 & 9.74 / 13.77 & 9.75 / 13.87 & RidgeReg \textbf{9.65} / CNN1D \textbf{13.01} \\
\hline
Potassium (MIMIC-III) & 0.47 $\pm$ 0.00 & 0.47 $\pm$ 0.00 & \textbf{0.46 $\pm$ 0.00} & \textbf{LSTM: 0.46 $\pm$ 0.00} \\
\hline
Sodium (MIMIC-III) & 3.78 $\pm$ 0.03 & 4.08 $\pm$ 0.03 & 4.08 $\pm$ 0.04 & LSTM: 3.65 $\pm$ 0.04 \\
\hline
Glucose (MIMIC-III) & 35.07 $\pm$ 0.44 & 35.52 $\pm$ 0.44 & 35.65 $\pm$ 0.43 & CNN1D: 34.91 $\pm$ 0.45 \\
\hline
A1c (Inst. A) & 1.49 $\pm$ 0.03 & 1.38 $\pm$ 0.03 & 1.33 $\pm$ 0.03 & \textbf{LSTM: 1.31 $\pm$ 0.03} \\
\hline
Troponin (MIMIC-III) & \textbf{1.11 $\pm$ 0.02} & 1.36 $\pm$ 0.02 & 1.35 $\pm$ 0.02 & LSTM: 1.15 $\pm$ 0.02 \\
\hline
Lactate (MIMIC-III) & \textbf{1.15 $\pm$ 0.01} & 1.23 $\pm$ 0.01 & 1.19 $\pm$ 0.01 & CNN1D: 1.18 $\pm$ 0.02 \\
\hline
\end{tabular}%
}
\caption{Detailed task-level results for the PPG-specialist
models PulsePPG and GPT-PPG (19M and 85M) and the strongest
traditional ML baseline. Classification tasks use F1 score or
accuracy (higher is better); regression tasks use MAE (lower is
better). Blood-pressure rows report diastolic / systolic MAE, and
the diastolic and systolic best traditional ML baselines may differ. Bold marks the best reported mean across both Tables~\ref{tab:app_models_a} and~\ref{tab:app_models_b}; ties at the displayed precision are bolded. Diastolic and systolic results are ranked separately. Bolding does not indicate statistical significance.}
\label{tab:app_models_b}
\end{table}

%% file: appendix_2.tex
\section{Detailed Classification Results for the Focused Generalist--Specialist Analysis}
\label{app:classification_details}

This appendix provides the detailed classification results underlying the seven-dimensional summary in Section~\ref{sec:classification}.
We report both head-only tuning, where the FM is frozen and only the task-specific prediction head is trained, and full-model tuning, where the entire model is updated.
Under head-only tuning, GPT-PPG achieves strong performance on atrial fibrillation detection, especially for the smaller 19M and 85M variants.
However, the largest 345M GPT-PPG model performs worse than the smaller GPT-PPG variants on Stanford AF detection and human identification.
These declines may reflect optimization or head-training instability.
Comparing the best configuration from each family, MOMENT outperforms GPT-PPG on MAUS, activity recognition, and human identification under head-only tuning, with non-overlapping 95\% intervals.

\begin{table}[htbp]
\small
\centering
\resizebox{\textwidth}{!}{%
\begin{tabular}{lccccc}
\toprule
\multirow{2}{*}{Task/Dataset (Metric)} & \multicolumn{3}{c}{MOMENT} & \multicolumn{2}{c}{Chronos-2} \\
\cmidrule(lr){2-4} \cmidrule(lr){5-6}
& 40M & 125M & 385M & 28M & 120M \\
\midrule
\multicolumn{6}{l}{\textbf{AF Detection (F1 Score, \%) $\uparrow$}} \\
Stanford & 55.47 $\pm$ 1.55 & 61.20 $\pm$ 1.45 & 66.74 $\pm$ 1.54 & 62.95 $\pm$ 1.40 & 66.77 $\pm$ 1.43 \\
Simband & 59.74 $\pm$ 5.06 & 58.19 $\pm$ 4.36 & 58.33 $\pm$ 4.07 & 59.76 $\pm$ 4.15 & 57.58 $\pm$ 4.35 \\
\midrule
\multicolumn{6}{l}{\textbf{Mental Workload (F1 Score, \%) $\uparrow$}} \\
MAUS & \underline{65.96 $\pm$ 2.44} & 62.90 $\pm$ 2.47 & \textbf{72.72 $\pm$ 2.56} & 60.38 $\pm$ 2.62 & 64.12 $\pm$ 2.28 \\
Mental Stress PPG & \underline{71.91 $\pm$ 5.53} & 71.11 $\pm$ 6.06 & \textbf{72.97 $\pm$ 6.08} & 69.47 $\pm$ 5.49 & 71.73 $\pm$ 5.26 \\
\midrule
\multicolumn{6}{l}{\textbf{Activity Recognition (Accuracy, \%) $\uparrow$}} \\
Pulse Transit & 77.03 $\pm$ 0.95 & 75.99 $\pm$ 0.97 & \textbf{82.09 $\pm$ 0.96} & \underline{81.97 $\pm$ 0.94} & 80.68 $\pm$ 0.98 \\
\midrule
\multicolumn{6}{l}{\textbf{Human Identification (Accuracy, \%) $\uparrow$}} \\
Real-world PPG & 83.57 $\pm$ 1.83 & \textbf{89.71 $\pm$ 1.84} & 81.00 $\pm$ 1.84 & 33.49 $\pm$ 1.80 & 30.61 $\pm$ 1.75 \\
\midrule
SCORE-class-head & 0/6 & 1/6 & \textbf{3/6} & 0/6 & 0/6 \\
SCORE-class-head-comb & \multicolumn{3}{c}{\textbf{4/6}} & \multicolumn{2}{c}{0/6} \\
\bottomrule
\end{tabular}
}
\caption{Detailed classification results for generalist FMs (MOMENT and Chronos-2) under head-only tuning. Bold and underline indicate the best and second-best task-level mean performance across all 11 model configurations in this table and Table~\ref{tab:cls_head_spec}. Win scores use this joint comparison; combined scores sum wins across sizes within each model family. Uncertainty is the bootstrap standard error over 1000 test-set resamples.}
\label{tab:cls_head_gen}
\end{table}

\begin{table}[htbp]
\small
\centering
\resizebox{\textwidth}{!}{%
\begin{tabular}{lcccccc}
\toprule
\multirow{2}{*}{Task/Dataset (Metric)} & \multicolumn{3}{c}{GPT-PPG} & \multicolumn{3}{c}{MOMENT-PPG} \\
\cmidrule(lr){2-4} \cmidrule(lr){5-7}
& 19M & 85M & 345M & 40M & 125M & 385M \\
\midrule
\multicolumn{7}{l}{\textbf{AF Detection (F1 Score, \%) $\uparrow$}} \\
Stanford & \underline{84.34 $\pm$ 0.95} & \textbf{85.83 $\pm$ 0.95} & 62.80 $\pm$ 1.31 & 68.36 $\pm$ 1.28 & 74.53 $\pm$ 1.27 & 75.89 $\pm$ 1.18 \\
Simband & 58.95 $\pm$ 3.89 & \textbf{64.89 $\pm$ 3.57} & 52.00 $\pm$ 4.46 & 53.10 $\pm$ 4.09 & 60.34 $\pm$ 3.77 & \underline{60.50 $\pm$ 3.75} \\
\midrule
\multicolumn{7}{l}{\textbf{Mental Workload (F1 Score, \%) $\uparrow$}} \\
MAUS & 57.92 $\pm$ 2.81 & 54.92 $\pm$ 2.78 & 60.04 $\pm$ 2.72 & 64.79 $\pm$ 2.61 & 59.73 $\pm$ 2.70 & 62.58 $\pm$ 2.65 \\
Mental Stress PPG & 71.11 $\pm$ 5.38 & 69.66 $\pm$ 5.76 & 64.00 $\pm$ 6.51 & 60.27 $\pm$ 6.96 & 64.71 $\pm$ 6.80 & 52.94 $\pm$ 7.45 \\
\midrule
\multicolumn{7}{l}{\textbf{Activity Recognition (Accuracy, \%) $\uparrow$}} \\
Pulse Transit & 68.25 $\pm$ 1.25 & 66.02 $\pm$ 1.28 & 64.16 $\pm$ 1.31 & 68.62 $\pm$ 1.26 & 70.19 $\pm$ 1.25 & 73.01 $\pm$ 1.18 \\
\midrule
\multicolumn{7}{l}{\textbf{Human Identification (Accuracy, \%) $\uparrow$}} \\
Real-world PPG & 82.14 $\pm$ 1.44 & 81.86 $\pm$ 1.47 & 44.00 $\pm$ 1.91 & 85.71 $\pm$ 1.34 & \underline{89.00 $\pm$ 1.16} & 85.29 $\pm$ 1.33 \\
\midrule
SCORE-class-head & 0/6 & 2/6 & 0/6 & 0/6 & 0/6 & 0/6 \\
SCORE-class-head-comb & \multicolumn{3}{c}{2/6} & \multicolumn{3}{c}{0/6} \\
\bottomrule
\end{tabular}
}
\caption{Detailed classification results for PPG-specialist FMs (GPT-PPG and MOMENT-PPG) under head-only tuning. Higher values are better. Bold and underline indicate the best and second-best task-level mean performance across all 11 model configurations in this table and Table~\ref{tab:cls_head_gen}. Win scores use this joint comparison; combined scores sum wins across sizes within each model family. Uncertainty is the bootstrap standard error over 1000 test-set resamples.}
\label{tab:cls_head_spec}
\end{table}

Under full-model tuning, MOMENT obtains the highest combined mean-based classification win score (3/6), followed by Chronos-2 (2/6), MOMENT-PPG (1/6), and GPT-PPG (0/6).
MOMENT-PPG and Chronos-2 obtain the highest mean F1 scores on Stanford and Simband, respectively, although the leading configuration on each dataset is tied with the best configuration from another family under the 95\% interval-overlap criterion.
These results indicate that full-model adaptation improves performance in selected settings, but its effects depend on the model and task.

\begin{table}[htbp]
\small
\centering
\resizebox{\textwidth}{!}{%
\begin{tabular}{lccccc}
\toprule
\multirow{2}{*}{Task/Dataset (Metric)} & \multicolumn{3}{c}{MOMENT} & \multicolumn{2}{c}{Chronos-2} \\
\cmidrule(lr){2-4} \cmidrule(lr){5-6}
& 40M & 125M & 385M & 28M & 120M \\
\midrule
\multicolumn{6}{l}{\textbf{AF Detection (F1 Score, \%) $\uparrow$}} \\
Stanford & 85.59 $\pm$ 1.09 & 83.55 $\pm$ 1.09 & 89.63 $\pm$ 1.11 & 82.30 $\pm$ 1.06 & 87.77 $\pm$ 0.85 \\
Simband & 64.32 $\pm$ 4.34 & 55.22 $\pm$ 3.94 & 64.84 $\pm$ 3.72 & 67.45 $\pm$ 3.87 & \textbf{69.66 $\pm$ 3.78} \\
\midrule
\multicolumn{6}{l}{\textbf{Mental Workload (F1 Score, \%) $\uparrow$}} \\
MAUS & \underline{68.24 $\pm$ 2.35} & 68.01 $\pm$ 2.37 & 66.03 $\pm$ 2.19 & 68.22 $\pm$ 2.05 & \textbf{68.54 $\pm$ 2.00} \\
Mental Stress PPG & \underline{72.50 $\pm$ 5.49} & \textbf{72.73 $\pm$ 5.49} & 71.11 $\pm$ 5.53 & 70.96 $\pm$ 5.31 & 68.10 $\pm$ 5.71 \\
\midrule
\multicolumn{6}{l}{\textbf{Activity Recognition (Accuracy, \%) $\uparrow$}} \\
Pulse Transit & \underline{86.99 $\pm$ 1.15} & \textbf{87.96 $\pm$ 1.06} & 84.23 $\pm$ 1.10 & 76.55 $\pm$ 1.16 & 77.72 $\pm$ 1.14 \\
\midrule
\multicolumn{6}{l}{\textbf{Human Identification (Accuracy, \%) $\uparrow$}} \\
Real-world PPG & \textbf{95.30 $\pm$ 1.62} & \underline{92.43 $\pm$ 1.50} & 89.86 $\pm$ 1.46 & 83.28 $\pm$ 1.40 & 87.98 $\pm$ 1.26 \\
\midrule
SCORE-class-full & 1/6 & \textbf{2/6} & 0/6 & 0/6 & \textbf{2/6} \\
SCORE-class-full-comb & \multicolumn{3}{c}{\textbf{3/6}} & \multicolumn{2}{c}{2/6} \\
\bottomrule
\end{tabular}
}
\caption{Detailed classification results for generalist FMs (MOMENT and Chronos-2) under full-model tuning. Higher values are better. Bold and underline indicate the best and second-best task-level mean performance across all 11 model configurations in this table and Table~\ref{tab:cls_full_spec}. Win scores use this joint comparison; combined scores sum wins across sizes within each model family. Uncertainty is the bootstrap standard error over 1000 test-set resamples.}
\label{tab:cls_full_gen}
\end{table}

\begin{table}[htbp]
\small
\centering
\resizebox{\textwidth}{!}{%
\begin{tabular}{lcccccc}
\toprule
\multirow{2}{*}{Task/Dataset (Metric)} & \multicolumn{3}{c}{GPT-PPG} & \multicolumn{3}{c}{MOMENT-PPG} \\
\cmidrule(lr){2-4} \cmidrule(lr){5-7}
& 19M & 85M & 345M & 40M & 125M & 385M \\
\midrule
\multicolumn{7}{l}{\textbf{AF Detection (F1 Score, \%) $\uparrow$}} \\
Stanford & 86.99 $\pm$ 0.91 & 87.63 $\pm$ 0.84 & 87.25 $\pm$ 0.88 & 85.89 $\pm$ 0.92 & \underline{89.83 $\pm$ 0.81} & \textbf{91.02 $\pm$ 0.72} \\
Simband & 67.91 $\pm$ 3.84 & 66.01 $\pm$ 3.82 & 61.11 $\pm$ 4.04 & 50.21 $\pm$ 3.89 & 61.54 $\pm$ 3.95 & \underline{68.77 $\pm$ 3.40} \\
\midrule
\multicolumn{7}{l}{\textbf{Mental Workload (F1 Score, \%) $\uparrow$}} \\
MAUS & 61.57 $\pm$ 2.83 & 64.32 $\pm$ 2.58 & 60.91 $\pm$ 2.80 & 67.53 $\pm$ 2.20 & 51.17 $\pm$ 3.11 & 62.30 $\pm$ 2.68 \\
Mental Stress PPG & 71.11 $\pm$ 5.55 & 64.86 $\pm$ 6.61 & 63.49 $\pm$ 7.47 & 58.06 $\pm$ 7.59 & 62.16 $\pm$ 6.51 & 65.82 $\pm$ 6.17 \\
\midrule
\multicolumn{7}{l}{\textbf{Activity Recognition (Accuracy, \%) $\uparrow$}} \\
Pulse Transit & 78.81 $\pm$ 1.12 & 80.37 $\pm$ 1.07 & 75.09 $\pm$ 1.18 & 72.04 $\pm$ 1.24 & 72.19 $\pm$ 1.21 & 72.04 $\pm$ 1.21 \\
\midrule
\multicolumn{7}{l}{\textbf{Human Identification (Accuracy, \%) $\uparrow$}} \\
Real-world PPG & 86.57 $\pm$ 1.31 & 88.14 $\pm$ 1.18 & 63.29 $\pm$ 1.82 & 58.71 $\pm$ 1.84 & 80.86 $\pm$ 1.50 & 78.00 $\pm$ 1.50 \\
\midrule
SCORE-class-full & 0/6 & 0/6 & 0/6 & 0/6 & 0/6 & 1/6 \\
SCORE-class-full-comb & \multicolumn{3}{c}{0/6} & \multicolumn{3}{c}{1/6} \\
\bottomrule
\end{tabular}
}
\caption{Detailed classification results for PPG-specialist FMs (GPT-PPG and MOMENT-PPG) under full-model tuning. Higher values are better. Bold and underline indicate the best and second-best task-level mean performance across all 11 model configurations in this table and Table~\ref{tab:cls_full_gen}. Win scores use this joint comparison; combined scores sum wins across sizes within each model family. Uncertainty is the bootstrap standard error over 1000 test-set resamples.}
\label{tab:cls_full_spec}
\end{table}

Taken together, the detailed classification tables show three main patterns.
First, MOMENT obtains higher combined win scores across the full classification suite.
Second, GPT-PPG performs strongly on atrial fibrillation detection, including an F1 score of 86.99\% for GPT-PPG-19M on Stanford under full-model tuning.
Third, model scaling is not uniformly beneficial: larger models improve some results but degrade others, especially when the backbone is frozen.
These detailed results motivate the seven-dimensional summary reported in the main text.

\section{Detailed Regression Results for the Focused Generalist--Specialist Analysis}
\label{app:regression_details}

This appendix provides the detailed regression results underlying the seven-dimensional summary in Section~\ref{sec:regression}.
The focused regression analysis includes 14 regression task types and 45 task-level evaluations after counting systolic/diastolic blood pressure and HDL/LDL cholesterol separately.
We report both head-only tuning, where the FM backbone is frozen and only the prediction head is trained, and full-model tuning, where the entire model is updated.
For readability, the detailed tables are split into two groups: Group 1 covers vital-sign and blood-pressure regression tasks, and Group 2 covers laboratory, ECG, ECHO, and cholesterol regression tasks.

\subsection{Head-only tuning on regression tasks}

\begin{table}[htbp]
\small
\centering
\resizebox{0.99\textwidth}{!}{%
\begin{tabular}{lccccc}
\toprule
\multirow{2}{*}{Task/Dataset (MAE)} & \multicolumn{3}{c}{MOMENT} & \multicolumn{2}{c}{Chronos-2} \\
\cmidrule(lr){2-4} \cmidrule(lr){5-6}
& 40M & 125M & 385M & 28M & 120M \\
\midrule
\multicolumn{6}{l}{\textbf{SpO$_2$ Estimation (MAE) $\downarrow$}} \\
Sleep Disorder & \underline{3.68 $\pm$ 0.05} & \textbf{3.53 $\pm$ 0.05} & 5.42 $\pm$ 0.05 & 5.26 $\pm$ 0.05 & 5.17 $\pm$ 0.05 \\
BIDMC & 5.07 $\pm$ 0.07 & 3.54 $\pm$ 0.07 & 3.64 $\pm$ 0.07 & 2.40 $\pm$ 0.06 & 2.44 $\pm$ 0.07 \\
MIMIC-III & 2.56 $\pm$ 0.02 & 2.57 $\pm$ 0.02 & 2.59 $\pm$ 0.02 & 2.21 $\pm$ 0.02 & 2.21 $\pm$ 0.02 \\
\midrule
\multicolumn{6}{l}{\textbf{Skin Temperature Estimation (MAE) $\downarrow$}} \\
MIMIC-III & 1.24 $\pm$ 0.01 & 1.45 $\pm$ 0.01 & 1.40 $\pm$ 0.01 & 1.70 $\pm$ 0.01 & 1.83 $\pm$ 0.01 \\
Pulse Transit & 2.20 $\pm$ 0.04 & 1.28 $\pm$ 0.06 & 1.42 $\pm$ 0.05 & 1.20 $\pm$ 0.04 & 1.20 $\pm$ 0.04 \\
\midrule
\multicolumn{6}{l}{\textbf{Heart Rate Estimation (MAE) $\downarrow$}} \\
WESAD & 10.41 $\pm$ 0.07 & 10.03 $\pm$ 0.07 & 9.69 $\pm$ 0.07 & 9.70 $\pm$ 0.07 & 10.19 $\pm$ 0.08 \\
DALIA & 13.24 $\pm$ 0.09 & 12.61 $\pm$ 0.09 & \textbf{10.59 $\pm$ 0.09} & 17.64 $\pm$ 0.13 & 17.87 $\pm$ 0.13 \\
BIDMC & 8.72 $\pm$ 0.26 & 9.40 $\pm$ 0.27 & 9.28 $\pm$ 0.27 & 8.03 $\pm$ 0.24 & 8.49 $\pm$ 0.25 \\
Gyro-Acc-PPG & 21.73 $\pm$ 0.62 & 21.03 $\pm$ 0.61 & 21.34 $\pm$ 0.61 & 20.02 $\pm$ 0.61 & \underline{19.97 $\pm$ 0.62} \\
Welltory-PPG & 8.33 $\pm$ 0.47 & 8.42 $\pm$ 0.46 & 8.76 $\pm$ 0.46 & \underline{7.25 $\pm$ 0.55} & 7.31 $\pm$ 0.48 \\
BUT-PPG & \underline{10.36 $\pm$ 0.30} & 10.43 $\pm$ 0.31 & 10.54 $\pm$ 0.31 & 10.69 $\pm$ 0.30 & 10.54 $\pm$ 0.30 \\
\midrule
\multicolumn{6}{l}{\textbf{Respiration Rate Estimation (MAE) $\downarrow$}} \\
BIDMC & 2.18 $\pm$ 0.06 & 2.15 $\pm$ 0.06 & 2.13 $\pm$ 0.06 & 1.92 $\pm$ 0.06 & 1.97 $\pm$ 0.06 \\
MIMIC-III & 4.63 $\pm$ 0.04 & 4.63 $\pm$ 0.04 & 4.64 $\pm$ 0.04 & \textbf{4.42 $\pm$ 0.04} & \underline{4.46 $\pm$ 0.04} \\
\midrule
\multicolumn{6}{l}{\textbf{Blood Pressure Estimation (Diastolic/Systolic MAE) $\downarrow$}} \\
\multirow{2}{*}{PPG-BP} & 9.13 $\pm$ 0.65 / & 9.17 $\pm$ 0.64 / & 9.57 $\pm$ 0.64 / & \textbf{8.74 $\pm$ 0.62} / & \underline{8.94 $\pm$ 0.62} / \\
& 16.68 $\pm$ 1.12 & 17.28 $\pm$ 1.11 & 17.01 $\pm$ 1.12 & 17.04 $\pm$ 1.12 & 17.07 $\pm$ 1.11 \\
\multirow{2}{*}{Aurora-Oscillometric} & 22.74 $\pm$ 0.31 / & 22.48 $\pm$ 0.29 / & 21.49 $\pm$ 0.27 / & 14.42 $\pm$ 0.22 / & 14.14 $\pm$ 0.21 / \\
& 25.44 $\pm$ 0.34 & 25.60 $\pm$ 0.34 & 25.07 $\pm$ 0.34 & 21.43 $\pm$ 0.33 & 21.60 $\pm$ 0.33 \\
\multirow{2}{*}{Aurora-Auscultatory} & 11.34 $\pm$ 0.39 / & 10.00 $\pm$ 0.39 / & 10.70 $\pm$ 0.39 / & \underline{9.52 $\pm$ 0.38} / & \textbf{9.49 $\pm$ 0.38} / \\
& 15.18 $\pm$ 0.48 & 14.10 $\pm$ 0.49 & 16.54 $\pm$ 0.49 & 13.99 $\pm$ 0.51 & 13.70 $\pm$ 0.49 \\
\multirow{2}{*}{CAS-BP} & 8.81 $\pm$ 0.10 / & 8.96 $\pm$ 0.10 / & 8.94 $\pm$ 0.10 / & 8.39 $\pm$ 0.10 / & 8.38 $\pm$ 0.10 / \\
& 14.69 $\pm$ 0.15 & 14.98 $\pm$ 0.15 & 14.91 $\pm$ 0.16 & 13.93 $\pm$ 0.16 & 13.90 $\pm$ 0.16 \\
\multirow{2}{*}{Vital Videos} & \textbf{7.89 $\pm$ 0.81} / & \underline{7.91 $\pm$ 0.80} / & 7.98 $\pm$ 0.83 / & 8.05 $\pm$ 0.73 / & 8.11 $\pm$ 0.72 / \\
& \underline{18.22 $\pm$ 2.20} & 18.31 $\pm$ 2.20 & \textbf{18.19 $\pm$ 2.20} & 18.23 $\pm$ 2.31 & 18.76 $\pm$ 2.09 \\
\multirow{2}{*}{BCG} & 5.75 $\pm$ 0.67 / & 5.81 $\pm$ 0.57 / & \textbf{5.32 $\pm$ 0.58} / & 5.82 $\pm$ 0.63 / & 5.90 $\pm$ 0.67 / \\
& 5.71 $\pm$ 0.62 & 4.39 $\pm$ 0.51 & 6.47 $\pm$ 0.23 & 4.45 $\pm$ 0.31 & 4.61 $\pm$ 0.21 \\
\multirow{2}{*}{BUT-PPG} & 8.67 $\pm$ 0.14 / & 9.49 $\pm$ 0.14 / & 13.09 $\pm$ 0.14 / & \textbf{7.51 $\pm$ 0.14} / & \underline{7.55 $\pm$ 0.14} / \\
& 47.33 $\pm$ 0.28 & 41.94 $\pm$ 0.28 & 48.30 $\pm$ 0.28 & \underline{14.18 $\pm$ 0.28} & \textbf{14.02 $\pm$ 0.29} \\
\midrule
SCORE-reg-head-p1 & 1/27 & 1/27 & 3/27 & 3/27 & 2/27 \\
SCORE-reg-head-p1-comb & \multicolumn{3}{c}{5/27} & \multicolumn{2}{c}{5/27} \\
\bottomrule
\end{tabular}
}
\caption{Detailed regression results for generalist FMs (MOMENT and Chronos-2) under head-only tuning on Group 1 tasks. Lower MAE is better. Bold and underline indicate the best and second-best task-level mean performance across all 11 model configurations in this table and Table~\ref{tab:reg1_head_spec}. Win scores use this joint comparison, with equal reported mean MAEs sharing the win credit; combined scores sum wins across sizes within each model family. Diastolic and systolic blood-pressure targets are scored separately, yielding 27 task-level evaluations. Uncertainty is the bootstrap standard error over 1000 test-set resamples. Diastolic and systolic blood-pressure MAEs are reported on separate lines, each accompanied by its bootstrap standard error.}
\label{tab:reg1_head_gen}
\end{table}

\begin{table}[htbp]
\small
\centering
\resizebox{0.99\textwidth}{!}{%
\begin{tabular}{lcccccc}
\toprule
\multirow{2}{*}{Task/Dataset (MAE)} & \multicolumn{3}{c}{GPT-PPG} & \multicolumn{3}{c}{MOMENT-PPG} \\
\cmidrule(lr){2-4} \cmidrule(lr){5-7}
& 19M & 85M & 345M & 40M & 125M & 385M \\
\midrule
\multicolumn{7}{l}{\textbf{SpO$_2$ Estimation (MAE) $\downarrow$}} \\
Sleep Disorder & 6.67 $\pm$ 0.08 & 4.04 $\pm$ 0.08 & 4.05 $\pm$ 0.08 & 3.93 $\pm$ 0.07 & 4.14 $\pm$ 0.07 & 3.98 $\pm$ 0.07 \\
BIDMC & 2.04 $\pm$ 0.07 & 2.58 $\pm$ 0.08 & 2.44 $\pm$ 0.07 & 2.07 $\pm$ 0.06 & \textbf{1.92 $\pm$ 0.06} & \underline{1.94 $\pm$ 0.07} \\
MIMIC-III & 2.83 $\pm$ 0.03 & 2.29 $\pm$ 0.03 & 2.46 $\pm$ 0.03 & \underline{2.19 $\pm$ 0.02} & \textbf{2.18 $\pm$ 0.02} & 2.20 $\pm$ 0.02 \\
\midrule
\multicolumn{7}{l}{\textbf{Skin Temperature Estimation (MAE) $\downarrow$}} \\
MIMIC-III & \underline{1.18 $\pm$ 0.01} & \textbf{0.84 $\pm$ 0.01} & 1.24 $\pm$ 0.01 & 1.84 $\pm$ 0.01 & 1.68 $\pm$ 0.01 & 1.76 $\pm$ 0.01 \\
Pulse Transit & 1.07 $\pm$ 0.04 & 1.25 $\pm$ 0.04 & 1.22 $\pm$ 0.05 & 1.01 $\pm$ 0.04 & \textbf{0.98 $\pm$ 0.04} & \underline{0.99 $\pm$ 0.04} \\
\midrule
\multicolumn{7}{l}{\textbf{Heart Rate Estimation (MAE) $\downarrow$}} \\
WESAD & 10.89 $\pm$ 0.09 & 10.44 $\pm$ 0.08 & 10.32 $\pm$ 0.08 & 9.57 $\pm$ 0.07 & \textbf{9.41 $\pm$ 0.07} & \underline{9.51 $\pm$ 0.07} \\
DALIA & 16.81 $\pm$ 0.13 & 16.05 $\pm$ 0.13 & 17.14 $\pm$ 0.13 & 11.27 $\pm$ 0.10 & 10.88 $\pm$ 0.09 & \underline{10.60 $\pm$ 0.09} \\
BIDMC & 10.58 $\pm$ 0.30 & 10.35 $\pm$ 0.29 & 9.67 $\pm$ 0.30 & 6.54 $\pm$ 0.20 & \underline{6.17 $\pm$ 0.20} & \textbf{6.00 $\pm$ 0.20} \\
Gyro-Acc-PPG & 20.23 $\pm$ 0.64 & 20.25 $\pm$ 0.65 & 20.26 $\pm$ 0.59 & \textbf{19.92 $\pm$ 0.62} & 19.98 $\pm$ 0.62 & 20.42 $\pm$ 0.64 \\
Welltory-PPG & 10.72 $\pm$ 1.29 & 11.92 $\pm$ 1.30 & 8.79 $\pm$ 0.69 & \textbf{7.23 $\pm$ 0.54} & 9.06 $\pm$ 0.66 & 9.13 $\pm$ 0.82 \\
BUT-PPG & 11.48 $\pm$ 0.34 & 11.78 $\pm$ 0.37 & 11.66 $\pm$ 0.34 & 10.47 $\pm$ 0.29 & \textbf{10.27 $\pm$ 0.28} & 10.40 $\pm$ 0.26 \\
\midrule
\multicolumn{7}{l}{\textbf{Respiration Rate Estimation (MAE) $\downarrow$}} \\
BIDMC & 1.86 $\pm$ 0.06 & 1.98 $\pm$ 0.06 & 2.05 $\pm$ 0.06 & \underline{1.81 $\pm$ 0.05} & \textbf{1.72 $\pm$ 0.05} & \textbf{1.72 $\pm$ 0.05} \\
MIMIC-III & 4.64 $\pm$ 0.04 & 4.61 $\pm$ 0.04 & 4.63 $\pm$ 0.04 & 4.50 $\pm$ 0.04 & 4.47 $\pm$ 0.04 & 4.49 $\pm$ 0.04 \\
\midrule
\multicolumn{7}{l}{\textbf{Blood Pressure Estimation (Diastolic/Systolic MAE) $\downarrow$}} \\
\multirow{2}{*}{PPG-BP} & 43.31 $\pm$ 1.02 / & 59.96 $\pm$ 1.03 / & 31.28 $\pm$ 1.02 / & 9.21 $\pm$ 0.65 / & 9.28 $\pm$ 0.68 / & 9.31 $\pm$ 0.67 / \\
& 17.44 $\pm$ 1.06 & 18.04 $\pm$ 1.16 & 16.62 $\pm$ 1.11 & \underline{16.46 $\pm$ 1.07} & 16.89 $\pm$ 1.14 & \textbf{15.68 $\pm$ 1.13} \\
\multirow{2}{*}{Aurora-Oscillometric} & \underline{8.42 $\pm$ 0.22} / & \textbf{8.18 $\pm$ 0.23} / & 8.44 $\pm$ 0.22 / & 9.16 $\pm$ 0.22 / & 8.86 $\pm$ 0.22 / & 9.12 $\pm$ 0.22 / \\
& 20.98 $\pm$ 0.36 & 20.94 $\pm$ 0.36 & 20.91 $\pm$ 0.34 & \underline{20.22 $\pm$ 0.34} & \textbf{20.11 $\pm$ 0.35} & 20.39 $\pm$ 0.33 \\
\multirow{2}{*}{Aurora-Auscultatory} & 9.73 $\pm$ 0.42 / & 9.74 $\pm$ 0.42 / & 9.71 $\pm$ 0.41 / & 10.35 $\pm$ 0.44 / & 9.97 $\pm$ 0.43 / & 9.97 $\pm$ 0.41 / \\
& 13.77 $\pm$ 0.54 & 13.77 $\pm$ 0.58 & 13.77 $\pm$ 0.55 & \underline{13.51 $\pm$ 0.50} & \textbf{13.45 $\pm$ 0.53} & 13.75 $\pm$ 0.53 \\
\multirow{2}{*}{CAS-BP} & 8.40 $\pm$ 0.10 / & 8.41 $\pm$ 0.11 / & 8.41 $\pm$ 0.10 / & 8.34 $\pm$ 0.10 / & \textbf{8.20 $\pm$ 0.10} / & \underline{8.33 $\pm$ 0.10} / \\
& 14.33 $\pm$ 0.16 & 14.33 $\pm$ 0.17 & 14.31 $\pm$ 0.16 & 13.66 $\pm$ 0.14 & \underline{13.55 $\pm$ 0.14} & \textbf{13.43 $\pm$ 0.14} \\
\multirow{2}{*}{Vital Videos} & 22.88 $\pm$ 1.28 / & 8.89 $\pm$ 0.84 / & 8.06 $\pm$ 0.80 / & 8.03 $\pm$ 0.91 / & 8.30 $\pm$ 0.89 / & 8.24 $\pm$ 0.74 / \\
& 67.69 $\pm$ 3.26 & 41.92 $\pm$ 3.39 & 18.59 $\pm$ 2.19 & 18.76 $\pm$ 2.31 & 19.20 $\pm$ 2.27 & 18.59 $\pm$ 2.27 \\
\multirow{2}{*}{BCG} & 5.82 $\pm$ 0.60 / & \underline{5.60 $\pm$ 0.71} / & 6.56 $\pm$ 0.82 / & 7.11 $\pm$ 0.58 / & 7.19 $\pm$ 0.97 / & 7.11 $\pm$ 0.60 / \\
& \underline{4.19 $\pm$ 0.41} & \textbf{4.12 $\pm$ 0.41} & 4.45 $\pm$ 0.10 & 4.53 $\pm$ 0.31 & 5.13 $\pm$ 0.41 & 4.92 $\pm$ 0.64 \\
\multirow{2}{*}{BUT-PPG} & 9.80 $\pm$ 0.28 / & 9.12 $\pm$ 0.26 / & 8.66 $\pm$ 0.20 / & 7.70 $\pm$ 0.14 / & 7.94 $\pm$ 0.15 / & 8.57 $\pm$ 0.16 / \\
& 15.77 $\pm$ 0.41 & 16.83 $\pm$ 0.49 & 16.48 $\pm$ 0.35 & 14.50 $\pm$ 0.30 & 15.63 $\pm$ 0.30 & 20.18 $\pm$ 0.42 \\
\midrule
SCORE-reg-head-p1 & 0/27 & 3/27 & 0/27 & 2/27 & \textbf{8.5/27} & 3.5/27 \\
SCORE-reg-head-p1-comb & \multicolumn{3}{c}{3/27} & \multicolumn{3}{c}{\textbf{14/27}} \\
\bottomrule
\end{tabular}
}
\caption{Detailed regression results for PPG-specialist FMs (GPT-PPG and MOMENT-PPG) under head-only tuning on Group 1 tasks. Lower MAE is better. Bold and underline indicate the best and second-best task-level mean performance across all 11 model configurations in this table and Table~\ref{tab:reg1_head_gen}. Win scores use this joint comparison, with equal reported mean MAEs sharing the win credit; combined scores sum wins across sizes within each model family. Diastolic and systolic blood-pressure targets are scored separately, yielding 27 task-level evaluations. Uncertainty is the bootstrap standard error over 1000 test-set resamples. Diastolic and systolic blood-pressure MAEs are reported on separate lines, each accompanied by its bootstrap standard error.}
\label{tab:reg1_head_spec}
\end{table}

Tables~\ref{tab:reg1_head_gen} and~\ref{tab:reg1_head_spec} summarize head-only tuning results for vital-sign and blood-pressure regression tasks.
MOMENT-PPG obtains the highest combined mean-based win score in Group 1 (14/27), compared with 5/27 for both MOMENT and Chronos-2 and 3/27 for GPT-PPG.
MOMENT outperforms GPT-PPG on WESAD and DALIA heart-rate estimation, with non-overlapping 95\% intervals when comparing the best configuration from each family.
Blood-pressure estimation shows dataset-dependent behavior, with different configurations achieving the lowest reported mean MAEs across diastolic and systolic targets.

\begin{table}[htbp]
\small
\centering
\resizebox{0.99\textwidth}{!}{%
\begin{tabular}{lccccc}
\toprule
\multirow{2}{*}{Task/Dataset (MAE)} & \multicolumn{3}{c}{MOMENT} & \multicolumn{2}{c}{Chronos-2} \\
\cmidrule(lr){2-4} \cmidrule(lr){5-6}
& 40M & 125M & 385M & 28M & 120M \\
\midrule
\multicolumn{6}{l}{\textbf{Potassium Estimation (MAE) $\downarrow$}} \\
MIMIC-III & 0.49 $\pm$ 0.004 & 0.48 $\pm$ 0.004 & 0.50 $\pm$ 0.004 & \textbf{0.46 $\pm$ 0.004} & \textbf{0.46 $\pm$ 0.004} \\
Inst. A & \underline{0.46 $\pm$ 0.01} & \underline{0.46 $\pm$ 0.01} & \textbf{0.45 $\pm$ 0.01} & 0.47 $\pm$ 0.01 & 0.48 $\pm$ 0.01 \\
\midrule
\multicolumn{6}{l}{\textbf{Sodium Estimation (MAE) $\downarrow$}} \\
MIMIC-III & 5.17 $\pm$ 0.03 & 5.84 $\pm$ 0.03 & 5.73 $\pm$ 0.03 & \underline{3.70 $\pm$ 0.03} & \textbf{3.68 $\pm$ 0.03} \\
Inst. A & 9.85 $\pm$ 0.06 & 7.55 $\pm$ 0.06 & 9.45 $\pm$ 0.06 & \underline{3.63 $\pm$ 0.06} & \textbf{3.59 $\pm$ 0.06} \\
\midrule
\multicolumn{6}{l}{\textbf{Glucose Estimation (MAE) $\downarrow$}} \\
MIMIC-III & 38.77 $\pm$ 0.39 & 38.19 $\pm$ 0.39 & 35.66 $\pm$ 0.40 & 36.46 $\pm$ 0.41 & 36.85 $\pm$ 0.40 \\
Inst. A & \underline{51.62 $\pm$ 0.66} & 52.22 $\pm$ 0.65 & 55.02 $\pm$ 0.71 & \textbf{51.17 $\pm$ 0.66} & 53.29 $\pm$ 0.71 \\
\midrule
\multicolumn{6}{l}{\textbf{A1c Estimation (MAE) $\downarrow$}} \\
Inst. A & 1.43 $\pm$ 0.02 & \underline{1.41 $\pm$ 0.02} & 1.43 $\pm$ 0.02 & 1.51 $\pm$ 0.03 & 1.48 $\pm$ 0.03 \\
\midrule
\multicolumn{6}{l}{\textbf{Troponin Estimation (MAE) $\downarrow$}} \\
MIMIC-III & 1.18 $\pm$ 0.02 & 1.25 $\pm$ 0.02 & \textbf{1.11 $\pm$ 0.02} & 1.27 $\pm$ 0.02 & 1.31 $\pm$ 0.02 \\
Inst. A & 6.26 $\pm$ 0.22 & \underline{6.03 $\pm$ 0.21} & \textbf{5.82 $\pm$ 0.21} & 8.26 $\pm$ 0.23 & 8.97 $\pm$ 0.22 \\
\midrule
\multicolumn{6}{l}{\textbf{Lactate Estimation (MAE) $\downarrow$}} \\
MIMIC-III & \textbf{1.15 $\pm$ 0.01} & 1.24 $\pm$ 0.01 & \textbf{1.15 $\pm$ 0.01} & 1.26 $\pm$ 0.01 & 1.29 $\pm$ 0.01 \\
Inst. A & 1.23 $\pm$ 0.01 & 1.24 $\pm$ 0.02 & 1.29 $\pm$ 0.02 & 1.36 $\pm$ 0.02 & 1.41 $\pm$ 0.02 \\
\midrule
\multicolumn{6}{l}{\textbf{ECG Metrics Estimation (MAE) $\downarrow$}} \\
PR interval (Inst. A) & 30.51 $\pm$ 0.26 & 29.81 $\pm$ 0.26 & 29.96 $\pm$ 0.25 & 26.64 $\pm$ 0.26 & 26.56 $\pm$ 0.25 \\
QRS duration (Inst. A) & 18.19 $\pm$ 0.13 & 17.72 $\pm$ 0.12 & 19.89 $\pm$ 0.12 & 16.63 $\pm$ 0.12 & 16.59 $\pm$ 0.12 \\
QT interval (Inst. A) & 55.44 $\pm$ 0.27 & 52.81 $\pm$ 0.26 & 52.96 $\pm$ 0.26 & 35.34 $\pm$ 0.26 & 35.45 $\pm$ 0.25 \\
\midrule
\multicolumn{6}{l}{\textbf{ECHO Metrics Estimation (MAE) $\downarrow$}} \\
LVEF (Inst. A) & 13.39 $\pm$ 0.22 & 13.14 $\pm$ 0.21 & 13.06 $\pm$ 0.21 & 13.72 $\pm$ 0.18 & 13.81 $\pm$ 0.18 \\
LVmass (Inst. A) & 32.83 $\pm$ 0.38 & 31.73 $\pm$ 0.38 & 31.71 $\pm$ 0.38 & \textbf{30.43 $\pm$ 0.39} & 30.53 $\pm$ 0.40 \\
\midrule
\multicolumn{6}{l}{\textbf{Cholesterol Estimation (HDL/LDL MAE) $\downarrow$}} \\
\multirow{2}{*}{Inst. A} & 14.42 $\pm$ 0.15 / & 13.90 $\pm$ 0.16 / & \textbf{13.13 $\pm$ 0.16} / & 14.65 $\pm$ 0.17 / & 14.90 $\pm$ 0.16 / \\
& 31.12 $\pm$ 0.37 & 30.83 $\pm$ 0.37 & 31.08 $\pm$ 0.37 & \textbf{29.92 $\pm$ 0.37} & \underline{30.35 $\pm$ 0.38} \\
\midrule
SCORE-reg-head-p2 & 0.5/18 & 0/18 & \textbf{4.5/18} & 3.17/18 & 2.17/18 \\
SCORE-reg-head-p2-comb & \multicolumn{3}{c}{5/18} & \multicolumn{2}{c}{\textbf{5.33/18}} \\
\midrule
SCORE-reg-head (p1+p2) & 1.5/45 & 1/45 & 7.5/45 & 6.17/45 & 4.17/45 \\
SCORE-reg-head-comb & \multicolumn{3}{c}{10/45} & \multicolumn{2}{c}{10.33/45} \\
\bottomrule
\end{tabular}
}
\caption{Detailed regression results for generalist FMs (MOMENT and Chronos-2) under head-only tuning on Group 2 tasks. Lower MAE is better. Bold and underline indicate the best and second-best reported mean performance across all 11 model configurations in this table and Table~\ref{tab:reg2_head_spec}. Win scores use this joint comparison, with equal reported mean MAEs sharing the win credit; combined scores sum wins across sizes within each model family. HDL and LDL are reported on separate lines and scored separately, yielding 18 Group 2 evaluations. The p1+p2 rows aggregate all 45 evaluations across Groups 1 and 2. Fractional scores are rounded to two decimal places. Uncertainty is the bootstrap standard error over 1000 test-set resamples.}
\label{tab:reg2_head_gen}
\end{table}

\begin{table}[htbp]
\small
\centering
\resizebox{0.99\textwidth}{!}{%
\begin{tabular}{lcccccc}
\toprule
\multirow{2}{*}{Task/Dataset (MAE)} & \multicolumn{3}{c}{GPT-PPG} & \multicolumn{3}{c}{MOMENT-PPG} \\
\cmidrule(lr){2-4} \cmidrule(lr){5-7}
& 19M & 85M & 345M & 40M & 125M & 385M \\
\midrule
\multicolumn{7}{l}{\textbf{Potassium Estimation (MAE) $\downarrow$}} \\
MIMIC-III & \underline{0.47 $\pm$ 0.00} & \underline{0.47 $\pm$ 0.00} & \textbf{0.46 $\pm$ 0.00} & \textbf{0.46 $\pm$ 0.00} & \textbf{0.46 $\pm$ 0.00} & \textbf{0.46 $\pm$ 0.00} \\
Inst. A & 0.57 $\pm$ 0.01 & 0.58 $\pm$ 0.01 & 0.48 $\pm$ 0.01 & 0.48 $\pm$ 0.01 & 0.50 $\pm$ 0.01 & 0.49 $\pm$ 0.01 \\
\midrule
\multicolumn{7}{l}{\textbf{Sodium Estimation (MAE) $\downarrow$}} \\
MIMIC-III & 4.80 $\pm$ 0.04 & 3.78 $\pm$ 0.03 & 3.74 $\pm$ 0.03 & 3.77 $\pm$ 0.04 & 3.78 $\pm$ 0.03 & 3.81 $\pm$ 0.03 \\
Inst. A & 8.24 $\pm$ 0.21 & 9.17 $\pm$ 0.26 & 10.69 $\pm$ 0.34 & 3.75 $\pm$ 0.07 & 4.25 $\pm$ 0.07 & 3.74 $\pm$ 0.06 \\
\midrule
\multicolumn{7}{l}{\textbf{Glucose Estimation (MAE) $\downarrow$}} \\
MIMIC-III & 35.19 $\pm$ 0.44 & \underline{35.05 $\pm$ 0.45} & \textbf{35.03 $\pm$ 0.45} & 35.54 $\pm$ 0.46 & 35.72 $\pm$ 0.44 & 35.66 $\pm$ 0.44 \\
Inst. A & 55.83 $\pm$ 0.81 & 57.11 $\pm$ 0.84 & 56.59 $\pm$ 0.84 & 56.99 $\pm$ 0.79 & 56.73 $\pm$ 0.78 & 59.42 $\pm$ 0.86 \\
\midrule
\multicolumn{7}{l}{\textbf{A1c Estimation (MAE) $\downarrow$}} \\
Inst. A & 1.47 $\pm$ 0.03 & 1.49 $\pm$ 0.03 & \textbf{1.33 $\pm$ 0.03} & 1.64 $\pm$ 0.02 & 1.66 $\pm$ 0.02 & 1.70 $\pm$ 0.02 \\
\midrule
\multicolumn{7}{l}{\textbf{Troponin Estimation (MAE) $\downarrow$}} \\
MIMIC-III & 1.20 $\pm$ 0.02 & \underline{1.17 $\pm$ 0.02} & 1.21 $\pm$ 0.02 & 1.46 $\pm$ 0.02 & 1.39 $\pm$ 0.02 & 1.38 $\pm$ 0.02 \\
Inst. A & 6.29 $\pm$ 0.26 & 6.54 $\pm$ 0.25 & 6.51 $\pm$ 0.25 & 7.10 $\pm$ 0.22 & 7.31 $\pm$ 0.23 & 7.59 $\pm$ 0.22 \\
\midrule
\multicolumn{7}{l}{\textbf{Lactate Estimation (MAE) $\downarrow$}} \\
MIMIC-III & 1.22 $\pm$ 0.01 & \underline{1.18 $\pm$ 0.02} & 1.20 $\pm$ 0.02 & 1.20 $\pm$ 0.02 & 1.19 $\pm$ 0.01 & 1.20 $\pm$ 0.01 \\
Inst. A & \textbf{1.12 $\pm$ 0.02} & \underline{1.15 $\pm$ 0.02} & \textbf{1.12 $\pm$ 0.02} & 1.20 $\pm$ 0.02 & 1.28 $\pm$ 0.02 & 1.37 $\pm$ 0.03 \\
\midrule
\multicolumn{7}{l}{\textbf{ECG Metrics Estimation (MAE) $\downarrow$}} \\
PR interval (Inst. A) & 28.35 $\pm$ 0.26 & 28.44 $\pm$ 0.27 & 28.60 $\pm$ 0.26 & \textbf{26.00 $\pm$ 0.25} & \underline{26.28 $\pm$ 0.25} & 27.04 $\pm$ 0.25 \\
QRS duration (Inst. A) & 17.43 $\pm$ 0.15 & 17.48 $\pm$ 0.15 & 17.38 $\pm$ 0.14 & \underline{16.32 $\pm$ 0.13} & \textbf{16.20 $\pm$ 0.13} & 16.66 $\pm$ 0.13 \\
QT interval (Inst. A) & 49.67 $\pm$ 0.33 & 50.79 $\pm$ 0.34 & 53.68 $\pm$ 0.35 & \underline{33.39 $\pm$ 0.23} & \textbf{32.75 $\pm$ 0.24} & 34.46 $\pm$ 0.24 \\
\midrule
\multicolumn{7}{l}{\textbf{ECHO Metrics Estimation (MAE) $\downarrow$}} \\
LVEF (Inst. A) & 14.11 $\pm$ 0.21 & 14.15 $\pm$ 0.23 & 14.52 $\pm$ 0.23 & \textbf{12.16 $\pm$ 0.19} & \underline{12.83 $\pm$ 0.20} & 13.32 $\pm$ 0.18 \\
LVmass (Inst. A) & 31.79 $\pm$ 0.46 & 32.01 $\pm$ 0.45 & 31.70 $\pm$ 0.45 & 31.63 $\pm$ 0.35 & \underline{30.46 $\pm$ 0.38} & 33.09 $\pm$ 0.33 \\
\midrule
\multicolumn{7}{l}{\textbf{Cholesterol Estimation (HDL/LDL MAE) $\downarrow$}} \\
\multirow{2}{*}{Inst. A} & \underline{13.79 $\pm$ 0.15} / & 14.04 $\pm$ 0.15 / & 13.93 $\pm$ 0.15 / & 15.07 $\pm$ 0.19 / & 15.14 $\pm$ 0.19 / & 16.04 $\pm$ 0.19 / \\
& 31.70 $\pm$ 0.38 & 31.62 $\pm$ 0.37 & 32.37 $\pm$ 0.39 & 31.47 $\pm$ 0.34 & 32.15 $\pm$ 0.36 & 31.30 $\pm$ 0.39 \\
\midrule
SCORE-reg-head-p2 & 0.5/18 & 0/18 & 2.67/18 & 2.17/18 & 2.17/18 & 0.17/18 \\
SCORE-reg-head-p2-comb & \multicolumn{3}{c}{3.17/18} & \multicolumn{3}{c}{4.5/18} \\
\midrule
SCORE-reg-head (p1+p2) & 0.5/45 & 3/45 & 2.67/45 & 4.17/45 & \textbf{10.67/45} & 3.67/45 \\
SCORE-reg-head-comb & \multicolumn{3}{c}{6.17/45} & \multicolumn{3}{c}{\textbf{18.5/45}} \\
\bottomrule
\end{tabular}
}
\caption{Detailed regression results for PPG-specialist FMs (GPT-PPG and MOMENT-PPG) under head-only tuning on Group 2 tasks. Lower MAE is better. Bold and underline indicate the best and second-best reported mean performance across all 11 model configurations in this table and Table~\ref{tab:reg2_head_gen}. Win scores use this joint comparison, with equal reported mean MAEs sharing the win credit; combined scores sum wins across sizes within each model family. HDL and LDL are reported on separate lines and scored separately, yielding 18 Group 2 evaluations. The p1+p2 rows aggregate all 45 evaluations across Groups 1 and 2. Fractional scores are rounded to two decimal places. Uncertainty is the bootstrap standard error over 1000 test-set resamples.}
\label{tab:reg2_head_spec}
\end{table}

Tables~\ref{tab:reg2_head_gen} and~\ref{tab:reg2_head_spec} summarize head-only tuning results for laboratory, ECG, ECHO, and cholesterol regression tasks.
The combined mean-based win scores in Group 2 are 5.33/18 for Chronos-2, 5/18 for MOMENT, 4.5/18 for MOMENT-PPG, and 3.17/18 for GPT-PPG.
GPT-PPG remains competitive on MIMIC-III sodium and glucose, where its best configuration is tied with the best configuration from another family under the 95\% interval-overlap criterion.
On QT interval estimation, the best Chronos-2 and MOMENT-PPG configurations achieve lower MAE than the best GPT-PPG configuration, with non-overlapping 95\% intervals.
These results suggest that the relative performance of the four families depends on the target and dataset.

\subsection{Full-model tuning on regression tasks}

\begin{table}[htbp]
\small
\centering
\resizebox{0.99\textwidth}{!}{%
\begin{tabular}{lccccc}
\toprule
\multirow{2}{*}{Task/Dataset (MAE)} & \multicolumn{3}{c}{MOMENT} & \multicolumn{2}{c}{Chronos-2} \\
\cmidrule(lr){2-4} \cmidrule(lr){5-6}
& 40M & 125M & 385M & 28M & 120M \\
\midrule
\multicolumn{6}{l}{\textbf{SpO$_2$ Estimation (MAE) $\downarrow$}} \\
Sleep Disorder & 3.94 $\pm$ 0.10 & 3.44 $\pm$ 0.14 & 4.28 $\pm$ 0.20 & 3.77 $\pm$ 0.05 & 3.79 $\pm$ 0.05 \\
BIDMC & 2.54 $\pm$ 0.07 & 2.56 $\pm$ 0.09 & 2.57 $\pm$ 0.06 & 1.22 $\pm$ 0.05 & \underline{1.15 $\pm$ 0.04} \\
MIMIC-III & 2.25 $\pm$ 0.02 & 2.25 $\pm$ 0.02 & 2.24 $\pm$ 0.02 & \textbf{2.20 $\pm$ 0.02} & \underline{2.21 $\pm$ 0.02} \\
\midrule
\multicolumn{6}{l}{\textbf{Skin Temperature Estimation (MAE) $\downarrow$}} \\
MIMIC-III & 2.52 $\pm$ 0.01 & 2.02 $\pm$ 0.01 & 1.98 $\pm$ 0.01 & 1.78 $\pm$ 0.01 & 1.97 $\pm$ 0.01 \\
Pulse Transit & 1.28 $\pm$ 0.04 & 1.28 $\pm$ 0.05 & 1.28 $\pm$ 0.05 & 1.06 $\pm$ 0.04 & \textbf{0.64 $\pm$ 0.03} \\
\midrule
\multicolumn{6}{l}{\textbf{Heart Rate Estimation (MAE) $\downarrow$}} \\
WESAD & 10.87 $\pm$ 0.09 & 10.86 $\pm$ 0.08 & 10.86 $\pm$ 0.09 & 9.78 $\pm$ 0.07 & \underline{9.60 $\pm$ 0.07} \\
DALIA & 12.98 $\pm$ 0.11 & 11.67 $\pm$ 0.11 & 10.84 $\pm$ 0.10 & 9.30 $\pm$ 0.11 & 9.03 $\pm$ 0.10 \\
BIDMC & 9.83 $\pm$ 0.31 & 9.57 $\pm$ 0.18 & 4.55 $\pm$ 0.16 & 2.98 $\pm$ 0.13 & 2.05 $\pm$ 0.11 \\
Gyro-Acc-PPG & 20.79 $\pm$ 0.61 & 20.72 $\pm$ 0.59 & 20.48 $\pm$ 0.58 & 20.21 $\pm$ 0.62 & \textbf{19.89 $\pm$ 0.59} \\
Welltory-PPG & 7.30 $\pm$ 0.49 & 7.25 $\pm$ 0.46 & 7.18 $\pm$ 0.48 & \underline{5.66 $\pm$ 0.46} & \textbf{3.90 $\pm$ 0.43} \\
BUT-PPG & 10.68 $\pm$ 0.27 & 10.77 $\pm$ 0.23 & 10.78 $\pm$ 0.29 & 10.49 $\pm$ 0.29 & \textbf{10.05 $\pm$ 0.29} \\
\midrule
\multicolumn{6}{l}{\textbf{Respiration Rate Estimation (MAE) $\downarrow$}} \\
BIDMC & 2.09 $\pm$ 0.06 & 2.09 $\pm$ 0.06 & 2.09 $\pm$ 0.06 & 1.57 $\pm$ 0.05 & 1.44 $\pm$ 0.05 \\
MIMIC-III & 4.64 $\pm$ 0.04 & 4.63 $\pm$ 0.04 & 4.63 $\pm$ 0.04 & 4.45 $\pm$ 0.04 & \textbf{4.17 $\pm$ 0.04} \\
\midrule
\multicolumn{6}{l}{\textbf{Blood Pressure Estimation (Diastolic/Systolic MAE) $\downarrow$}} \\
\multirow{2}{*}{PPG-BP} & 9.17 $\pm$ 0.64 / & 9.14 $\pm$ 0.63 / & 9.16 $\pm$ 0.62 / & 9.20 $\pm$ 0.64 / & \textbf{8.50 $\pm$ 0.69} / \\
& 17.07 $\pm$ 1.11 & 17.10 $\pm$ 1.12 & 16.99 $\pm$ 1.12 & 17.04 $\pm$ 1.11 & \underline{15.36 $\pm$ 1.16} \\
\multirow{2}{*}{Aurora-Oscillometric} & 21.11 $\pm$ 0.28 / & 22.19 $\pm$ 0.30 / & 23.82 $\pm$ 0.30 / & 8.54 $\pm$ 0.24 / & \textbf{7.02 $\pm$ 0.21} / \\
& 25.11 $\pm$ 0.35 & 25.12 $\pm$ 0.34 & 25.11 $\pm$ 0.35 & 20.67 $\pm$ 0.34 & \textbf{19.13 $\pm$ 0.32} \\
\multirow{2}{*}{Aurora-Auscultatory} & 13.79 $\pm$ 0.39 / & \textbf{9.70 $\pm$ 0.38} / & 9.79 $\pm$ 0.39 / & 9.86 $\pm$ 0.40 / & 10.07 $\pm$ 0.34 / \\
& 13.82 $\pm$ 0.49 & 13.77 $\pm$ 0.50 & 13.83 $\pm$ 0.49 & 13.37 $\pm$ 0.49 & \textbf{13.25 $\pm$ 0.47} \\
\multirow{2}{*}{CAS-BP} & \textbf{8.39 $\pm$ 0.10} / & \textbf{8.39 $\pm$ 0.10} / & \underline{8.42 $\pm$ 0.11} / & 8.98 $\pm$ 0.11 / & 8.76 $\pm$ 0.11 / \\
& 14.39 $\pm$ 0.15 & 14.30 $\pm$ 0.15 & 14.30 $\pm$ 0.15 & 13.98 $\pm$ 0.17 & 13.37 $\pm$ 0.16 \\
\multirow{2}{*}{Vital Videos} & 7.95 $\pm$ 0.83 / & 7.97 $\pm$ 0.79 / & 7.86 $\pm$ 0.80 / & 7.87 $\pm$ 0.78 / & 7.92 $\pm$ 0.78 / \\
& \underline{18.19 $\pm$ 2.20} & \underline{18.19 $\pm$ 2.20} & 18.24 $\pm$ 2.21 & \textbf{18.18 $\pm$ 2.24} & 19.14 $\pm$ 2.21 \\
\multirow{2}{*}{BCG} & \underline{5.86 $\pm$ 0.56} / & \underline{5.86 $\pm$ 0.39} / & \underline{5.86 $\pm$ 0.56} / & 8.04 $\pm$ 0.94 / & 8.27 $\pm$ 0.76 / \\
& 4.46 $\pm$ 0.64 & 4.46 $\pm$ 0.63 & 4.60 $\pm$ 0.41 & 8.62 $\pm$ 0.49 & 7.00 $\pm$ 0.61 \\
\multirow{2}{*}{BUT-PPG} & 8.46 $\pm$ 0.13 / & 8.93 $\pm$ 0.23 / & 8.46 $\pm$ 0.17 / & \textbf{7.59 $\pm$ 0.13} / & 7.74 $\pm$ 0.13 / \\
& 33.00 $\pm$ 0.31 & 17.39 $\pm$ 0.30 & 16.61 $\pm$ 0.35 & \underline{14.32 $\pm$ 0.28} & \textbf{14.01 $\pm$ 0.28} \\
\midrule
SCORE-reg-full-p1 & 0.5/27 & 1.5/27 & 0/27 & 2.25/27 & \textbf{10/27} \\
SCORE-reg-full-p1-comb & \multicolumn{3}{c}{2/27} & \multicolumn{2}{c}{\textbf{12.25/27}} \\
\bottomrule
\end{tabular}
}
\caption{Detailed regression results for generalist FMs (MOMENT and Chronos-2) under full-model tuning on Group 1 tasks. Lower MAE is better. Bold and underline indicate the best and second-best distinct reported means, respectively, across all 11 configurations in this table and Table~\ref{tab:reg1_full_spec}, including ties. SCORE counts wins over 27 evaluations, splitting win credit equally among configurations with equal reported mean MAEs; combined scores sum wins across sizes within each model family. Diastolic and systolic MAE and their standard errors are reported on separate lines and scored separately. Uncertainty is the bootstrap standard error over 1000 test-set resamples.}
\label{tab:reg1_full_gen}
\end{table}

\begin{table}[htbp]
\small
\centering
\resizebox{0.99\textwidth}{!}{%
\begin{tabular}{lcccccc}
\toprule
\multirow{2}{*}{Task/Dataset (MAE)} & \multicolumn{3}{c}{GPT-PPG} & \multicolumn{3}{c}{MOMENT-PPG} \\
\cmidrule(lr){2-4} \cmidrule(lr){5-7}
& 19M & 85M & 345M & 40M & 125M & 385M \\
\midrule
\multicolumn{7}{l}{\textbf{SpO$_2$ Estimation (MAE) $\downarrow$}} \\
Sleep Disorder & 3.82 $\pm$ 0.08 & 3.88 $\pm$ 0.08 & 3.90 $\pm$ 0.08 & \underline{3.27 $\pm$ 0.07} & 3.28 $\pm$ 0.07 & \textbf{3.20 $\pm$ 0.06} \\
BIDMC & 1.48 $\pm$ 0.05 & \textbf{1.04 $\pm$ 0.04} & 1.18 $\pm$ 0.04 & 1.62 $\pm$ 0.05 & 1.46 $\pm$ 0.05 & \underline{1.15 $\pm$ 0.04} \\
MIMIC-III & 2.46 $\pm$ 0.03 & 2.39 $\pm$ 0.03 & 2.43 $\pm$ 0.03 & \textbf{2.20 $\pm$ 0.03} & \textbf{2.20 $\pm$ 0.03} & \textbf{2.20 $\pm$ 0.03} \\
\midrule
\multicolumn{7}{l}{\textbf{Skin Temperature Estimation (MAE) $\downarrow$}} \\
MIMIC-III & \textbf{1.17 $\pm$ 0.01} & \underline{1.18 $\pm$ 0.01} & 1.19 $\pm$ 0.01 & 1.58 $\pm$ 0.01 & 1.76 $\pm$ 0.01 & 2.14 $\pm$ 0.01 \\
Pulse Transit & \underline{0.90 $\pm$ 0.03} & 0.91 $\pm$ 0.04 & 0.98 $\pm$ 0.05 & \underline{0.90 $\pm$ 0.04} & 0.99 $\pm$ 0.04 & 0.93 $\pm$ 0.04 \\
\midrule
\multicolumn{7}{l}{\textbf{Heart Rate Estimation (MAE) $\downarrow$}} \\
WESAD & 10.84 $\pm$ 0.08 & 10.51 $\pm$ 0.08 & 10.41 $\pm$ 0.07 & \textbf{9.54 $\pm$ 0.07} & 9.71 $\pm$ 0.08 & 10.19 $\pm$ 0.08 \\
DALIA & 9.39 $\pm$ 0.09 & 8.03 $\pm$ 0.09 & 8.18 $\pm$ 0.09 & \underline{7.01 $\pm$ 0.08} & 7.16 $\pm$ 0.08 & \textbf{6.29 $\pm$ 0.08} \\
BIDMC & 1.97 $\pm$ 0.11 & 1.93 $\pm$ 0.09 & 2.07 $\pm$ 0.12 & 2.42 $\pm$ 0.13 & \underline{1.66 $\pm$ 0.10} & \textbf{1.54 $\pm$ 0.08} \\
Gyro-Acc-PPG & 20.66 $\pm$ 0.62 & 20.65 $\pm$ 0.65 & 21.13 $\pm$ 0.61 & 20.23 $\pm$ 0.63 & \underline{20.20 $\pm$ 0.60} & 20.40 $\pm$ 0.60 \\
Welltory-PPG & 8.56 $\pm$ 1.68 & 8.21 $\pm$ 1.71 & 6.46 $\pm$ 1.19 & 6.03 $\pm$ 0.56 & 8.80 $\pm$ 0.58 & 7.00 $\pm$ 0.58 \\
BUT-PPG & 11.44 $\pm$ 0.30 & 12.28 $\pm$ 0.29 & \underline{10.22 $\pm$ 0.28} & 10.74 $\pm$ 0.28 & 10.46 $\pm$ 0.27 & 10.99 $\pm$ 0.30 \\
\midrule
\multicolumn{7}{l}{\textbf{Respiration Rate Estimation (MAE) $\downarrow$}} \\
BIDMC & 1.30 $\pm$ 0.05 & \underline{1.29 $\pm$ 0.04} & 1.42 $\pm$ 0.06 & 1.40 $\pm$ 0.05 & \textbf{1.08 $\pm$ 0.04} & \textbf{1.08 $\pm$ 0.04} \\
MIMIC-III & 4.50 $\pm$ 0.04 & 4.47 $\pm$ 0.04 & 4.57 $\pm$ 0.04 & 4.38 $\pm$ 0.04 & \underline{4.27 $\pm$ 0.04} & 4.33 $\pm$ 0.04 \\
\midrule
\multicolumn{7}{l}{\textbf{Blood Pressure Estimation (Diastolic/Systolic MAE) $\downarrow$}} \\
\multirow{2}{*}{PPG-BP} & 9.60 $\pm$ 0.63 / & 10.24 $\pm$ 0.67 / & 9.66 $\pm$ 0.61 / & \underline{9.04 $\pm$ 0.66} / & 10.06 $\pm$ 0.66 / & 9.28 $\pm$ 0.60 / \\
& \textbf{15.31 $\pm$ 1.28} & 16.56 $\pm$ 1.30 & 15.62 $\pm$ 1.12 & 16.71 $\pm$ 1.17 & 16.69 $\pm$ 1.14 & 16.49 $\pm$ 1.16 \\
\multirow{2}{*}{Aurora-Oscillometric} & 7.96 $\pm$ 0.23 / & 7.93 $\pm$ 0.23 / & \underline{7.07 $\pm$ 0.24} / & 8.14 $\pm$ 0.22 / & 7.13 $\pm$ 0.23 / & 7.17 $\pm$ 0.24 / \\
& 20.50 $\pm$ 0.35 & 20.76 $\pm$ 0.34 & 20.42 $\pm$ 0.35 & 20.22 $\pm$ 0.35 & \underline{19.76 $\pm$ 0.32} & 20.34 $\pm$ 0.33 \\
\multirow{2}{*}{Aurora-Auscultatory} & \underline{9.74 $\pm$ 0.43} / & 9.75 $\pm$ 0.42 / & 9.75 $\pm$ 0.41 / & 10.00 $\pm$ 0.44 / & 9.78 $\pm$ 0.43 / & 10.13 $\pm$ 0.42 / \\
& 13.77 $\pm$ 0.54 & 13.87 $\pm$ 0.51 & 13.85 $\pm$ 0.54 & 13.37 $\pm$ 0.50 & \underline{13.34 $\pm$ 0.52} & 13.77 $\pm$ 0.54 \\
\multirow{2}{*}{CAS-BP} & 8.49 $\pm$ 0.10 / & 8.50 $\pm$ 0.10 / & 8.48 $\pm$ 0.10 / & 8.87 $\pm$ 0.10 / & 8.85 $\pm$ 0.11 / & 8.94 $\pm$ 0.11 / \\
& 14.35 $\pm$ 0.15 & 14.36 $\pm$ 0.15 & 14.37 $\pm$ 0.15 & 13.50 $\pm$ 0.16 & \textbf{12.61 $\pm$ 0.15} & \underline{13.15 $\pm$ 0.15} \\
\multirow{2}{*}{Vital Videos} & 8.01 $\pm$ 0.80 / & 7.93 $\pm$ 0.82 / & \underline{7.85 $\pm$ 0.80} / & \textbf{7.84 $\pm$ 0.84} / & 8.39 $\pm$ 0.87 / & 8.50 $\pm$ 0.90 / \\
& 39.90 $\pm$ 3.27 & 18.36 $\pm$ 2.27 & 18.36 $\pm$ 2.19 & 18.73 $\pm$ 2.24 & 22.03 $\pm$ 2.24 & 18.61 $\pm$ 2.22 \\
\multirow{2}{*}{BCG} & \textbf{5.73 $\pm$ 0.62} / & 6.17 $\pm$ 0.72 / & 6.16 $\pm$ 0.82 / & 6.02 $\pm$ 0.67 / & 6.16 $\pm$ 0.57 / & 7.40 $\pm$ 0.85 / \\
& \underline{4.34 $\pm$ 0.11} & 5.81 $\pm$ 0.57 & 4.71 $\pm$ 0.46 & \textbf{4.05 $\pm$ 0.34} & 5.37 $\pm$ 0.48 & 5.93 $\pm$ 0.45 \\
\multirow{2}{*}{BUT-PPG} & 7.85 $\pm$ 0.15 / & \underline{7.71 $\pm$ 0.16} / & 7.80 $\pm$ 0.13 / & 8.96 $\pm$ 0.17 / & 8.83 $\pm$ 0.18 / & 9.75 $\pm$ 0.22 / \\
& 15.50 $\pm$ 0.32 & 15.71 $\pm$ 0.32 & 15.43 $\pm$ 0.34 & 14.84 $\pm$ 0.32 & 14.83 $\pm$ 0.32 & 16.76 $\pm$ 0.39 \\
\midrule
SCORE-reg-full-p1 & 3/27 & 1/27 & 0/27 & 3.25/27 & 1.75/27 & 3.75/27 \\
SCORE-reg-full-p1-comb & \multicolumn{3}{c}{4/27} & \multicolumn{3}{c}{8.75/27} \\
\bottomrule
\end{tabular}
}
\caption{Detailed regression results for PPG-specialist FMs (GPT-PPG and MOMENT-PPG) under full-model tuning on Group 1 tasks. Lower MAE is better. Bold and underline indicate the best and second-best distinct reported means, respectively, across all 11 configurations in this table and Table~\ref{tab:reg1_full_gen}, including ties. SCORE counts wins over 27 evaluations, splitting win credit equally among configurations with equal reported mean MAEs; combined scores sum wins across sizes within each model family. Diastolic and systolic MAE and their standard errors are reported on separate lines and scored separately. Uncertainty is the bootstrap standard error over 1000 test-set resamples.}
\label{tab:reg1_full_spec}
\end{table}


Tables~\ref{tab:reg1_full_gen} and~\ref{tab:reg1_full_spec} show that full-model tuning improves GPT-PPG on several Group 1 tasks, including selected heart-rate, respiration-rate, and blood-pressure settings.
For example, GPT-PPG improves substantially on some blood-pressure datasets relative to its head-only tuning behavior.
MOMENT remains competitive on several diastolic blood-pressure and heart-rate settings, but its full-tuning gains are less consistent across datasets.

\begin{table}[htbp]
\small
\centering
\resizebox{0.99\textwidth}{!}{%
\begin{tabular}{lccccc}
\toprule
\multirow{2}{*}{Task/Dataset (MAE)} & \multicolumn{3}{c}{MOMENT} & \multicolumn{2}{c}{Chronos-2} \\
\cmidrule(lr){2-4} \cmidrule(lr){5-6}
& 40M & 125M & 385M & 28M & 120M \\
\midrule
\multicolumn{6}{l}{\textbf{Potassium Estimation (MAE) $\downarrow$}} \\
MIMIC-III & \underline{0.47 $\pm$ 0.004} & \underline{0.47 $\pm$ 0.004} & \underline{0.47 $\pm$ 0.004} & \underline{0.47 $\pm$ 0.004} & \underline{0.47 $\pm$ 0.004} \\
Inst. A & \textbf{0.45 $\pm$ 0.01} & \textbf{0.45 $\pm$ 0.01} & \textbf{0.45 $\pm$ 0.01} & \underline{0.47 $\pm$ 0.01} & 0.52 $\pm$ 0.01 \\
\midrule
\multicolumn{6}{l}{\textbf{Sodium Estimation (MAE) $\downarrow$}} \\
MIMIC-III & \textbf{3.63 $\pm$ 0.03} & \textbf{3.63 $\pm$ 0.03} & \textbf{3.63 $\pm$ 0.03} & \underline{3.66 $\pm$ 0.03} & 3.67 $\pm$ 0.03 \\
Inst. A & 3.99 $\pm$ 0.06 & 3.90 $\pm$ 0.05 & \underline{3.84 $\pm$ 0.06} & 4.30 $\pm$ 0.06 & 4.68 $\pm$ 0.07 \\
\midrule
\multicolumn{6}{l}{\textbf{Glucose Estimation (MAE) $\downarrow$}} \\
MIMIC-III & 36.19 $\pm$ 0.41 & 36.13 $\pm$ 0.39 & 35.72 $\pm$ 0.41 & \textbf{35.00 $\pm$ 0.43} & 35.64 $\pm$ 0.41 \\
Inst. A & \underline{51.88 $\pm$ 0.76} & 52.15 $\pm$ 0.77 & 51.89 $\pm$ 0.72 & \textbf{51.40 $\pm$ 0.69} & 57.47 $\pm$ 0.82 \\
\midrule
\multicolumn{6}{l}{\textbf{A1c Estimation (MAE) $\downarrow$}} \\
Inst. A & 1.37 $\pm$ 0.03 & 1.35 $\pm$ 0.03 & 1.34 $\pm$ 0.03 & 1.53 $\pm$ 0.02 & 1.77 $\pm$ 0.03 \\
\midrule
\multicolumn{6}{l}{\textbf{Troponin Estimation (MAE) $\downarrow$}} \\
MIMIC-III & 1.27 $\pm$ 0.02 & 1.26 $\pm$ 0.02 & 1.29 $\pm$ 0.02 & 1.28 $\pm$ 0.02 & 1.26 $\pm$ 0.02 \\
Inst. A & \underline{6.50 $\pm$ 0.32} & 6.59 $\pm$ 0.25 & \textbf{6.49 $\pm$ 0.24} & 8.24 $\pm$ 0.25 & 8.23 $\pm$ 0.26 \\
\midrule
\multicolumn{6}{l}{\textbf{Lactate Estimation (MAE) $\downarrow$}} \\
MIMIC-III & 1.21 $\pm$ 0.01 & \textbf{1.18 $\pm$ 0.01} & 1.24 $\pm$ 0.02 & 1.23 $\pm$ 0.02 & 1.25 $\pm$ 0.01 \\
Inst. A & 1.36 $\pm$ 0.03 & 1.34 $\pm$ 0.04 & 1.29 $\pm$ 0.03 & 1.36 $\pm$ 0.02 & 1.50 $\pm$ 0.02 \\
\midrule
\multicolumn{6}{l}{\textbf{ECG Metrics Estimation (MAE) $\downarrow$}} \\
PR interval (Inst. A) & 28.45 $\pm$ 0.26 & 28.43 $\pm$ 0.27 & 28.38 $\pm$ 0.28 & 27.85 $\pm$ 0.26 & 27.98 $\pm$ 0.26 \\
QRS duration (Inst. A) & 17.19 $\pm$ 0.14 & 17.04 $\pm$ 0.15 & 17.23 $\pm$ 0.16 & 16.79 $\pm$ 0.14 & \underline{16.56 $\pm$ 0.14} \\
QT interval (Inst. A) & 50.62 $\pm$ 0.34 & 50.62 $\pm$ 0.33 & 50.62 $\pm$ 0.33 & \underline{33.68 $\pm$ 0.25} & 33.72 $\pm$ 0.24 \\
\midrule
\multicolumn{6}{l}{\textbf{ECHO Metrics Estimation (MAE) $\downarrow$}} \\
LVEF (Inst. A) & 14.13 $\pm$ 0.25 & 14.42 $\pm$ 0.24 & 13.98 $\pm$ 0.25 & 13.53 $\pm$ 0.18 & 13.42 $\pm$ 0.19 \\
LVmass (Inst. A) & 31.61 $\pm$ 0.40 & 31.61 $\pm$ 0.42 & 31.61 $\pm$ 0.38 & 32.69 $\pm$ 0.45 & \underline{31.27 $\pm$ 0.38} \\
\midrule
\multicolumn{6}{l}{\textbf{Cholesterol Estimation (HDL/LDL MAE) $\downarrow$}} \\
\multirow{2}{*}{Inst. A} & \underline{12.27 $\pm$ 0.18} / & \textbf{12.24 $\pm$ 0.19} / & 13.38 $\pm$ 0.19 / & 13.08 $\pm$ 0.16 / & 13.66 $\pm$ 0.20 / \\
& \textbf{29.72 $\pm$ 0.38} & 29.83 $\pm$ 0.43 & \underline{29.79 $\pm$ 0.48} & 31.80 $\pm$ 0.39 & 32.77 $\pm$ 0.41 \\
\midrule
SCORE-reg-full-p2 & 1.67/18 & 2.67/18 & 1.67/18 & 2/18 & 0/18 \\
SCORE-reg-full-p2-comb & \multicolumn{3}{c}{6/18} & \multicolumn{2}{c}{2/18} \\
\midrule
SCORE-reg-full (p1+p2) & 2.17/45 & 4.17/45 & 1.67/45 & 4.25/45 & \textbf{10/45} \\
SCORE-reg-full-comb & \multicolumn{3}{c}{8/45} & \multicolumn{2}{c}{14.25/45} \\
\bottomrule
\end{tabular}
}
\caption{Detailed regression results for generalist FMs (MOMENT and Chronos-2) under full-model tuning on Group 2 tasks. Lower MAE is better. Bold and underline indicate the best and second-best distinct reported means, respectively, across all 11 configurations in this table and Table~\ref{tab:reg2_full_spec}, including ties. Win scores use this joint comparison, with equal reported mean MAEs sharing the win credit; combined scores sum wins across sizes within each model family. HDL and LDL MAE and their standard errors are reported on separate lines and scored separately, yielding 18 Group 2 evaluations. The p1+p2 rows aggregate all 45 evaluations across Groups 1 and 2. Fractional scores are rounded to two decimal places. Uncertainty is the bootstrap standard error over 1000 test-set resamples.}
\label{tab:reg2_full_gen}
\end{table}

\begin{table}[htbp]
\small
\centering
\resizebox{0.99\textwidth}{!}{%
\begin{tabular}{lcccccc}
\toprule
\multirow{2}{*}{Task/Dataset (MAE)} & \multicolumn{3}{c}{GPT-PPG} & \multicolumn{3}{c}{MOMENT-PPG} \\
\cmidrule(lr){2-4} \cmidrule(lr){5-7}
& 19M & 85M & 345M & 40M & 125M & 385M \\
\midrule
\multicolumn{7}{l}{\textbf{Potassium Estimation (MAE) $\downarrow$}} \\
MIMIC-III & \underline{0.47 $\pm$ 0.00} & \textbf{0.46 $\pm$ 0.00} & 0.49 $\pm$ 0.00 & \textbf{0.46 $\pm$ 0.00} & \textbf{0.46 $\pm$ 0.00} & \textbf{0.46 $\pm$ 0.00} \\
Inst. A & 0.49 $\pm$ 0.01 & 0.48 $\pm$ 0.01 & 0.53 $\pm$ 0.01 & \underline{0.47 $\pm$ 0.01} & 0.58 $\pm$ 0.01 & 0.64 $\pm$ 0.01 \\
\midrule
\multicolumn{7}{l}{\textbf{Sodium Estimation (MAE) $\downarrow$}} \\
MIMIC-III & 4.08 $\pm$ 0.03 & 4.08 $\pm$ 0.04 & 3.98 $\pm$ 0.04 & 3.75 $\pm$ 0.04 & 3.75 $\pm$ 0.03 & 3.74 $\pm$ 0.04 \\
Inst. A & 6.65 $\pm$ 0.19 & 7.02 $\pm$ 0.21 & 4.72 $\pm$ 0.08 & \textbf{3.59 $\pm$ 0.06} & 6.06 $\pm$ 0.09 & 6.49 $\pm$ 0.08 \\
\midrule
\multicolumn{7}{l}{\textbf{Glucose Estimation (MAE) $\downarrow$}} \\
MIMIC-III & 35.52 $\pm$ 0.44 & 35.65 $\pm$ 0.43 & 35.42 $\pm$ 0.44 & 35.26 $\pm$ 0.44 & 35.29 $\pm$ 0.45 & \underline{35.24 $\pm$ 0.45} \\
Inst. A & 59.15 $\pm$ 0.85 & 57.60 $\pm$ 0.77 & 54.13 $\pm$ 0.76 & 54.92 $\pm$ 0.83 & 60.65 $\pm$ 0.83 & 63.18 $\pm$ 0.83 \\
\midrule
\multicolumn{7}{l}{\textbf{A1c Estimation (MAE) $\downarrow$}} \\
Inst. A & 1.38 $\pm$ 0.03 & \underline{1.33 $\pm$ 0.03} & \textbf{1.30 $\pm$ 0.03} & 1.65 $\pm$ 0.03 & 1.61 $\pm$ 0.03 & 1.64 $\pm$ 0.03 \\
\midrule
\multicolumn{7}{l}{\textbf{Troponin Estimation (MAE) $\downarrow$}} \\
MIMIC-III & 1.36 $\pm$ 0.02 & 1.35 $\pm$ 0.02 & \textbf{1.22 $\pm$ 0.02} & 1.38 $\pm$ 0.02 & 1.38 $\pm$ 0.02 & \underline{1.23 $\pm$ 0.02} \\
Inst. A & 6.82 $\pm$ 0.25 & 6.83 $\pm$ 0.25 & 7.33 $\pm$ 0.24 & 7.74 $\pm$ 0.24 & 8.01 $\pm$ 0.22 & 6.68 $\pm$ 0.22 \\
\midrule
\multicolumn{7}{l}{\textbf{Lactate Estimation (MAE) $\downarrow$}} \\
MIMIC-III & 1.23 $\pm$ 0.01 & \underline{1.19 $\pm$ 0.01} & \underline{1.19 $\pm$ 0.02} & 1.26 $\pm$ 0.01 & 1.24 $\pm$ 0.01 & 1.21 $\pm$ 0.01 \\
Inst. A & \textbf{1.09 $\pm$ 0.02} & 1.12 $\pm$ 0.02 & \underline{1.10 $\pm$ 0.02} & 1.30 $\pm$ 0.02 & 1.54 $\pm$ 0.03 & 1.68 $\pm$ 0.03 \\
\midrule
\multicolumn{7}{l}{\textbf{ECG Metrics Estimation (MAE) $\downarrow$}} \\
PR interval (Inst. A) & \underline{27.33 $\pm$ 0.27} & 27.47 $\pm$ 0.26 & 28.34 $\pm$ 0.25 & \textbf{26.56 $\pm$ 0.25} & 30.11 $\pm$ 0.28 & 33.93 $\pm$ 0.31 \\
QRS duration (Inst. A) & 17.12 $\pm$ 0.15 & 17.14 $\pm$ 0.15 & 17.25 $\pm$ 0.15 & \textbf{16.09 $\pm$ 0.14} & 21.00 $\pm$ 0.15 & 21.75 $\pm$ 0.18 \\
QT interval (Inst. A) & 35.58 $\pm$ 0.28 & 34.07 $\pm$ 0.25 & 35.80 $\pm$ 0.27 & \textbf{33.23 $\pm$ 0.23} & 42.32 $\pm$ 0.31 & 37.19 $\pm$ 0.26 \\
\midrule
\multicolumn{7}{l}{\textbf{ECHO Metrics Estimation (MAE) $\downarrow$}} \\
LVEF (Inst. A) & 14.50 $\pm$ 0.20 & 14.17 $\pm$ 0.22 & 14.61 $\pm$ 0.20 & \underline{12.31 $\pm$ 0.19} & \textbf{11.44 $\pm$ 0.19} & 14.81 $\pm$ 0.21 \\
LVmass (Inst. A) & 31.33 $\pm$ 0.47 & 31.57 $\pm$ 0.45 & 31.52 $\pm$ 0.44 & \textbf{30.32 $\pm$ 0.51} & 31.82 $\pm$ 0.41 & 32.53 $\pm$ 0.46 \\
\midrule
\multicolumn{7}{l}{\textbf{Cholesterol Estimation (HDL/LDL MAE) $\downarrow$}} \\
\multirow{2}{*}{Inst. A} & 13.44 $\pm$ 0.16 / & 13.52 $\pm$ 0.15 / & 13.43 $\pm$ 0.15 / & 13.78 $\pm$ 0.16 / & 15.23 $\pm$ 0.20 / & 17.59 $\pm$ 0.22 / \\
& 31.55 $\pm$ 0.38 & 33.11 $\pm$ 0.37 & 31.97 $\pm$ 0.37 & 31.58 $\pm$ 0.39 & 33.01 $\pm$ 0.40 & 31.47 $\pm$ 0.43 \\
\midrule
SCORE-reg-full-p2 & 1/18 & 0.25/18 & 2/18 & \textbf{5.25/18} & 1.25/18 & 0.25/18 \\
SCORE-reg-full-p2-comb & \multicolumn{3}{c}{3.25/18} & \multicolumn{3}{c}{\textbf{6.75/18}} \\
\midrule
SCORE-reg-full (p1+p2) & 4/45 & 1.25/45 & 2/45 & 8.5/45 & 3/45 & 4/45 \\
SCORE-reg-full-comb & \multicolumn{3}{c}{7.25/45} & \multicolumn{3}{c}{\textbf{15.5/45}} \\
\bottomrule
\end{tabular}
}
\caption{Detailed regression results for PPG-specialist FMs (GPT-PPG and MOMENT-PPG) under full-model tuning on Group 2 tasks. Lower MAE is better. Bold and underline indicate the best and second-best distinct reported means, respectively, across all 11 configurations in this table and Table~\ref{tab:reg2_full_gen}, including ties. Win scores use this joint comparison, with equal reported mean MAEs sharing the win credit; combined scores sum wins across sizes within each model family. HDL and LDL MAE and their standard errors are reported on separate lines and scored separately, yielding 18 Group 2 evaluations. The p1+p2 rows aggregate all 45 evaluations across Groups 1 and 2. Fractional scores are rounded to two decimal places. Uncertainty is the bootstrap standard error over 1000 test-set resamples.}
\label{tab:reg2_full_spec}
\end{table}

Tables~\ref{tab:reg2_full_gen} and~\ref{tab:reg2_full_spec} show that full-model tuning produces mixed behavior on laboratory and cross-modal clinical regression tasks.
On LVEF, MOMENT-PPG-125M achieves lower MAE than the best configuration from every other family, with non-overlapping 95\% intervals.
For LVmass, the best configurations from all four families are tied under the 95\% interval-overlap criterion.
Several biomarker comparisons have overlapping 95\% intervals; for QT interval estimation, MOMENT has higher MAE than the best configuration from every other family, with non-overlapping 95\% intervals.
These patterns indicate that regression performance depends strongly on the target variable and dataset rather than on model family alone.

\begin{table*}[htbp]
\small\centering
\caption{Holistic win scores across all 11 model configurations on the full suite and the shared Stage~1 core. The full suite contains six classification and 45 regression evaluations; the core retains the same six classification evaluations and 17 regression evaluations. Higher classification scores and lower regression mean absolute errors are better. Each dataset--target evaluation contributes one win, split equally among tied best reported means. Family scores sum model-level credits; combined scores sum credits within the generalist or specialist group. Bold indicates the highest family, model, and group score within each panel.}
\label{tab:holistic_win_scores}
\resizebox{\textwidth}{!}{%
\begin{tabular}{lccccccccccc}
\toprule
\multirow{2}{*}{Score} & \multicolumn{3}{c}{MOMENT} & \multicolumn{2}{c}{Chronos-2} & \multicolumn{3}{c}{GPT-PPG} & \multicolumn{3}{c}{MOMENT-PPG} \\
\cmidrule(lr){2-4}\cmidrule(lr){5-6}\cmidrule(lr){7-9}\cmidrule(lr){10-12}
 & 40M & 125M & 385M & 28M & 120M & 19M & 85M & 345M & 40M & 125M & 385M \\
\midrule
\multicolumn{12}{l}{\textbf{Full suite --- classification: head-only tuning}} \\
Family score & \multicolumn{3}{c}{\textbf{4/6}} & \multicolumn{2}{c}{0/6} & \multicolumn{3}{c}{2/6} & \multicolumn{3}{c}{0/6} \\
Win score & 0/6 & 1/6 & \textbf{3/6} & 0/6 & 0/6 & 0/6 & 2/6 & 0/6 & 0/6 & 0/6 & 0/6 \\
Combined & \multicolumn{5}{c}{Generalists: \textbf{4/6}} & \multicolumn{6}{c}{Specialists: 2/6} \\
\midrule
\multicolumn{12}{l}{\textbf{Full suite --- classification: full-model tuning}} \\
Family score & \multicolumn{3}{c}{\textbf{3/6}} & \multicolumn{2}{c}{2/6} & \multicolumn{3}{c}{0/6} & \multicolumn{3}{c}{1/6} \\
Win score & 1/6 & \textbf{2/6} & 0/6 & 0/6 & \textbf{2/6} & 0/6 & 0/6 & 0/6 & 0/6 & 0/6 & 1/6 \\
Combined & \multicolumn{5}{c}{Generalists: \textbf{5/6}} & \multicolumn{6}{c}{Specialists: 1/6} \\
\midrule
\multicolumn{12}{l}{\textbf{Full suite --- regression: head-only tuning}} \\
Family score & \multicolumn{3}{c}{10/45} & \multicolumn{2}{c}{10.33/45} & \multicolumn{3}{c}{6.17/45} & \multicolumn{3}{c}{\textbf{18.5/45}} \\
Win score & 1.5/45 & 1/45 & 7.5/45 & 6.17/45 & 4.17/45 & 0.5/45 & 3/45 & 2.67/45 & 4.17/45 & \textbf{10.67/45} & 3.67/45 \\
Combined & \multicolumn{5}{c}{Generalists: 20.33/45} & \multicolumn{6}{c}{Specialists: \textbf{24.67/45}} \\
\midrule
\multicolumn{12}{l}{\textbf{Full suite --- regression: full-model tuning}} \\
Family score & \multicolumn{3}{c}{8/45} & \multicolumn{2}{c}{14.25/45} & \multicolumn{3}{c}{7.25/45} & \multicolumn{3}{c}{\textbf{15.5/45}} \\
Win score & 2.17/45 & 4.17/45 & 1.67/45 & 4.25/45 & \textbf{10/45} & 4/45 & 1.25/45 & 2/45 & 8.5/45 & 3/45 & 4/45 \\
Combined & \multicolumn{5}{c}{Generalists: 22.25/45} & \multicolumn{6}{c}{Specialists: \textbf{22.75/45}} \\
\midrule
\midrule
\multicolumn{12}{l}{\textbf{Shared core --- classification: head-only tuning}} \\
Family score & \multicolumn{3}{c}{\textbf{4/6}} & \multicolumn{2}{c}{0/6} & \multicolumn{3}{c}{2/6} & \multicolumn{3}{c}{0/6} \\
Win score & 0/6 & 1/6 & \textbf{3/6} & 0/6 & 0/6 & 0/6 & 2/6 & 0/6 & 0/6 & 0/6 & 0/6 \\
Combined & \multicolumn{5}{c}{Generalists: \textbf{4/6}} & \multicolumn{6}{c}{Specialists: 2/6} \\
\midrule
\multicolumn{12}{l}{\textbf{Shared core --- classification: full-model tuning}} \\
Family score & \multicolumn{3}{c}{\textbf{3/6}} & \multicolumn{2}{c}{2/6} & \multicolumn{3}{c}{0/6} & \multicolumn{3}{c}{1/6} \\
Win score & 1/6 & \textbf{2/6} & 0/6 & 0/6 & \textbf{2/6} & 0/6 & 0/6 & 0/6 & 0/6 & 0/6 & 1/6 \\
Combined & \multicolumn{5}{c}{Generalists: \textbf{5/6}} & \multicolumn{6}{c}{Specialists: 1/6} \\
\midrule
\multicolumn{12}{l}{\textbf{Shared core --- regression: head-only tuning}} \\
Family score & \multicolumn{3}{c}{3/17} & \multicolumn{2}{c}{4.33/17} & \multicolumn{3}{c}{4.17/17} & \multicolumn{3}{c}{\textbf{5.5/17}} \\
Win score & 0.5/17 & 0/17 & 2.5/17 & 2.17/17 & 2.17/17 & 0/17 & 2/17 & 2.17/17 & 0.17/17 & \textbf{4.17/17} & 1.17/17 \\
Combined & \multicolumn{5}{c}{Generalists: 7.33/17} & \multicolumn{6}{c}{Specialists: \textbf{9.67/17}} \\
\midrule
\multicolumn{12}{l}{\textbf{Shared core --- regression: full-model tuning}} \\
Family score & \multicolumn{3}{c}{3/17} & \multicolumn{2}{c}{\textbf{6.25/17}} & \multicolumn{3}{c}{4.25/17} & \multicolumn{3}{c}{3.5/17} \\
Win score & 0.33/17 & 2.33/17 & 0.33/17 & 1.25/17 & \textbf{5/17} & 2/17 & 0.25/17 & 2/17 & 1.5/17 & 0.5/17 & 1.5/17 \\
Combined & \multicolumn{5}{c}{Generalists: \textbf{9.25/17}} & \multicolumn{6}{c}{Specialists: 7.75/17} \\
\bottomrule
\end{tabular}%
}
\par\smallskip
\begin{minipage}{\textwidth}\footnotesize
Classification results are repeated because all six classification evaluations belong to the shared core. Scores are earned credit / available credit. Fractional credits are summed exactly before rounding to at most two decimal places. The full and core panels use the same 11 competitors. Bolding does not imply statistical significance.
\end{minipage}
\end{table*}

\begin{table*}[htbp]
\centering\small
\setlength{\tabcolsep}{5pt}
\renewcommand{\arraystretch}{1.08}
\caption{Paired head-only / full fine-tuning results. Scores are averaged equally across datasets and model scales within each family. Bold and underlining indicate the best and second-best unrounded means separately for each tuning setting. Balanced wins assign half of each task's credit to each tuning setting. Win scores are reported as earned credit / maximum possible credit.}
\label{tab:paired_head_full_fraction_wins}
\resizebox{0.99\textwidth}{!}{%
\begin{tabular}{@{}lrrrr@{}}
\toprule
 & \multicolumn{2}{c}{Generalists} & \multicolumn{2}{c}{Specialists} \\
\cmidrule(lr){2-3}\cmidrule(lr){4-5}
Actual task & MOMENT & Chronos-2 & GPT-PPG & MOMENT-PPG \\
 & \textit{Head / Full} & \textit{Head / Full} & \textit{Head / Full} & \textit{Head / Full} \\
\midrule
\multicolumn{5}{@{}l}{\textit{Classification: F1 or accuracy (\%), $\uparrow$}} \\
AF detection & 59.95 / 73.86 & 61.77 / \textbf{76.80} & \textbf{68.14} / \underline{76.15} & \underline{65.45} / 74.54 \\
Mental workload & \textbf{69.60} / \textbf{69.77} & \underline{66.43} / \underline{68.96} & 62.94 / 64.38 & 60.84 / 61.17 \\
Activity recognition & \underline{78.37} / \textbf{86.39} & \textbf{81.33} / 77.14 & 66.14 / \underline{78.09} & 70.61 / 72.09 \\
Human identification & \underline{84.76} / \textbf{92.53} & 32.05 / \underline{85.63} & 69.33 / 79.33 & \textbf{86.67} / 72.52 \\
\midrule
Head wins & 1/4 & 1/4 & 1/4 & 1/4 \\
Full wins & 3/4 & 1/4 & 0/4 & 0/4 \\
Balanced family wins & 2/4 & 1/4 & 0.5/4 & 0.5/4 \\
Balanced group wins & \multicolumn{2}{c}{Generalists: 3/4} & \multicolumn{2}{c}{Specialists: 1/4} \\
\midrule
\multicolumn{5}{@{}l}{\textit{Regression: mean absolute error, $\downarrow$}} \\
SpO$_2$ estimation & 3.62 / 2.90 & 3.28 / \underline{2.39} & \underline{3.27} / 2.51 & \textbf{2.73} / \textbf{2.29} \\
Skin temperature estimation & 1.50 / 1.73 & 1.48 / \underline{1.36} & \textbf{1.13} / \textbf{1.06} & \underline{1.38} / 1.38 \\
HR estimation & \underline{11.94} / 11.55 & 12.31 / \textbf{9.41} & 13.30 / 10.16 & \textbf{10.94} / \underline{9.47} \\
RR estimation & 3.39 / 3.36 & \underline{3.19} / \underline{2.91} & 3.30 / 2.93 & \textbf{3.12} / \textbf{2.76} \\
DBP estimation & 10.73 / 10.49 & \underline{8.93} / \underline{8.45} & 14.26 / \textbf{8.21} & \textbf{8.59} / 8.54 \\
SBP estimation & 20.30 / 16.48 & \textbf{14.78} / \underline{14.82} & 18.54 / 15.80 & \underline{14.88} / \textbf{14.81} \\
Potassium estimation & \underline{0.47} / \textbf{0.46} & \textbf{0.47} / \underline{0.48} & 0.51 / 0.49 & 0.48 / 0.51 \\
Sodium estimation & 7.27 / \textbf{3.77} & \textbf{3.65} / \underline{4.08} & 6.74 / 5.09 & \underline{3.85} / 4.56 \\
Glucose estimation & \underline{45.25} / \textbf{43.99} & \textbf{44.44} / \underline{44.88} & 45.80 / 46.25 & 46.68 / 47.42 \\
A1c estimation & \textbf{1.42} / \underline{1.35} & 1.50 / 1.65 & \underline{1.43} / \textbf{1.34} & 1.67 / 1.63 \\
Troponin estimation & \textbf{3.61} / \textbf{3.90} & 4.95 / 4.75 & \underline{3.82} / \underline{4.15} & 4.37 / 4.40 \\
Lactate estimation & \underline{1.22} / \underline{1.27} & 1.33 / 1.34 & \textbf{1.17} / \textbf{1.15} & 1.24 / 1.37 \\
PR interval estimation & 30.09 / 28.42 & \underline{26.60} / \underline{27.92} & 28.46 / \textbf{27.71} & \textbf{26.44} / 30.20 \\
QRS duration estimation & 18.60 / \underline{17.15} & \underline{16.61} / \textbf{16.68} & 17.43 / 17.17 & \textbf{16.39} / 19.61 \\
QT interval estimation & 53.74 / 50.62 & \underline{35.40} / \textbf{33.70} & 51.38 / \underline{35.15} & \textbf{33.53} / 37.58 \\
LVEF estimation & \underline{13.20} / 14.18 & 13.77 / \underline{13.48} & 14.26 / 14.43 & \textbf{12.77} / \textbf{12.85} \\
LV mass estimation & 32.09 / 31.61 & \textbf{30.48} / 31.98 & 31.83 / \textbf{31.47} & \underline{31.73} / \underline{31.56} \\
HDL estimation & \textbf{13.82} / \textbf{12.63} & 14.78 / \underline{13.37} & \underline{13.92} / 13.46 & 15.42 / 15.53 \\
LDL estimation & \underline{31.01} / \textbf{29.78} & \textbf{30.14} / 32.29 & 31.90 / 32.21 & 31.64 / \underline{32.02} \\
\midrule
Head wins & 3/19 & 6/19 & 2/19 & 8/19 \\
Full wins & 6/19 & 3/19 & 6/19 & 4/19 \\
Balanced family wins & 4.5/19 & 4.5/19 & 4/19 & 6/19 \\
Balanced group wins & \multicolumn{2}{c}{Generalists: 9/19} & \multicolumn{2}{c}{Specialists: 10/19} \\
Balanced group wins (through BP task) & \multicolumn{2}{c}{Generalists: 1/6} & \multicolumn{2}{c}{Specialists: 5/6} \\
\bottomrule
\end{tabular}%
}
\par\vspace{3pt}
\begin{minipage}{0.99\textwidth}\footnotesize
For each task, head and full tuning each contribute half a win, shared equally among any tied winning families. Balanced family wins equal $(W_{\mathrm{head}}+W_{\mathrm{full}})/2$; group wins sum their constituent family credits. Each task contributes one total balanced win. Denominators are 4 classification tasks or 19 regression tasks. Rankings and wins use unrounded means. Regression units are unspecified in the source. Bold and underlining do not imply statistical significance.
\end{minipage}
\end{table*}

\begin{table*}[htbp]
\centering\small
\setlength{\tabcolsep}{5pt}
\renewcommand{\arraystretch}{1.08}
\caption{Paired head-only / full fine-tuning results restricted to the shared Stage~1 core. The 23 core dataset--target evaluations are summarized into four classification and 12 regression task categories. Within each category, scores are averaged equally across the retained datasets and model scales within each family. Bold and underlining indicate the best and second-best distinct unrounded family means separately for each tuning setting. Balanced wins assign half of each category's credit to each tuning setting.}
\label{tab:paired_head_full_core_fraction_wins}
\resizebox{0.99\textwidth}{!}{%
\begin{tabular}{@{}lrrrr@{}}
\toprule
 & \multicolumn{2}{c}{Generalists} & \multicolumn{2}{c}{Specialists} \\
\cmidrule(lr){2-3}\cmidrule(lr){4-5}
Actual task & MOMENT & Chronos-2 & GPT-PPG & MOMENT-PPG \\
 & \textit{Head / Full} & \textit{Head / Full} & \textit{Head / Full} & \textit{Head / Full} \\
\midrule
\multicolumn{5}{@{}l}{\textit{Classification: 4 task categories; F1 or accuracy (\%), $\uparrow$}} \\
AF detection & 59.95 / 73.86 & 61.77 / \textbf{76.80} & \textbf{68.14} / \underline{76.15} & \underline{65.45} / 74.54 \\
Mental workload / stress & \textbf{69.60} / \textbf{69.77} & \underline{66.43} / \underline{68.96} & 62.94 / 64.38 & 60.84 / 61.17 \\
Activity recognition & \underline{78.37} / \textbf{86.39} & \textbf{81.33} / 77.14 & 66.14 / \underline{78.09} & 70.61 / 72.09 \\
Human identification & \underline{84.76} / \textbf{92.53} & 32.05 / \underline{85.63} & 69.33 / 79.33 & \textbf{86.67} / 72.52 \\
\midrule
Head wins & 1/4 & 1/4 & 1/4 & 1/4 \\
Full wins & 3/4 & 1/4 & 0/4 & 0/4 \\
Balanced family wins & 2/4 & 1/4 & 0.5/4 & 0.5/4 \\
Balanced group wins & \multicolumn{2}{c}{Generalists: 3/4} & \multicolumn{2}{c}{Specialists: 1/4} \\
\midrule
\multicolumn{5}{@{}l}{\textit{Regression: 12 task categories; mean absolute error, $\downarrow$}} \\
SpO$_2$ estimation & 2.57 / 2.25 & \underline{2.21} / \underline{2.21} & 2.53 / 2.43 & \textbf{2.19} / \textbf{2.20} \\
Skin temperature estimation & \underline{1.36} / 2.17 & 1.77 / 1.88 & \textbf{1.09} / \textbf{1.18} & 1.76 / \underline{1.83} \\
HR estimation & \underline{11.10} / 11.35 & 13.85 / \underline{9.43} & 13.61 / 9.56 & \textbf{10.21} / \textbf{8.32} \\
RR estimation & 4.63 / 4.63 & \textbf{4.44} / \textbf{4.31} & 4.63 / 4.51 & \underline{4.49} / \underline{4.33} \\
DBP estimation & 14.07 / 14.21 & \underline{10.88} / \textbf{8.87} & 20.97 / 9.08 & \textbf{9.47} / \underline{8.97} \\
SBP estimation & 19.21 / 18.66 & 17.47 / \textbf{16.47} & \underline{17.36} / \underline{16.74} & \textbf{16.72} / 16.74 \\
Potassium estimation & 0.49 / \underline{0.47} & \textbf{0.46} / \underline{0.47} & \underline{0.47} / 0.47 & \textbf{0.46} / \textbf{0.46} \\
Sodium estimation & 5.58 / \textbf{3.63} & \textbf{3.69} / \underline{3.67} & 4.11 / 4.05 & \underline{3.79} / 3.75 \\
Glucose estimation & 37.54 / 36.01 & 36.66 / \underline{35.32} & \textbf{35.09} / 35.53 & \underline{35.64} / \textbf{35.26} \\
A1c estimation & \textbf{1.42} / \underline{1.35} & 1.50 / 1.65 & \underline{1.43} / \textbf{1.34} & 1.67 / 1.63 \\
Troponin estimation & \textbf{1.18} / \underline{1.27} & 1.29 / \textbf{1.27} & \underline{1.19} / 1.31 & 1.41 / 1.33 \\
Lactate estimation & \textbf{1.18} / \underline{1.21} & 1.28 / 1.24 & 1.20 / \textbf{1.20} & \underline{1.20} / 1.24 \\
\midrule
Head wins & 3/12 & 2.5/12 & 2/12 & 4.5/12 \\
Full wins & 1/12 & 4/12 & 3/12 & 4/12 \\
Balanced family wins & 2/12 & 3.25/12 & 2.5/12 & 4.25/12 \\
Balanced group wins & \multicolumn{2}{c}{Generalists: 5.25/12} & \multicolumn{2}{c}{Specialists: 6.75/12} \\
Balanced group wins (through BP) & \multicolumn{2}{c}{Generalists: 2/6} & \multicolumn{2}{c}{Specialists: 4/6} \\
\bottomrule
\end{tabular}%
}
\par\vspace{3pt}
\begin{minipage}{0.99\textwidth}\footnotesize
Only the Stage~1 core dataset--target pairs are included. Mental workload and mental stress are combined to preserve the category grouping of the full-suite paired table. Each category contributes one balanced win: head and full tuning each contribute half a win, shared equally among tied winning families. Balanced family wins equal $(W_{\mathrm{head}}+W_{\mathrm{full}})/2$; group wins sum constituent family credits. Denominators are four classification or 12 regression categories; the final row covers the first six regression categories. These category-level wins differ from the individual-model dataset--target wins in Table~\ref{tab:holistic_win_scores}. Rankings use unrounded family means; displayed values have two decimal places. Regression values retain the source units. Bold and underlining do not imply statistical significance.
\end{minipage}
\end{table*}

\subsection{Additional result summary}

Across the 51 dataset–task evaluations, generalists perform better in classification, whereas specialists hold a narrower advantage in regression (Table~\ref{tab:holistic_win_scores}). MOMENT earns the highest classification win score among the four families under both tuning settings, while MOMENT-PPG leads regression. Chronos-2 is the strongest generalist in the evaluation-level regression comparison, particularly under full tuning, where its 14.25/45 win credits approach MOMENT-PPG’s 15.5/45 and exceed MOMENT’s 8/45. Its strengths include heart-rate, QRS-duration, and QT-interval estimation, for which it achieves the best family-average performance under full tuning. The comparison that weights each task equally and uses both tuning settings shows the same overall pattern: generalists earn 3/4 classification credits versus 1/4 for specialists, while specialists earn 10/19 regression credits versus 9/19 for generalists (Table~\ref{tab:paired_head_full_fraction_wins}). Thus, Chronos-2 contributes substantially to generalist regression performance, alongside MOMENT’s stronger classification results.
Table~\ref{tab:holistic_win_scores} and Table~\ref{tab:paired_head_full_core_fraction_wins} additionally include results for the core (23-task) suite (stage 1 of our benchmarking) for fair comparison. The slight differences in the core vs expanded suite (esp. in regression tasks) highlight the need for bigger suites for more general conclusions.

Specialists show their clearest advantage in physiological monitoring, earning 5/6 available task-win credits across oxygen saturation, skin temperature, heart rate, respiration rate, and systolic and diastolic blood-pressure estimation. MOMENT-PPG achieves the best family-average performance for oxygen saturation, respiration rate, and left ventricular ejection fraction under both tuning settings, while GPT-PPG consistently leads skin-temperature and lactate estimation. Specialist families also lead diastolic blood-pressure and PR-interval estimation in both settings, although the winning family changes. Generalists are stronger across biochemical targets, earning 6.5/8 credits and consistently leading potassium, sodium, glucose, troponin, HDL, and LDL estimation; they also lead mental-workload and activity classification. These task-level patterns nevertheless conceal variation across datasets: blood-pressure rankings depend strongly on the dataset and differ between systolic and diastolic targets. Some aggregate advantages are also numerically small, particularly for systolic blood pressure under full tuning. Overall, specialization provides its clearest benefits in physiological monitoring, while generalists remain stronger for several biochemical targets and classification tasks.

\begin{table*}[htbp]
\small
\centering
\caption{Performance as a function of specialist pretraining-source overlap and downstream supervised-data scale. Downstream data scale is defined using training-split PPG signal hours only. Specialist advantage is the median [IQR] percentage reduction in the mean MAE of the two specialist FMs relative to the mean MAE of the two generalist FMs on each task; positive values favor the specialist FMs. The small tier uses MOMENT-40M, Chronos-2-28M, GPT-PPG-19M, and MOMENT-PPG-40M, while the medium tier uses MOMENT-125M, Chronos-2-120M, GPT-PPG-85M, and MOMENT-PPG-125M. All models are fully tuned. Inst. A is the pretraining-source-overlap group for GPT-PPG and MOMENT-PPG.}
\label{tab:domain_data_regime}
\resizebox{0.99\textwidth}{!}{%
\begin{tabular}{llcccc}
\toprule
\multirow{2}{*}{\shortstack{Specialist pretraining-\\source overlap}} & \multirow{2}{*}{\shortstack{Downstream\\scale}} & \multirow{2}{*}{$N$} & \multirow{2}{*}{\shortstack{Train\\hours}} & \multicolumn{2}{c}{Specialist advantage} \\
\cmidrule(lr){5-6}
& & & & Small tier & Medium tier \\
\midrule
Inst.A & All & 13 & 105.0 [99.0--381.0] & +1.2 [$-$4.5--4.1] & $-$2.1 [$-$9.3--5.8] \\
Non-Inst.A & $\leq 10$ h & 14 & 3.8 [0.2--9.8] & +3.2 [$-$3.9--25.4] & +0.1 [$-$6.8--2.7] \\
Non-Inst.A & $10$--$<100$ h & 5 & 23.0 [12.6--94.2] & +16.5 [1.3--17.6] & +1.2 [1.2--32.6] \\
Non-Inst.A & $\geq 100$ h & 13 & 100.0 [100.0--322.3] & +1.1 [$-$2.0--8.0] & +1.0 [$-$1.2--2.5] \\
\bottomrule
\end{tabular}}
\end{table*}

Pretraining-source overlap and downstream data scale: In Table~\ref{tab:domain_data_regime}, to examine whether the performance of PPG-specialist FMs is explained by overlap with their Inst.A pretraining source, we stratified regression tasks into Inst. A endpoints and non-Inst.A endpoints, further dividing the latter by the amount of labeled PPG available for downstream training. 
We used training-split signal hours rather than total dataset duration and fixed all models to their small, fully tuned configurations to avoid confounding the analysis with model scale. Across UCSF tasks, the specialist pair showed only a small median advantage over the two generalist FMs, comparable to that observed on large non-UCSF tasks. Outside UCSF, specialist advantage did not vary monotonically with downstream data size. A matched-target comparison of five laboratory endpoints available in both UCSF and MIMIC-III similarly showed no consistent increase in specialist advantage on the UCSF data. Together, these results suggest that the benefits of PPG-specialist pretraining are not explained solely by exact pretraining-source overlap and can transfer beyond the source cohort, although the magnitude of this benefit remains task- and model-scale-dependent.

\section{Architectural ablation details}
\label{app:arch_ablation}

\paragraph{Setup.}
All variants share the GPT-PPG 85M Transformer skeleton; only the token mixer in every block is replaced, with widths adjusted so that parameter counts stay within about 2\% of the Transformer's. The seven alternatives are local depthwise convolution, global depthwise convolution, Hyena, a selective state-space mixer, a linear recurrent unit (LRU), a GRU and an LSTM; an eighth variant keeps attention but replaces rotary position embeddings with ALiBi biases. On the UCSF corpus (104.8M 30-s segments at 40\,Hz, 30 patches of 40 samples), every variant was trained with MOMENT's masked-reconstruction objective for 50{,}000 steps at a batch of 2{,}048, with three seeds and the same data order. Every variant's peak learning rate, the Transformer's included, was chosen by the same rule: two 3{,}000-step pilots at $10^{-4}$ and $4\times10^{-4}$, keeping, among pilots that did not collapse, the one with the lower held-out loss. On the second corpus (9.6M segments from PulseDB, i.e.\ MIMIC-III and VitalDB, and MC-MED), the Transformer and the convolutional, state-space and LRU variants were trained for 10{,}000 steps at the same batch size under three objectives: masked reconstruction, a contrastive objective, and next-patch prediction (GPT-PPG's objective). Each objective kept its own warm-up and decay schedule and the peak learning rate that pilots had selected for it on the UCSF corpus ($4\times10^{-4}$ for masked reconstruction and next-patch prediction, $2.5\times10^{-4}$ for contrastive learning). For next-patch prediction the mixers are made causal: the convolution is left-padded, and the state-space mixer and the LRU run forward only. Every encoder is evaluated with the frozen linear probe of the 51-task suite; the 16 informative tasks and the composite score are defined in Section~\ref{sec:arch_ablation}. On the UCSF corpus, the composite is standardized over the 27 runs of the nine variants; on the second corpus, over the 12 runs of each objective.

\paragraph{Every run on the second corpus.}
Table~\ref{tab:arch_composite} lists the composite of every run. Under masked reconstruction and next-patch prediction, every run with an alternative mixer scores better than every Transformer run; under contrastive learning the ranges overlap. The held-out losses show that this is not a difference in fit: under masked reconstruction, the Transformer reaches a held-out reconstruction error of 0.0091--0.0093, within the range of the alternatives (state-space 0.0087--0.0088, convolution 0.0093, LRU 0.0095--0.0097; three seeds each), and under next-patch prediction the held-out loss is $-1.688$ for the Transformer against $-1.697$, $-1.737$ and $-1.700$ for the convolutional, state-space and LRU variants (seed means). Per-task gains of the alternatives over the Transformer under masked reconstruction agree between the two corpora (Spearman $\rho = 0.75$, $0.47$ and $0.61$ for the convolutional, state-space and LRU variants). Compared across objectives, contrastive training improves the Transformer over masked reconstruction on 10 of the 16 tasks (median $+1.8\%$), but not the alternatives (7 to 9 of the 16 tasks, median $-1.0\%$ to $+0.4\%$). The objectives differ in learning rate and schedule, so this comparison is less controlled than the comparisons between mixers.

\begin{table}[h]
\centering
\caption{Composite score of every run on the second corpus (z-scored log error over the 16 informative tasks and over the runs of each objective; lower is better). Each cell lists the three seeds.}
\label{tab:arch_composite}
\footnotesize
\setlength{\tabcolsep}{3pt}
\begin{tabular}{lcccc}
\toprule
\textbf{Objective} & \textbf{Transformer} & \textbf{Convolution} & \textbf{State-space} & \textbf{LRU} \\
\midrule
Masked & $+0.75, +0.70, +0.93$ & $-0.15, -0.00, +0.03$ & $-0.47, -0.37, -0.69$ & $-0.52, -0.13, -0.09$ \\
Next-patch & $+0.74, +0.85, +0.85$ & $-0.05, -0.09, +0.13$ & $-0.66, -0.30, -0.32$ & $-0.29, -0.36, -0.48$ \\
Contrastive & $+0.33, +0.07, -0.16$ & $+0.05, +0.18, +0.25$ & $-0.59, -0.26, -0.14$ & $+0.05, +0.11, +0.11$ \\
\bottomrule
\end{tabular}
\end{table}

\paragraph{Synthetic factor test.}
To test whether the deficit is built into the architecture, we used synthetic PPG in which beat timing and pulse morphology vary independently: two timing factors (the coefficient of variation of the inter-beat intervals, and their lag-1 autocorrelation, which changes the order of the intervals but not their distribution) and two morphology factors (a reflected-wave amplitude and a shape exponent that moves no fiducial time). Each factor was regressed from the frozen embeddings of every UCSF-corpus encoder with a linear probe, and the four z-scored log errors were averaged (lower is better). At random initialization, the attention mixers are not separated from the alternatives (Transformer $+0.32$, ALiBi $-0.04$; alternatives $-0.65$ to $+0.88$). After pre-training, both encode the factors worse than all seven alternatives (Transformer $+0.54$, ALiBi $+0.23$; alternatives $-0.90$ to $+0.02$), so the deficit arises during pre-training.

\section{Frozen-encoder MOMENT nearest-neighbor baseline}
\label{app:moment_frozen_knn}

This appendix reports frozen-encoder nearest-neighbor baselines for MOMENT, Chronos-2, GPT-PPG, and MOMENT-PPG, as described in Section~\ref{subsec:moment_frozen_knn}.
For classification tasks, Table~\ref{tab:moment_frozen_knn_cls} reports F1 scores.
Binary tasks use binary averaging.
For the 35-class single-label human-identification task, F1 is micro-averaged and is therefore equal to accuracy.
For regression tasks, Table~\ref{tab:frozen_knn_reg} reports mean absolute error (MAE).
Values are reported as mean $\pm$ bootstrap standard error, with 95\% confidence intervals computed from 1000 test-set bootstrap resamples.

\begin{table}[htbp]
\small
\centering
\resizebox{\textwidth}{!}{%
\begin{tabular}{lcccc}
\toprule
\textbf{Dataset} & \textbf{MOMENT} & \textbf{Chronos-2} & \textbf{GPT-PPG} & \textbf{MOMENT-PPG} \\
\midrule
Stanford & 68.20 $\pm$ 1.28 [65.61, 70.61] & 68.57 $\pm$ 1.27 [66.09, 70.96] & \textbf{81.25 $\pm$ 1.07} [79.13, 83.31] & 64.69 $\pm$ 1.27 [62.15, 67.21] \\
Simband & 54.17 $\pm$ 4.15 [46.07, 62.69] & \textbf{65.81 $\pm$ 3.95} [57.71, 73.51] & 50.00 $\pm$ 4.42 [40.94, 58.59] & 47.62 $\pm$ 4.47 [38.67, 55.88] \\
MAUS & 55.39 $\pm$ 2.80 [49.76, 60.91] & 57.92 $\pm$ 2.80 [52.54, 63.31] & \textbf{64.86 $\pm$ 2.48} [59.78, 69.62] & 61.34 $\pm$ 2.64 [55.84, 66.37] \\
Mental Stress& 44.61 $\pm$ 8.23 [28.07, 59.70] & 59.20 $\pm$ 7.41 [44.44, 73.02] & 61.76 $\pm$ 7.16 [46.87, 74.67] & \textbf{73.53 $\pm$ 6.14} [59.65, 83.96] \\
Pulse Transit & \textbf{82.69 $\pm$ 0.97} [80.75, 84.66] & 78.13 $\pm$ 1.10 [76.05, 80.34] & 73.10 $\pm$ 1.34 [70.57, 75.72] & 63.50 $\pm$ 1.39 [60.74, 66.16] \\
Real-world & 73.07 $\pm$ 1.68 [69.85, 76.29] & 25.00 $\pm$ 1.67 [21.71, 28.29] & \textbf{76.57 $\pm$ 1.65} [73.29, 79.72] & 65.71 $\pm$ 1.79 [62.29, 69.29] \\
\bottomrule
\end{tabular}
}
\caption{Frozen-encoder nearest-neighbor baselines on classification tasks, reported as F1 (\%). Each encoder is kept frozen; segment embeddings are mean-pooled over patches and classified with a $k$-nearest-neighbor classifier ($k=5$, cosine distance) fit on the training split. Values are mean $\pm$ standard error with 95\% bootstrap confidence intervals. For the 35-class Real-world PPG (human identification) task, F1 is micro-averaged.}
\label{tab:moment_frozen_knn_cls}
\end{table}

\begin{table}[htbp]
\small
\centering
\resizebox{\textwidth}{!}{%
\begin{tabular}{llcccc}
\toprule
\textbf{Task} & \textbf{Dataset} & \textbf{MOMENT} & \textbf{Chronos-2} & \textbf{GPT-PPG} & \textbf{MOMENT-PPG} \\
\midrule
SpO$_2$ Estimation & MIMIC-III
& 2.41 $\pm$ 0.03 [2.36, 2.46]
& 2.44 $\pm$ 0.02 [2.39, 2.48]
& 2.44 $\pm$ 0.03 [2.40, 2.50]
& \textbf{2.41} $\pm$ 0.03 [2.36, 2.46] \\

Heart Rate Estimation & WESAD
& \textbf{10.95} $\pm$ 0.08 [10.80, 11.11]
& 11.01 $\pm$ 0.08 [10.85, 11.16]
& 11.43 $\pm$ 0.08 [11.27, 11.60]
& 11.21 $\pm$ 0.08 [11.05, 11.37] \\

Heart Rate Estimation & DALIA
& \textbf{12.27} $\pm$ 0.11 [12.07, 12.50]
& 14.66 $\pm$ 0.12 [14.44, 14.90]
& 18.44 $\pm$ 0.14 [18.18, 18.70]
& 12.67 $\pm$ 0.11 [12.47, 12.88] \\

Respiration Rate Estimation & MIMIC-III
& 5.03 $\pm$ 0.04 [4.95, 5.11]
& 5.02 $\pm$ 0.04 [4.94, 5.10]
& 5.05 $\pm$ 0.04 [4.97, 5.14]
& \textbf{5.00} $\pm$ 0.04 [4.92, 5.08] \\

Skin Temperature Estimation & MIMIC-III
& 1.89 $\pm$ 0.02 [1.86, 1.92]
& 1.87 $\pm$ 0.01 [1.84, 1.89]
& \textbf{1.72} $\pm$ 0.02 [1.69, 1.75]
& 1.91 $\pm$ 0.02 [1.87, 1.94] \\

Diastolic BP Estimation & PPG-BP
& 10.00 $\pm$ 0.66 [8.78, 11.38]
& 9.25 $\pm$ 0.68 [7.98, 10.66]
& \textbf{9.18} $\pm$ 0.69 [7.89, 10.60]
& 9.68 $\pm$ 0.72 [8.32, 11.09] \\

Systolic BP Estimation & PPG-BP
& 18.45 $\pm$ 1.18 [16.23, 20.80]
& \textbf{17.54} $\pm$ 1.22 [15.24, 20.00]
& 17.65 $\pm$ 1.26 [15.22, 20.15]
& 18.18 $\pm$ 1.28 [15.79, 20.87] \\

Diastolic BP Estimation & Aurora-Oscillometric
& 13.65 $\pm$ 0.29 [13.10, 14.24]
& 10.71 $\pm$ 0.26 [10.22, 11.22]
& 8.51 $\pm$ 0.24 [8.08, 9.00]
& \textbf{7.97} $\pm$ 0.23 [7.56, 8.45] \\

Systolic BP Estimation & Aurora-Oscillometric
& 23.80 $\pm$ 0.36 [23.11, 24.49]
& 22.77 $\pm$ 0.33 [22.11, 23.45]
& 23.30 $\pm$ 0.37 [22.57, 24.01]
& \textbf{21.97} $\pm$ 0.34 [21.30, 22.69] \\

Diastolic BP Estimation & Aurora-Auscultatory
& 10.55 $\pm$ 0.38 [9.81, 11.26]
& 10.40 $\pm$ 0.41 [9.63, 11.20]
& 10.62 $\pm$ 0.43 [9.76, 11.46]
& \textbf{10.28} $\pm$ 0.39 [9.55, 11.06] \\

Systolic BP Estimation & Aurora-Auscultatory
& \textbf{14.34} $\pm$ 0.53 [13.37, 15.41]
& 15.02 $\pm$ 0.54 [13.97, 16.04]
& 15.56 $\pm$ 0.59 [14.46, 16.77]
& 15.22 $\pm$ 0.51 [14.19, 16.27] \\

Glucose Estimation & MIMIC-III
& 45.44 $\pm$ 0.48 [44.53, 46.39]
& 43.93 $\pm$ 0.48 [42.96, 44.84]
& \textbf{43.65} $\pm$ 0.48 [42.68, 44.61]
& 44.51 $\pm$ 0.48 [43.57, 45.48] \\

Lactate Estimation & MIMIC-III
& 1.48 $\pm$ 0.02 [1.44, 1.50]
& 1.44 $\pm$ 0.02 [1.41, 1.47]
& 1.47 $\pm$ 0.02 [1.44, 1.49]
& \textbf{1.42} $\pm$ 0.02 [1.39, 1.45] \\

Potassium Estimation & MIMIC-III
& \textbf{0.53} $\pm$ 0.00 [0.52, 0.54]
& 0.53 $\pm$ 0.00 [0.52, 0.54]
& 0.53 $\pm$ 0.00 [0.52, 0.54]
& 0.53 $\pm$ 0.00 [0.52, 0.54] \\

Sodium Estimation & MIMIC-III
& 4.27 $\pm$ 0.04 [4.19, 4.33]
& \textbf{4.23} $\pm$ 0.04 [4.16, 4.31]
& 4.32 $\pm$ 0.04 [4.25, 4.39]
& 4.30 $\pm$ 0.04 [4.23, 4.38] \\

Troponin Estimation & MIMIC-III
& 1.50 $\pm$ 0.02 [1.46, 1.55]
& 1.46 $\pm$ 0.02 [1.42, 1.50]
& \textbf{1.40} $\pm$ 0.02 [1.35, 1.44]
& 1.50 $\pm$ 0.02 [1.46, 1.55] \\

A1c Estimation & Inst. A
& 1.55 $\pm$ 0.03 [1.50, 1.60]
& 1.51 $\pm$ 0.02 [1.47, 1.56]
& \textbf{1.43} $\pm$ 0.02 [1.38, 1.47]
& 1.49 $\pm$ 0.02 [1.44, 1.54] \\
\bottomrule
\end{tabular}
}
\caption{Frozen-encoder nearest-neighbor regression results across four foundation models. Embeddings were mean-pooled, L2-normalized, and evaluated using uniform-mean 5-nearest-neighbor regression with cosine distance fitted only on each task's training split. Values are mean absolute error estimates $\pm$ bootstrap standard error with percentile 95\% confidence intervals from 1,000 paired sample-level bootstrap resamples. The lowest unrounded mean absolute error in each row is shown in bold.}
\label{tab:frozen_knn_reg}
\end{table}

\clearpage

\section{Additional Baseline, Robustness, and Mechanistic Analyses}
\label{app:further_analysis_details}

\subsection{Statistical, clinical-history, and traditional baselines}
\label{app:baseline_details}

Here, we investigate whether FMs provide additional value over simpler statistical, clinical history-based, or traditional ML baselines.

In Table~\ref{tab:stat_bline}, we choose a representative subset of our 51-task benchmark and compare the performance of the two generalist (MOMENT, Chronos-2) and two specialist (GPT-PPG, MOMENT-PPG) FMs with simpler statistical and clinical-history baselines based on demographics~\cite{chowdhury2020estimating} or the Last-Observed-Carry-Forward (LOCF) value~\cite{im2025labtop}, respectively.
For BP and HR estimation, we fit a linear regression model based on demographic information (age, height, and weight).
In BP prediction, the simple (demographic) baseline is significantly better, especially in the systolic case.
In HR prediction, FMs can surpass the demographic baseline by a slight margin -- revealing that non-invasive HR task is more approachable by FMs compared to non-invasive BP estimation.
For the remaining eight laboratory-test tasks, we use LOCF, where the most recent previously measured value of the same laboratory variable in the patient's record is used as the prediction for the current time.
This baseline is not a waveform forecasting baseline and is not applied to tasks without longitudinal response measurements.
LOCF is simple, but it is a strong clinical-history baseline when laboratory values are temporally persistent.
We note that for glucose estimation, FMs can match or surpass LOCF, but for all other tasks, LOCF remains stronger -- revealing the challenges in spot prediction.

Next, in Table~\ref{tab:carryfwd}, we take a deeper look by evaluating on subsets of patients with positive or negative state as marked by practitioners using clinical thresholds.
We also note performance for transition states, i.e., positive-to-negative (indicating recovery) and negative-to-positive (exhibiting degradation).
We find that during patient health transitions, LOCF can become inaccurate, while FMs sometimes produce lower errors because they rely on the current waveform rather than the history.
We enclose a win score matrix in Table~\ref{tab:carryfwd} indicating cumulative win for each FM in all subsets of patient status (\texttt{all}, \texttt{pos}, \texttt{neg}, \texttt{pos2neg}, \texttt{neg2pos}).
The discrepancy between LOCF and FM predictions may be used as an indicator for earlier testing of patient but for translation of this practice in clinical environment will require further validation.

\begin{table}[htbp]
\centering
\resizebox{\textwidth}{!}{%
\begin{tabular}{l|c|c|c|c|c}
\hline
\shortstack{Task $/$ \\ (bline type)} & MOMENT & Chronos-2 & GPT-PPG & MOMENT-PPG & \shortstack{Statistical \\ baseline} \\ 
\hline
\multicolumn{6}{l}{\textbf{Blood pressure estimation (Diastolic/Systolic)} (MAE) $\downarrow$ $/$ (Age,Height,Weight)} \\
\hline
BUT-PPG & \shortstack{8.46 $\pm$ 0.13 / \\ 16.61 $\pm$ 0.35} & \shortstack{7.51 $\pm$ 0.14 / \\ 14.01 $\pm$ 0.28} & \shortstack{7.71 $\pm$ 0.16 / \\ 15.43 $\pm$ 0.34} & \shortstack{7.70 $\pm$ 0.14 / \\ 14.50 $\pm$ 0.30} & \shortstack{\textbf{6.24 $\pm$ 0.13} / \\ \textbf{7.29 $\pm$ 0.19}} \\
\hline
\multicolumn{6}{l}{\textbf{Heart rate estimation} (MAE) $\downarrow$ $/$ (Age,Height,Weight)} \\
\hline
BUT-PPG & 10.36 $\pm$ 0.30 & \textbf{10.05 $\pm$ 0.29} & 10.22 $\pm$ 0.28 & 10.27 $\pm$ 0.28 & 10.67 $\pm$ 0.31 \\
\hline
\hline
\multicolumn{6}{l}{\textbf{Troponin estimation} (MAE) $\downarrow$ $/$ (Carry forward)} \\
\hline
MIMIC & 1.11 $\pm$ 0.02 & 1.26 $\pm$ 0.02 & 1.17 $\pm$ 0.02 & 1.23 $\pm$ 0.02 & \textbf{0.42 $\pm$ 0.04} \\
\hline
Inst. A & 5.82 $\pm$ 0.21 & 8.23 $\pm$ 0.26 & 6.29 $\pm$ 0.26 & 6.68 $\pm$ 0.22 & \textbf{1.05 $\pm$ 0.08} \\
\hline
\multicolumn{6}{l}{\textbf{Lactate estimation} (MAE) $\downarrow$ $/$ (Carry forward)} \\
\hline
MIMIC & 1.15 $\pm$ 0.01 & 1.23 $\pm$ 0.02 & 1.18 $\pm$ 0.02 & 1.19 $\pm$ 0.01 & \textbf{0.82 $\pm$ 0.02} \\
\hline
Inst. A & 1.23 $\pm$ 0.01 & 1.36 $\pm$ 0.02 & 1.09 $\pm$ 0.02 & 1.20 $\pm$ 0.02 & \textbf{0.51 $\pm$ 0.02} \\
\hline
\multicolumn{6}{l}{\textbf{Potassium estimation} (MAE) $\downarrow$ $/$ (Carry forward)} \\
\hline
MIMIC & 0.47 $\pm$ 0.00 & 0.46 $\pm$ 0.00 & 0.46 $\pm$ 0.00 & 0.46 $\pm$ 0.00 & \textbf{0.44 $\pm$ 0.01} \\
\hline
Inst. A & 0.45 $\pm$ 0.01 & 0.47 $\pm$ 0.01 & 0.48 $\pm$ 0.01 & 0.47 $\pm$ 0.01 & \textbf{0.36 $\pm$ 0.00} \\
\hline
\multicolumn{6}{l}{\textbf{Sodium estimation} (MAE) $\downarrow$ $/$ (Carry forward)} \\
\hline
MIMIC & 3.63 $\pm$ 0.03 & 3.66 $\pm$ 0.03 & 3.74 $\pm$ 0.03 & 3.74 $\pm$ 0.04 & \textbf{2.11 $\pm$ 0.02} \\
\hline
Inst. A & 3.84 $\pm$ 0.06 & 3.59 $\pm$ 0.06 & 4.72 $\pm$ 0.08 & 3.59 $\pm$ 0.06 & \textbf{1.66 $\pm$ 0.01} \\
\hline
\multicolumn{6}{l}{\textbf{Glucose estimation} (MAE) $\downarrow$ $/$ (Carry forward)} \\
\hline
MIMIC & 35.66 $\pm$ 0.40 & \textbf{35.00 $\pm$ 0.43} & 35.03 $\pm$ 0.45 & 35.24 $\pm$ 0.45 & 38.05 $\pm$ 0.75 \\
\hline
Inst. A & 51.62 $\pm$ 0.66 & 51.17 $\pm$ 0.66 & 54.13 $\pm$ 0.76 & 54.92 $\pm$ 0.83 & \textbf{27.76 $\pm$ 0.58} \\
\hline
\multicolumn{6}{l}{\textbf{A1c estimation} (MAE) $\downarrow$ $/$ (Carry forward)} \\
\hline
Inst. A & 1.34 $\pm$ 0.03 & 1.48 $\pm$ 0.03 & 1.30 $\pm$ 0.03 & 1.61 $\pm$ 0.03 & \textbf{0.06 $\pm$ 0.01} \\
\hline
\multicolumn{6}{l}{\textbf{ECHO LVEF} (MAE) $\downarrow$ $/$ (Carry forward)} \\
\hline
Inst. A & 13.06 $\pm$ 0.21 & 13.42 $\pm$ 0.19 & 14.11 $\pm$ 0.21 & 11.44 $\pm$ 0.19 & \textbf{0.78 $\pm$ 0.19} \\
\hline
\multicolumn{6}{l}{\textbf{ECHO LVmass} (MAE) $\downarrow$ $/$ (Carry forward)} \\
\hline
Inst. A & 31.61 $\pm$ 0.40 & 30.43 $\pm$ 0.39 & 31.33 $\pm$ 0.47 & 30.32 $\pm$ 0.51 & \textbf{0.48 $\pm$ 0.23} \\
\hline
\end{tabular}
}
\caption{Comparison of the best results obtained with MOMENT, Chronos-2, GPT-PPG, and MOMENT-PPG with statistical baselines. For each FM family, the best result across the evaluated model scales and tuning strategies is reported.}
\label{tab:stat_bline}
\end{table}

\begin{table}[htbp]
\centering
\small
\resizebox{\textwidth}{!}{%
\begin{tabular}{l|c|c|c|c|c}
\hline
Task / Subset & MOMENT & Chronos-2 & GPT-PPG & MOMENT-PPG & Carry forward \\
\hline
\multicolumn{6}{l}{\textbf{Troponin Estimation} (MAE) $\downarrow$} \\
\hline
all & $6.94\pm0.23$ & $8.82\pm0.27$ & $6.94\pm0.26$ & $7.76\pm0.24$ & \textbf{\boldmath$2.47\pm0.38$} \\
pos & $7.16\pm0.04$ & $7.49\pm0.04$ & $6.66\pm0.04$ & $7.09\pm0.04$ & \textbf{\boldmath$2.79\pm0.41$} \\
neg & $3.15\pm0.02$ & $3.35\pm0.04$ & $1.51\pm0.00$ & $3.22\pm0.02$ & \textbf{\boldmath$0.32\pm0.21$} \\
pos2neg & $3.53\pm0.06$ & $2.85\pm0.04$ & \textbf{\boldmath$1.43\pm0.01$} & $3.24\pm0.03$ & $1.54\pm1.08$ \\
neg2pos & $3.49\pm0.05$ & $5.59\pm0.22$ & \textbf{\boldmath$1.61\pm0.02$} & $3.27\pm0.03$ & $2.00\pm1.53$ \\
\hline
\multicolumn{6}{l}{\textbf{Lactate Estimation} (MAE) $\downarrow$} \\
\hline
all & $1.70\pm0.02$ & $1.23\pm0.02$ & $1.10\pm0.02$ & $1.30\pm0.02$ & \textbf{\boldmath$0.84\pm0.05$} \\
pos & $2.01\pm0.01$ & $2.69\pm0.01$ & $2.65\pm0.01$ & $2.25\pm0.01$ & \textbf{\boldmath$1.52\pm0.09$} \\
neg & $1.33\pm0.00$ & $0.72\pm0.00$ & $0.61\pm0.00$ & $0.91\pm0.00$ & \textbf{\boldmath$0.52\pm0.03$} \\
pos2neg & $1.10\pm0.01$ & $0.70\pm0.01$ & \textbf{\boldmath$0.53\pm0.01$} & $0.64\pm0.00$ & $1.94\pm0.15$ \\
neg2pos & \textbf{\boldmath$1.39\pm0.01$} & $2.04\pm0.01$ & $1.96\pm0.01$ & $1.55\pm0.01$ & $2.12\pm0.17$ \\
\hline
\multicolumn{6}{l}{\textbf{Potassium Estimation} (MAE) $\downarrow$} \\
\hline
all & $0.52\pm0.01$ & $0.45\pm0.01$ & $0.48\pm0.01$ & $0.47\pm0.01$ & \textbf{\boldmath$0.44\pm0.01$} \\
pos & $0.96\pm0.00$ & $0.89\pm0.00$ & $0.82\pm0.00$ & $0.96\pm0.00$ & \textbf{\boldmath$0.63\pm0.01$} \\
neg & $0.36\pm0.00$ & \textbf{\boldmath$0.32\pm0.00$} & $0.36\pm0.00$ & $0.35\pm0.00$ & $0.38\pm0.01$ \\
pos2neg & $0.36\pm0.00$ & \textbf{\boldmath$0.32\pm0.00$} & $0.35\pm0.00$ & $0.35\pm0.00$ & $0.83\pm0.01$ \\
neg2pos & $1.06\pm0.01$ & $1.00\pm0.01$ & $0.96\pm0.01$ & $1.07\pm0.01$ & \textbf{\boldmath$0.80\pm0.01$} \\
\hline
\multicolumn{6}{l}{\textbf{Sodium Estimation} (MAE) $\downarrow$} \\
\hline
all & $4.48\pm0.06$ & $4.04\pm0.07$ & $7.17\pm0.21$ & $3.59\pm0.06$ & \textbf{\boldmath$2.67\pm0.04$} \\
pos & $8.90\pm0.02$ & $8.64\pm0.02$ & $10.63\pm0.04$ & $7.91\pm0.02$ & \textbf{\boldmath$3.19\pm0.08$} \\
neg & $2.47\pm0.00$ & $2.55\pm0.01$ & $5.61\pm0.03$ & $2.46\pm0.01$ & \textbf{\boldmath$2.40\pm0.03$} \\
pos2neg & $2.70\pm0.01$ & $2.68\pm0.02$ & $4.41\pm0.08$ & \textbf{\boldmath$2.43\pm0.02$} & $5.33\pm0.09$ \\
neg2pos & $7.95\pm0.02$ & $7.84\pm0.03$ & $10.33\pm0.09$ & $7.26\pm0.03$ & \textbf{\boldmath$5.26\pm0.09$} \\
\hline
\multicolumn{6}{l}{\textbf{Win Score Matrix (Cumulative Across Tasks)}} \\
\hline
Subset & MOMENT & Chronos-2 & GPT-PPG & MOMENT-PPG & Carry forward \\
\hline
all & 0 & 0 & 0 & 0 & \textbf{4/4} \\
pos & 0 & 0 & 0 & 0 & \textbf{4/4} \\
neg & 0 & 1/4 & 0 & 0 & \textbf{3/4} \\
pos2neg & 0 & 1/4 & \textbf{2/4} & 1/4 & 0 \\
neg2pos & 1/4 & 0 & 1/4 & 0 & \textbf{2/4} \\
\hline
\textbf{Total} & 1/20 & 2/20 & 3/20 & 1/20 & \textbf{13/20} \\
\hline
\end{tabular}
}
\caption{Comparison of the carry-forward clinical-history baseline with four fully fine-tuned foundation models on Inst. A laboratory-status subsets: MOMENT (40M), Chronos-2 (28M), GPT-PPG (19M), and MOMENT-PPG (40M). Results are reported as mean $\pm$ bootstrap standard error (SE). Bold indicates the lowest mean MAE for each subset. Win Score represents the number of wins out of 20 evaluations. The cumulative win score matrix summarizes wins for each subset across all four laboratory estimation tasks.}
\label{tab:carryfwd}
\end{table}

Representative comparisons with traditional signal-processing and morphology-based baselines are reported in the main text (Table~\ref{tab:trad_bline}).

\subsection{Cross-domain robustness and cross-modal transfer}
\label{app:robustness_transfer_details}

This appendix provides the detailed cross-domain and cross-modal analyses summarized in Section~\ref{sec:further_analysis}.

\subsubsection{Cross-domain robustness}
In Table~\ref{tab:robustness}, we evaluate cross-domain robustness of FMs by fine-tuning on one (PPG) dataset and testing on another.
This is more challenging than the usual evaluation because it introduces shifts in sensor hardware, acquisition protocols, cohort composition, and target distributions.
We find that on our chosen 12 evaluations, no model dominates: MOMENT, Chronos-2, MOMENT-PPG, and GPT-PPG without the auxiliary Laplace loss each achieve three best point estimates.
The rankings are strongly task dependent: Chronos-2 leads both atrial-fibrillation transfers, MOMENT-PPG leads both heart-rate transfers, MOMENT performs best on three of the four blood-pressure endpoints, and GPT-PPG without Laplace leads both SpO$_2$ transfers.
The reversal of model rankings across transfer directions, particularly for sodium and blood pressure, further shows that robustness cannot be inferred from only one source-to-target direction.

GPT-PPG is especially sensitive to its fine-tuning objective.
Retaining its auxiliary generative Laplace reconstruction loss produces severe failures for some SpO$_2$, blood-pressure, and sodium transfers, whereas removing it eliminates the largest SpO$_2$ and sodium failures and gives GPT-PPG three wins.
However, this modification is not uniformly beneficial.
This therefore suggests task and domain-dependent interference between the supervised and generative objectives instead of a universally harmful auxiliary loss.
In conclusion, specialist PPG pre-training does not guarantee superior cross-domain robustness, since the general-purpose models remain competitive across several tasks.

\begin{table}[!h]
\centering
\resizebox{\textwidth}{!}{%
\begin{tabular}{l|l|c|c|c|c}
\hline
\multicolumn{2}{c|}{Task/Dataset (Metric)} & MOMENT & Chronos-2 & GPT-PPG (w/ \& w/o Laplace) & MOMENT-PPG \\
\hline
Train set & Test set &  &  &  &  \\
\hline
\multicolumn{6}{l}{\textbf{AF Detection} (F1 score) $\uparrow$} \\
\hline
Stanford & Simband & $45.90\pm4.73\%$ & \textbf{\boldmath$70.94\pm4.73\%$} & $66.15\pm3.45\%$ ($62.22\pm4.16\%$) & $53.55\pm3.62\%$ \\
\hline
Simband & Stanford & $62.23\pm1.36\%$ & \textbf{\boldmath$69.61\pm1.41\%$} & $65.10\pm1.40\%$ ($61.35\pm1.43\%$) & $49.74\pm1.03\%$ \\
\hline
\multicolumn{6}{l}{\textbf{SpO$_2$ Estimation} (MAE) $\downarrow$} \\
\hline
Sleep Disorder & MIMIC & $2.60\pm0.05$ & $2.34\pm0.02$ & $21.06\pm0.22$ (\textbf{\boldmath$2.23\pm0.03$}) & $3.56\pm0.10$ \\
\hline
MIMIC & Sleep Disorder & $4.39\pm0.08$ & $4.10\pm0.08$ & $4.21\pm0.08$ (\textbf{\boldmath$4.06\pm0.08$}) & $4.09\pm0.08$ \\
\hline
\multicolumn{6}{l}{\textbf{Heart Rate Estimation} (MAE) $\downarrow$} \\
\hline
DALIA & BUT-PPG & $14.27\pm0.38$ & $20.15\pm0.45$ & $20.17\pm0.43$ ($27.84\pm0.44$) & \textbf{\boldmath$13.11\pm0.35$} \\
\hline
BUT-PPG & DALIA & $22.72\pm0.14$ & $22.44\pm0.15$ & $24.87\pm0.18$ ($33.09\pm0.21$) & \textbf{\boldmath$22.41\pm0.15$} \\
\hline
\multicolumn{6}{l}{\textbf{Blood Pressure Estimation (Diastolic/Systolic MAE)} $\downarrow$} \\
\hline
\multirow{2}{*}{PPG-BP} & \multirow{2}{*}{BUT-PPG} & \textbf{\boldmath$7.36\pm0.19$} / & $7.48\pm0.20$ / & $72.68\pm0.36$ ($33.83\pm0.40$) / & $7.57\pm0.21$ / \\
& & \textbf{\boldmath$13.95\pm0.42$} & $15.43\pm0.31$ & $94.63\pm0.51$ ($106.38\pm0.58$) & $14.96\pm0.29$ \\
\hline
\multirow{2}{*}{BUT-PPG} & \multirow{2}{*}{PPG-BP} & $10.93\pm0.61$ / & $11.17\pm0.58$ / & $11.62\pm0.58$ ($10.83\pm0.59$) / & \textbf{\boldmath$10.15\pm0.61$} / \\
& & \textbf{\boldmath$16.50\pm1.14$} & $16.69\pm1.20$ & $16.57\pm1.11$ ($21.52\pm1.63$) & $17.38\pm1.40$ \\
\hline
\multicolumn{6}{l}{\textbf{Sodium Estimation} (MAE) $\downarrow$} \\
\hline
MIMIC & Inst. A & $4.79\pm0.06$ & \textbf{\boldmath$3.85\pm0.06$} & $114.47\pm0.10$ ($5.46\pm0.10$) & $3.89\pm0.06$ \\
\hline
Inst. A & MIMIC & $3.78\pm0.04$ & $4.07\pm0.04$ & $6.87\pm0.05$ (\textbf{\boldmath$3.70\pm0.03$}) & $4.10\pm0.04$ \\
\hline
\hline
\multicolumn{2}{l|}{Score-individual} & \textbf{3/12} & \textbf{3/12} & 0/12 (\textbf{3/12}) & \textbf{3/12} \\
\hline
\multicolumn{2}{l|}{Score-overall} & \multicolumn{2}{c|}{\textbf{6/12}} & \multicolumn{2}{c}{\textbf{6/12}} \\
\hline
\end{tabular}
}
\caption{Cross-domain generalization during full fine-tuning of four foundation models: MOMENT (40M), Chronos-2 (28M), GPT-PPG (19M), and MOMENT-PPG (40M). Results are reported as mean $\pm$ bootstrap standard error (SE). For GPT-PPG, values in parentheses correspond to training without the auxiliary Laplace loss. Blood pressure results are presented as diastolic/systolic MAE. Bold indicates the best mean performance across all models and variants for each evaluation. Win scores are computed over 12 cross-domain evaluations, counting diastolic and systolic blood pressure separately.}
\label{tab:robustness}
\end{table}

\begin{table}[!t]
\centering
\small
\resizebox{\textwidth}{!}{%
\begin{tabular}{l|c|c|c|c}
\hline
Task/Signal (Metric) & MOMENT & Chronos-2 & GPT-PPG & MOMENT-PPG \\
\hline
\multicolumn{5}{l}{\textbf{ICP Prediction} (MAE) $\downarrow$} \\
\hline
TCD CBFV signal & $3.89\pm0.10$ & $\bm{3.87\pm0.10}$ & $4.10\pm0.10$ & $3.98\pm0.10$ \\
\hline
\multicolumn{5}{l}{\textbf{Blood Pressure Estimation (Diastolic/Systolic MAE)} $\downarrow$} \\
\hline
\multirow{2}{*}{ECG signal} & $7.73\pm0.14$ / & $\bm{7.63\pm0.14}$ / & $9.30\pm0.20$ / & $8.97\pm0.20$ / \\
& $14.11\pm0.29$ & $\bm{13.86\pm0.28}$ & $18.54\pm0.38$ & $19.87\pm0.42$ \\
\hline
\multicolumn{5}{l}{\textbf{Sleep Detection} (F1 score) $\uparrow$} \\
\hline
EEG signal & $93.50\pm0.30\%$ & $91.79\pm0.33\%$ & $\bm{95.63\pm0.21}\%$ & $94.69\pm0.27\%$ \\
\hline
\textbf{Win Score} & 0/4 & \textbf{3/4} & 1/4 & 0/4 \\
\hline
\end{tabular}
}
\caption{Generalization of four fully fine-tuned foundation models to non-PPG physiological signals: transcranial Doppler cerebral blood flow velocity (TCD CBFV), electrocardiogram (ECG), and electroencephalogram (EEG). Results are reported as mean $\pm$ bootstrap standard error (SE). Blood pressure results are presented as diastolic/systolic MAE. Bold indicates the best mean performance across models. Win Score counts the number of wins across four evaluations, considering diastolic and systolic blood pressure separately.}
\label{tab:nonppg}
\end{table}

\subsubsection{Generalization to cross-modal biosignals}
In Table~\ref{tab:nonppg}, we evaluate whether foundation models (FMs) can adapt beyond photoplethysmography (PPG) to three distinct physiological modalities: transcranial Doppler cerebral blood flow velocity (TCD CBFV), electrocardiography (ECG), and electroencephalography (EEG).
These signals are used for intracranial pressure (ICP) prediction, diastolic and systolic blood pressure estimation, and sleep detection, respectively.
For BP prediction, we use the BUT-PPG dataset, while the other datasets are described in Section~\ref{S419}.
Chronos-2 achieves the best point estimate in all three regression evaluations, with mean absolute errors (MAEs) of 3.87 for ICP and 7.63/13.86 for diastolic/systolic blood pressure.
MOMENT follows closely with corresponding MAEs of 3.89 and 7.73/14.11, whereas the two PPG-specialized models transfer less effectively to ECG-based blood-pressure estimation.
The MOMENT FM has been adopted for forecasting time-series ICP and presents comparative results against LSTM within a 30-minute horizon \cite{vanBhattacharyay2024}.

For EEG-based sleep detection, the ranking reverses: GPT-PPG obtains the highest F1 score of 95.63\%, followed by MOMENT-PPG at 94.69\%, MOMENT at 93.50\%, and Chronos-2 at 91.79\%.
Overall, Chronos-2 wins in 3 of the 4 cases, suggesting strong transfer of its general time-series representations, while GPT-PPG wins on the sleep classification task suggesting its promise on EEG task evaluations.

\subsection{Controlled study of pre-training schemes}
\label{app:pretraining_scheme_details}

\paragraph{Pre-training.}
All PPG pre-training used the same corpus, 9.6M 30-s segments at 40\,Hz from PulseDB (MIMIC-III and VitalDB) and MC-MED. MOMENT-Small (flan-T5-small encoder, 40M parameters) was trained with its masked-reconstruction objective (mask ratio 0.3) on segments resampled to 512 samples (patches of 8), with AdamW (peak learning rate $10^{-4}$, 1{,}000 warm-up steps, cosine decay scheduled over 400{,}000 steps), a batch of 2{,}048 segments and bfloat16 precision; PT starts from random weights and CPT from MOMENT-1-small, with identical settings. Each run was stopped at 26{,}000 steps, with checkpoints every 2{,}000 steps, for three independently seeded CPT/PT pairs. (The first pair stopped at 26{,}773 and 26{,}783 steps under a time limit; two PT runs were resumed with optimizer and scheduler state after interruptions.) Chronos-2 (120M parameters) was trained with its forecasting objective (context 1{,}024, prediction 128) for 112{,}896 steps of 512 segments (six passes over the corpus) with a linear learning-rate decay; PT used a peak learning rate of $10^{-3}$ without warm-up and CPT $10^{-4}$ with 500 warm-up steps, because the higher rate diverged from the pre-trained checkpoint. For the early trajectory, the Chronos-2 pair was re-run with the same recipe and schedule, stopped at 10{,}000 steps with checkpoints every 2{,}000 steps. Because every PPG window forms its own group, Chronos-2's group attention reduces to each series attending to itself; we computed it directly instead of as a masked batch-by-batch attention, which leaves the outputs unchanged (maximum difference 0.0 on real windows) and halves the training time.

\paragraph{Random weights and probes.}
The random reference of MOMENT-Small is the initialization from which each PT run started (three seeds); those of MOMENT-Base and Chronos-2 are fresh initializations of the published architectures (two and three seeds). Each encoder's patch embeddings were mean-pooled and standardized. Regression probes are ridge regressions whose penalty was chosen by leave-one-out cross-validation over $10^{-2}$ to $10^{6}$; classification probes are class-balanced logistic regressions whose inverse penalty was chosen by five-fold cross-validated F1 over $10^{-4}$ to $10^{2}$. Probes were fitted on the training split (at most 50{,}000 windows) and scored on the test split. Of the 45 regression tasks of the 51-task suite, 30 have test $R^2 \le 0$ for every MOMENT model (off-the-shelf, PT and CPT) and are not analyzed; the remaining 15 (9 with $R^2 > 0.2$ for at least one of them, called solvable, and 6 weak) and the 6 classification tasks are. Summaries are medians over tasks, with bootstrap 95\% intervals over tasks.

\begin{table}[h]
\centering
\caption{Frozen features compared with random weights of the same architecture: median difference over tasks (pre-trained minus the mean of the random initializations), tuned probe. The encoders of Section~\ref{sec:arch_ablation} are the 85M Transformer-skeleton models pre-trained on the UCSF corpus (three seeds each, against three random builds of the same configuration).}
\label{tab:pretrain_random}
\footnotesize
\setlength{\tabcolsep}{3pt}
\begin{tabular}{lccc}
\toprule
\textbf{Encoder} & \textbf{Regression (15), $\Delta R^2$} & \textbf{Solvable (9)} & \textbf{Classif.\ (6), $\Delta$F1} \\
\midrule
\multicolumn{4}{l}{\textit{Generalist checkpoints}} \\
MOMENT-Small & $+0.005$ [$-0.001$, $+0.017$] & $+0.016$ & $0.000$ \\
MOMENT-Base & $-0.011$ [$-0.017$, $+0.002$] & $-0.016$ & $-0.012$ \\
MOMENT-Small, stat.-matched & $-0.003$ [$-0.011$, $+0.004$] & $-0.003$ & $-0.023$ \\
Chronos-2 & $+0.051$ [$+0.024$, $+0.094$] & $+0.094$ & $+0.110$ \\
\midrule
\multicolumn{4}{l}{\textit{After PPG pre-training}} \\
MOMENT PT & $+0.067$ [$+0.020$, $+0.194$] & $+0.194$ & $+0.110$ \\
MOMENT CPT & $+0.077$ [$+0.023$, $+0.183$] & $+0.180$ & $+0.093$ \\
Chronos-2 PT & $+0.080$ [$+0.027$, $+0.195$] & $+0.168$ & $+0.108$ \\
Chronos-2 CPT & $+0.065$ [$+0.021$, $+0.130$] & $+0.097$ & $+0.151$ \\
Attention, masked & $+0.058$ [$+0.013$, $+0.158$] & $+0.131$ & $+0.066$ \\
Attention, next-patch & $+0.106$ [$+0.071$, $+0.319$] & $+0.258$ & $+0.111$ \\
Attention, contrastive & $+0.092$ [$+0.034$, $+0.156$] & $+0.156$ & $+0.033$ \\
Local convolution, masked & $+0.073$ [$+0.018$, $+0.221$] & $+0.074$ & $+0.084$ \\
State-space, masked & $+0.070$ [$+0.015$, $+0.221$] & $+0.070$ & $+0.120$ \\
LRU, masked & $+0.075$ [$+0.030$, $+0.228$] & $+0.075$ & $+0.079$ \\
\bottomrule
\end{tabular}
\end{table}

\paragraph{Robustness of MOMENT's equivalence with random weights.}
The equivalence holds wherever we looked (tuned ridge, 15 regression tasks, median difference from a random MOMENT): at every layer of the encoder ($+0.000$ for the patch embedding, $0.000$ to $+0.021$ for blocks 1 to 8), for mean-and-standard-deviation pooling ($+0.008$) and max pooling ($+0.003$) of the last layer, and when the centre 512 samples of each 40-Hz window are fed at their native rate instead of resampling 30\,s to 512 samples ($+0.024$, 10 tasks long enough). For comparison, MOMENT PT is $+0.05$ above random weights at block 1 and $+0.08$ to $+0.10$ from block 2 on, and Chronos-2 is $+0.05$ to $+0.08$ above its random initialization at every block. Linear CKA does not explain the difference: MOMENT's checkpoint and MOMENT PT are equally similar to a random MOMENT (0.73--0.84 and 0.77--0.84 across blocks).

\begin{figure}[h]
\centering
\includegraphics[width=0.6\linewidth]{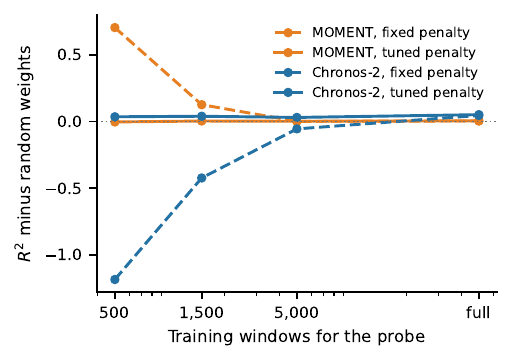}
\caption{Pre-trained minus random weights (median difference in $R^2$ over the 15 regression tasks) against the number of training windows of the probe, for a ridge probe with a fixed penalty ($\alpha = 1$, dashed) and with a tuned penalty (solid).}
\label{fig:probe_artifact}
\end{figure}

\paragraph{Probe regularization.}
With the fixed penalty of a ridge regression with $\alpha = 1$, pre-trained and random features are misranked when the probe has few training windows (Fig.~\ref{fig:probe_artifact}): MOMENT minus random weights is $+0.70$, $+0.13$ and $+0.00$ in $R^2$ at 500, 1{,}500 and 5{,}000 windows, and Chronos-2 minus random weights $-1.18$, $-0.42$ and $-0.06$, because the two kinds of features need different penalties (the tuned penalty is 32 for MOMENT and 100 for its random initialization, 100 for Chronos-2 and 3 for its random initialization). Two further defences against overfitting give the same answer as the tuned ridge at every size from 500 to 20{,}000 windows: PCA truncation with the number of components chosen by cross-validation (MOMENT minus random $-0.005$ to $+0.001$; Chronos-2 $+0.007$ to $+0.051$) and an MLP probe with dropout, weight decay and early stopping (MOMENT $-0.005$ to $+0.018$; Chronos-2 $+0.029$ to $+0.119$).

\begin{table}[h]
\centering
\caption{CPT compared with PT through PPG pre-training: the number of the 15 informative tasks on which CPT is better, and the median relative gain (MAE for regression, F1 for classification), tuned probe. MOMENT: mean $\pm$ s.d. over the three seed pairs. At step 0 the generalist checkpoint is compared with random weights (for MOMENT, the initialization each PT run starts from). End is step 26{,}000 for MOMENT and 112{,}896 for Chronos-2; a MOMENT step is 2{,}048 segments and a Chronos-2 step 512.}
\label{tab:pretrain_headstart}
\footnotesize
\setlength{\tabcolsep}{4pt}
\begin{tabular}{lccccc}
\toprule
 & \textbf{Step 0} & \textbf{2{,}000} & \textbf{6{,}000} & \textbf{10{,}000} & \textbf{End} \\
\midrule
MOMENT, tasks better (of 15) & $8.7 \pm 1.2$ & $12.3 \pm 0.6$ & $10.7 \pm 1.5$ & $10.3 \pm 0.6$ & $9.3 \pm 1.5$ \\
MOMENT, median gain (\%) & $+1.3 \pm 1.3$ & $+4.6 \pm 1.6$ & $+1.7 \pm 1.4$ & $+2.5 \pm 2.1$ & $+0.9 \pm 0.7$ \\
\midrule
Chronos-2, tasks better (of 15) & 14 & 12 & 9 & 9 & 6 \\
Chronos-2, median gain (\%) & $+9.9$ & $+7.9$ & $+1.7$ & $+2.3$ & $-1.2$ \\
\bottomrule
\end{tabular}
\end{table}

\paragraph{What predicts the head start.}
Table~\ref{tab:pretrain_r2pred} gives, for each checkpoint, the Spearman correlation across the 45 regression tasks between CPT's relative gain over PT (MAE) and the off-the-shelf model's $R^2$, the $R^2$ of the same architecture with random weights (mean of the random initializations), and the partial correlation of the off-the-shelf $R^2$ given the random-weight $R^2$ (Spearman ranks, linear residuals). All scores use the tuned ridge probe. The off-the-shelf and random-weight $R^2$ correlate at $0.97$ across tasks for MOMENT and $0.91$ for Chronos-2.

\begin{table}[h]
\centering
\caption{Spearman correlation across the 45 regression tasks between CPT's relative gain over PT and the off-the-shelf $R^2$ (Off.), the random-weight $R^2$ (Rand.), and the partial correlation of the former given the latter (Part.). MOMENT: mean $\pm$ s.d. over the three seed pairs. End is step 26{,}000 for MOMENT and 112{,}896 for Chronos-2.}
\label{tab:pretrain_r2pred}
\footnotesize
\setlength{\tabcolsep}{4pt}
\begin{tabular}{lcccccc}
\toprule
 & \multicolumn{3}{c}{\textbf{MOMENT}} & \multicolumn{3}{c}{\textbf{Chronos-2}} \\
\cmidrule(lr){2-4}\cmidrule(lr){5-7}
\textbf{Step} & Off. & Rand. & Part. & Off. & Rand. & Part. \\
\midrule
2{,}000 & $+0.60 \pm 0.08$ & $+0.60 \pm 0.07$ & $+0.10 \pm 0.08$ & $+0.65$ & $+0.54$ & $+0.45$ \\
4{,}000 & $+0.44 \pm 0.06$ & $+0.44 \pm 0.05$ & $+0.08 \pm 0.06$ & -- & -- & -- \\
6{,}000 & $+0.27 \pm 0.14$ & $+0.24 \pm 0.15$ & $+0.15 \pm 0.00$ & $+0.13$ & $+0.21$ & $-0.13$ \\
10{,}000 & $+0.27 \pm 0.09$ & $+0.28 \pm 0.11$ & $-0.01 \pm 0.14$ & $-0.13$ & $-0.03$ & $-0.24$ \\
16{,}000 & $+0.16 \pm 0.10$ & $+0.19 \pm 0.10$ & $-0.13 \pm 0.18$ & -- & -- & -- \\
20{,}000 & $-0.02 \pm 0.15$ & $+0.01 \pm 0.14$ & $-0.13 \pm 0.07$ & -- & -- & -- \\
End & $+0.17 \pm 0.02$ & $+0.23 \pm 0.02$ & $-0.21 \pm 0.10$ & $-0.21$ & $-0.20$ & $-0.09$ \\
\bottomrule
\end{tabular}
\end{table}

\paragraph{A statistics-matched start and a small late benefit.}
One further CPT run started from a copy of MOMENT in which every weight tensor was redrawn with its own mean and standard deviation, which keeps the scale of the weights but none of their learned structure (one run, with the seed and data order of the first pair); its frozen features are also equivalent to random weights (Table~\ref{tab:pretrain_random}). Early in pre-training it gets about half of CPT's head start on the 9 solvable tasks ($+3.9\%$ against $+7.2\%$ after 2{,}000 steps), but none across all 45 regression tasks ($-0.0\%$ against $+0.5\%$, fixed-penalty probe), so part of the early head start on tasks with signal may come from the scale of MOMENT's weights alone. Late in pre-training (steps 16{,}000--26{,}000), MOMENT's CPT keeps a small benefit over PT (median $+0.9 \pm 0.2\%$ over the three pairs), concentrated on a few tasks: oscillometric diastolic pressure ($+8.0\%$, mean of the pairs), DaLiA heart rate ($+6.7\%$) and echocardiographic LVEF ($+4.5\%$). Which tasks keep it is reproducible across pairs (Spearman between pairs $+0.81 \pm 0.06$ over the 15 regression tasks). The statistics-matched start keeps none of it on the four tasks where the first pair's benefit is largest ($0.0\%$ against $+4.4\%$).

\section{Bias study}
\label{sec:bias_study}
In Table~\ref{tab:type5_hr_gender_fairness}, we investigate whether female--male performance disparities in heart-rate estimation under controlled training imbalance and domain shift
Panel A compares female:male three training-window ratios of 20:80, 50:50, and 80:20 while holding the number of training samples fixed and evaluating on balanced test cohorts using three random seed runs.
DALIA is used only as testing domain due to less participants.
Balanced 50:50 training does not consistently optimize either predictive performance or fairness: it produces the lowest mean absolute error (MAE) in only 3/6 directions for MOMENT, 1/6 for Chronos-2, 1/6 for GPT-PPG, and 0/6 for MOMENT-PPG, while yielding the smallest absolute female--male error gap, $|\Delta|$, in 2/6, 3/6, 1/6, and 1/6 directions, respectively.
Thus, equal representation during fine-tuning is not sufficient to guarantee equal subgroup performance.
Across all 18 train-ratio and test-domain settings (three training ratio and six train-test domains), Chronos-2 provides the strongest aggregate trade-off, achieving both the lowest equal-cell (equal weightage of 18 cases) mean MAE of 7.20 beats per minute and the smallest mean $|\Delta|$ of 1.72 beats per minute.
By comparison, GPT-PPG has the largest aggregate error and disparity, particularly under cross-domain testing, whereas MOMENT-PPG offers the second-lowest aggregate MAE with a subgroup gap comparable to MOMENT.

To test inherent bias of FMs, Panel B examines the frozen representations using a labeled-reference $k$-nearest-neighbor probe.
MOMENT-PPG achieves the lowest MAE in three of the six reference--test directions, MOMENT in two, and GPT-PPG in one, but the model with the lowest MAE does not necessarily have the smallest $|\Delta|$.
More broadly, subgroup-gap direction depends strongly on the test domain: errors are generally lower for female participants on MIMIC and BUT-PPG, whereas female errors are consistently higher on DALIA across models and training ratios.
This indicates that apparent bias is shaped not only by the gender composition of the fine-tuning set but also by domain-specific signal characteristics.

\begin{table*}[t]
\centering
\scriptsize
\setlength{\tabcolsep}{4pt}
\renewcommand{\arraystretch}{1.50}
\resizebox{\textwidth}{!}{%
\begin{tabular}{c|c|c|c|c|c}
\hline
\multicolumn{6}{c}{\textbf{Panel A: Supervised heart-rate estimation under controlled female:male training-window ratios}} \\
\hline
\textbf{Train source} & \textbf{Balanced test} & \textbf{MOMENT} & \textbf{Chronos-2} & \textbf{GPT-PPG} & \textbf{MOMENT-PPG} \\
\hline
MIMIC & MIMIC
& \shortstack{$4.08/4.02/\mathbf{3.95}$\\[3pt]$\Delta$: $-0.64^{\dagger}/-0.87/-1.00$}
& \shortstack{$2.03/2.01/\mathbf{1.95}$\\[3pt]$\Delta$: $-0.45^{\dagger}/-0.45^{\dagger}/-0.54$}
& \shortstack{$\mathbf{2.12}/2.22/2.21$\\[3pt]$\Delta$: $-0.53^{\dagger}/-0.75/-0.73$}
& \shortstack{$\mathbf{2.24}/2.26/2.31$\\[3pt]$\Delta$: $-0.34^{\dagger}/-0.59/-0.78$} \\
\hline
MIMIC & BUT-PPG
& \shortstack{$61.98/58.20/\mathbf{57.87}$\\[3pt]$\Delta$: $-3.16/-3.29/-3.00^{\dagger}$}
& \shortstack{$5.57/\mathbf{4.28}/4.85$\\[3pt]$\Delta$: $-2.09/-1.31^{\dagger}/-1.53$}
& \shortstack{$42.03/40.97/\mathbf{40.66}$\\[3pt]$\Delta$: $+0.34^{\dagger}/+0.44/+0.43$}
& \shortstack{$\mathbf{12.01}/14.07/13.63$\\[3pt]$\Delta$: $-1.77^{\dagger}/-2.02/-1.77^{\dagger}$} \\
\hline
MIMIC & DALIA
& \shortstack{$\mathbf{8.62}/9.11/9.21$\\[3pt]$\Delta$: $+2.21/+2.24/+2.08^{\dagger}$}
& \shortstack{$6.00/6.09/\mathbf{5.24}$\\[3pt]$\Delta$: $+2.87/+2.99/+2.63^{\dagger}$}
& \shortstack{$47.73/\mathbf{47.54}/48.38$\\[3pt]$\Delta$: $+8.91^{\dagger}/+8.99/+8.98$}
& \shortstack{$\mathbf{6.03}/6.34/6.55$\\[3pt]$\Delta$: $+3.23/+3.11/+2.80^{\dagger}$} \\
\hline
BUT-PPG & BUT-PPG
& \shortstack{$13.10/\mathbf{13.06}/13.07$\\[3pt]$\Delta$: $-0.48/-0.45/-0.41^{\dagger}$}
& \shortstack{$\mathbf{5.24}/6.45/7.97$\\[3pt]$\Delta$: $-1.60/-1.19^{\dagger}/-1.33$}
& \shortstack{$10.71/10.54/\mathbf{10.26}$\\[3pt]$\Delta$: $-1.40/-1.32/-1.24^{\dagger}$}
& \shortstack{$\mathbf{9.20}/9.47/9.61$\\[3pt]$\Delta$: $-1.91/-1.64/-1.43^{\dagger}$} \\
\hline
BUT-PPG & MIMIC
& \shortstack{$21.58/\mathbf{21.31}/22.35$\\[3pt]$\Delta$: $-1.47/-1.46^{\dagger}/-1.56$}
& \shortstack{$\mathbf{9.22}/11.68/12.91$\\[3pt]$\Delta$: $-0.08^{\dagger}/-0.29/-0.65$}
& \shortstack{$16.26/16.46/\mathbf{16.17}$\\[3pt]$\Delta$: $-1.13/-0.57/+0.14^{\dagger}$}
& \shortstack{$\mathbf{15.65}/16.05/16.38$\\[3pt]$\Delta$: $-0.55^{\dagger}/-0.68/-0.75$} \\
\hline
BUT-PPG & DALIA
& \shortstack{$22.56/\mathbf{22.41}/23.04$\\[3pt]$\Delta$: $+3.06/+3.01^{\dagger}/+3.05$}
& \shortstack{$\mathbf{11.01}/12.89/14.15$\\[3pt]$\Delta$: $+3.94/+3.59/+3.37^{\dagger}$}
& \shortstack{$\mathbf{27.03}/27.34/28.60$\\[3pt]$\Delta$: $+14.31/+14.30^{\dagger}/+14.67$}
& \shortstack{$\mathbf{15.88}/16.67/16.94$\\[3pt]$\Delta$: $+3.42/+3.23^{\dagger}/+3.23^{\dagger}$} \\
\hline
\multicolumn{2}{c|}{\shortstack{50:50 best or tied\\MAE / smallest $|\Delta|$ (of 6)}}
& $3/6\;/\;2/6$ & $1/6\;/\;3/6$ & $1/6\;/\;1/6$ & $0/6\;/\;1/6$ \\
\hline
\multicolumn{2}{c|}{Equal-cell mean over all 18 settings: MAE / $|\Delta|$}
& $21.64/1.86$ & $\mathbf{7.20}/\mathbf{1.72}$ & $24.29/4.40$ & $10.63/1.85$ \\
\hline
\multicolumn{6}{c}{\textbf{Panel B: Frozen-backbone, labeled-reference k-nearest-neighbor probe}} \\
\hline
\textbf{Reference source} & \textbf{Balanced test} & \textbf{MOMENT} & \textbf{Chronos-2} & \textbf{GPT-PPG} & \textbf{MOMENT-PPG} \\
\hline
MIMIC & MIMIC & $\mathbf{3.76}\;(-0.89)$ & $7.81\;(+0.56)$ & $14.12\;(+0.33^{\dagger})$ & $4.90\;(-0.54)$ \\
\hline
MIMIC & BUT-PPG & $7.39\;(+1.29)$ & $7.25\;(-0.42)$ & $\mathbf{4.05}\;(-1.44)$ & $9.31\;(+0.23^{\dagger})$ \\
\hline
MIMIC & DALIA & $\mathbf{10.56}\;(+0.91^{\dagger})$ & $13.38\;(+3.66)$ & $13.62\;(+4.28)$ & $11.55\;(+1.53)$ \\
\hline
BUT-PPG & BUT-PPG & $10.37\;(-0.68)$ & $12.73\;(+0.10^{\dagger})$ & $10.23\;(-0.58)$ & $\mathbf{7.32}\;(-2.44)$ \\
\hline
BUT-PPG & MIMIC & $19.83\;(-1.25)$ & $18.72\;(-0.81)$ & $17.05\;(-0.86)$ & $\mathbf{15.47}\;(-0.22^{\dagger})$ \\
\hline
BUT-PPG & DALIA & $23.09\;(+2.39^{\dagger})$ & $19.69\;(+2.42)$ & $18.18\;(+3.06)$ & $\mathbf{17.83}\;(+2.82)$ \\
\hline
\end{tabular}%
}
\caption{Panel A reports supervised full-fine-tuning estimates. In every supervised model cell, the first line gives total mean absolute error for female:male training-window ratios 20:80/50:50/80:20; bold marks the lowest MAE within that cell. The second line gives the signed subgroup gap $\Delta=\mathrm{MAE}_{F}-\mathrm{MAE}_{M}$ in the same ratio order; a dagger ($\dagger$) marks the smallest absolute gap within that cell. Test cohorts were fixed at 50:50, validation matched the training ratio, and total training rows were fixed within each source. The ``50:50 best or tied MAE / smallest $|\Delta|$ (of 6)'' row counts, for each foundation model, how often balanced training attained the lowest total MAE and how often it attained the smallest absolute subgroup gap across the six train--test directions. The ``Equal-cell mean over all 18 settings: MAE / $|\Delta|$'' row averages the two training sources, three training ratios, and three test domains with every setting weighted equally rather than by its number of windows or participants. Panel B reports the frozen-backbone k-nearest-neighbor probe as MAE ($\Delta$); it uses labeled balanced reference data and is not label-free zero-shot inference. Bold marks the lowest MAE and a dagger marks the smallest $|\Delta|$ across models for each reference--test direction. Underlying uncertainty used 1000 participant-cluster bootstrap repetitions and, for supervised aggregates, hierarchical seed/participant resampling; intervals are omitted here for readability.}
\label{tab:type5_hr_gender_fairness}
\end{table*}